\documentclass[twoside]{article}
\usepackage[accepted]{aistats2022}
\usepackage[T1]{fontenc}
\usepackage[utf8]{inputenc}  
\usepackage[american]{babel}
\usepackage[usenames,dvipsnames]{xcolor}
\usepackage{amsfonts}
\usepackage[colorlinks=true,citecolor=RoyalPurple]{hyperref}
\usepackage{bigcenter}
\usepackage{amsthm}
\usepackage{amsmath}
\usepackage{amssymb}
\usepackage{dsfont}
\usepackage{bm}
\usepackage{mathtools}
\usepackage{float}
\usepackage{thmtools, thm-restate}
\usepackage[boxed]{algorithm}% http://ctan.org/pkg/algorithm
\usepackage{algpseudocode}% http://ctan.org/pkg/algorithmicx
\usepackage{subfig}
\usepackage{tikz}
\usetikzlibrary{arrows,automata}
\tikzset{terminal state/.style={draw,rectangle,minimum size=.3in}}
\usepackage{enumitem}
\usepackage{xargs}       % Use more than one optional parameter in a new commands
\usepackage{tcolorbox}
\usepackage{tabularx}
\usepackage{array}
\usepackage{colortbl}
\tcbuselibrary{skins}

\newcolumntype{Y}{>{\raggedleft\arraybackslash}X}

\tcbset{tab2/.style={enhanced,fonttitle=\bfseries,fontupper=\footnotesize\sffamily,
colback=yellow!10!white,colframe=red!50!black,colbacktitle=pink!40!white,
coltitle=black,center title}}

\newcommandx{\rlm}[2][1=]{\todo[linecolor=violet,backgroundcolor=violet!25,bordercolor=violet,#1]{\textbf{Romain:} #2}}
\newcommandx{\rtcm}[2][1=]{\todo[linecolor=red,backgroundcolor=red!25,bordercolor=red,#1]{\textbf{Remi:} #2}}

\newcommand*{\mytop}{\mathrel{\scalebox{0.5}{$\top$}}}
\newcommand*{\mybot}{\mathrel{\scalebox{0.5}{$\bot$}}}

\DeclareMathOperator*{\argmax}{argmax}

\tikzstyle{startstop} = [rectangle, rounded corners, minimum width=3cm, minimum height=1cm,text centered, draw=black, fill=red!30]
\tikzstyle{io} = [trapezium, trapezium left angle=70, trapezium right angle=110, minimum width=3cm, minimum height=1cm, text centered, draw=black, fill=blue!30]
\tikzstyle{process} = [rectangle, minimum width=3cm, minimum height=1cm, text centered, draw=black, fill=orange!30]
\tikzstyle{decision} = [diamond, minimum width=3cm, minimum height=1cm, text centered, draw=black, fill=green!30]
\tikzstyle{arrow} = [thick,->,>=stealth]

\newtheorem{theorem}{Theorem}%[section]

\newtheorem{corollary}[theorem]{Corollary}
\newtheorem{lemma}[theorem]{Lemma}

\floatname{algorithm}{Algorithm}

\usepackage{contour}
\usepackage{ulem}

\contourlength{1pt}

\contourlength{0.8pt}

\newcommand{\myuline}[1]{%
  \uline{\phantom{#1}}%
  \llap{\contour{white}{#1}}%
}

\algnewcommand{\InlineIfThen}[2]{% \IfThenElse{<if>}{<then>}{<else>}
  \State \algorithmicif\ #1\ \algorithmicthen\ #2\ }
  
\newcommandx{\yaHelper}[2][1=\empty]{%
\ifthenelse{\equal{#1}{\empty}}%
  { \ensuremath{ \scriptstyle{ #2 } } } % no offset
  { \raisebox{ #1 }[0pt][0pt]{ \ensuremath{ \scriptstyle{ #2 } } } }  % with offset
}

\newcommandx{\yrightarrow}[4][2=\empty, 3=\empty, 4=\empty, usedefault=@]{%
  \ifthenelse{\equal{#4}{\empty}}
  { \xrightarrow[ \protect{ \yaHelper[ #2 ]{ #1 } } ] {}} % there's no text below
  { \xrightarrow[ \protect{ \yaHelper[ #2 ]{ #1 } } ]{ \protect{ \yaHelper[ #4 ]{ #3 } } } } % there's text below
}

\newcommandx{\yleftarrow}[4][2=\empty, 3=\empty, 4=\empty, usedefault=@]{%
  \ifthenelse{\equal{#4}{\empty}}
  { \xleftarrow[ \protect{ \yaHelper[ #2 ]{ #1 } } ] {} } % there's no text below
  { \xleftarrow[ \protect{ \yaHelper[ #2 ]{ #1 } } ]{ \protect{ \yaHelper[ #4 ]{ #3 } } } } % there's text below
}

\newcommandx{\yRightarrow}[4][2=\empty, 3=\empty, 4=\empty, usedefault=@]{%
  \ifthenelse{\equal{#4}{\empty}}
  { \xRightarrow[ \protect{ \yaHelper[ #2 ]{ #1 } } ] {} } % there's no text below
  { \xRightarrow[ \protect{ \yaHelper[ #2 ]{ #1 } } ]{ \protect{ \yaHelper[ #4 ]{ #3 } } } } % there's text below
}

\newcommandx{\yLeftarrow}[4][2=\empty, 3=\empty, 4=\empty, usedefault=@]{%
  \ifthenelse{\equal{#4}{\empty}}
  { \xLeftarrow[ \protect{ \yaHelper[ #2 ]{ #1 } } ] {} } % there's no text below
  { \xLeftarrow[ \protect{ \yaHelper[ #2 ]{ #1 } } ]{ \protect{ \yaHelper[ #4 ]{ #3 } } } } % there's text below
}  

\newcommand{\R}{\mathbb{R}}
\usepackage[round]{natbib}

\begin{document}
% \cfoot{\thepage}



\twocolumn[

\aistatstitle{Beyond the Policy Gradient Theorem \\for Efficient Policy Updates in Actor-Critic Algorithms}

\aistatsauthor{ Romain Laroche \And R\'emi Tachet des Combes }

\aistatsaddress{ \hspace{0.15cm} Microsoft Research Montr\'eal \hspace{2.5cm} Microsoft Research Montr\'eal } ]

\begin{abstract}
    In Reinforcement Learning, the optimal action at a given state is dependent on policy decisions at subsequent states. As a consequence, the learning targets evolve with time and the policy optimization process must be efficient at unlearning what it previously learnt. In this paper, we discover that the policy gradient theorem prescribes policy updates that are slow to unlearn because of their structural symmetry with respect to the value target. To increase the unlearning speed, we study a novel policy update: the gradient of the cross-entropy loss with respect to the action maximizing $q$, but find that such updates may lead to a decrease in value. Consequently, we introduce a modified policy update devoid of that flaw, and prove its guarantees of convergence to global optimality in $\mathcal{O}(t^{-1})$ under classic assumptions. Further, we assess standard policy updates and our cross-entropy policy updates along six analytical dimensions. Finally, we empirically validate our theoretical findings.
    % In Reinforcement Learning, optimality at a given state, and therefore policy decisions, are dependent on optimality and policy decisions at subsequent states. As a consequence, the learning targets evolve with time and the optimization process must be efficient at unlearning past convergence. In this paper, we find that the policy gradient theorem prescribes updates that are slow to unlearn because of their structural symmetry with respect to the value target. While the projection on the probability simplex involved in the direct parametrization breaks this symmetry, it cannot readily be adapted to function approximation. To overcome this limitation, we study a novel policy update speeding up the unlearning: the gradient of the cross-entropy loss with respect to the action maximizing $q$, but find it potentially leads to disadvantageous policy updates. Consequently, we introduce a modified update devoid of that flaw, and prove it guarantees convergence to the global optimum under classic assumptions. We support our claims with theoretical and empirical evidence.
\end{abstract}
% Etant donne l'incertitude que l'on a vis a vis des resultats de deep, je me pose la question si on pourrait pas faire un papier (majoritairement) non deep pour AISTATS et recentrer les contributions sur la nouveaute de CE/MCE et le panorama des proprietes des updates de la litterature:
% convergence/optimality properties alla J&H (ce serait bien, mais pas obligatoire, d'ajouter des preuves pour escort, d'ou la necessite de se concentrer sur la partie theorique)
% gravity well
% damping (qui est ni plus ni moins que les conv rates asymptotiques en fait?)
% unlearning setting
% domino setting
% necessite d'expected updates?
% applicabilite en deep (non pour direct, oui pour les autres)
% + experiences en MDP finis pour montrer les perfs relatives
% Il n'y a que MCE qui cochera toutes les cases (a part pout la necessite d'expected updates).

\section{INTRODUCTION}

The policy gradient theorem derived for the first time in~\cite{Williams1992} is seminal to all the policy gradient theory~\citep{sutton2000policy,konda2000actor,Ahmed2019,kumar2019sample,zhang2020global,qiu2021finite}, and the actor-critic algorithmic innovations~\citep{mnih2016asynchronous,silver2016mastering,vinyals2019grandmaster}. In this paper, we discover that if, starting from a uniform policy, $n$ policy gradient updates have been performed with respect to some values $q$, then at least as many policy gradient updates with respect to opposite values will be needed to return back to a uniform policy. We argue that unlearning in $\mathcal{O}(n)$ is too slow for efficient Reinforcement Learning~\citep[RL,][]{Sutton1998}. Indeed, the optimal action at a given state is dependent on policy decisions at subsequent states. As a consequence, the learning targets evolve with time and an efficient policy optimization process must be fast at unlearning what it previously learnt. We show further that the unlearning slowness of policy gradient updates critically compounds with the number of such chained decisions as well as with decaying learning rates and/or state visitations. The structural flaw in the policy gradient lies in its symmetry with respect to the $q$-function.
% We show further that the unlearning slowness of policy gradient updates critically compounds with the number of such chained decisions and/or when the learning rate and/or the state visitations decay with time. The structural flaw in the policy gradient lies in its symmetry with respect to the $q$-function.

We therefore look for alternative solutions. As described in \cite{Agarwal2019}, the direct parametrization update also applies a classic policy gradient update, but follows it with a projection on the probability simplex. This projection breaks the symmetry, like a wall preventing the parameters from going further forward but still allowing them to go backwards. Unfortunately, an adaptation of the direct parametrization to non-tabular settings, \textit{i.e.} to function approximation, remains an open problem because the projection cannot be readily differentiated.

To overcome this limitation, we study a novel policy update that improves the unlearning speed: the cross-entropy policy update. It consists in updating the parameters with the gradient of the cross-entropy loss between the output of a softmax parametrization and the current local optimal action $a_q(s)=\argmax_{a\in\mathcal{A}}q(s,a)$. This policy update displays a consistent empiric convergence to global optimality in our experiments. But unfortunately, our analysis reveals that such updates may at times lead to a decrease in value, which is a serious dent in its theoretical grounding. We conjecture its convergence and global optimality to be true, but they remain an open problem.

In the meantime, we propose to alter the cross-entropy loss gradient in order to guarantee monotonicity of the value function. We prove that the resulting modified cross-entropy policy update converges in $\mathcal{O}(t^{-1})$ to a global optimal under the set of assumption/conditions made in \cite{Laroche2021}. We pursue our theoretical analysis with an overview of the main policy updates used in the literature, and analyze them along six axes: convergence to global optimality, asymptotic convergence rates, sensitivity to the gravity well exposed in \cite{Mei2020b}, unlearning speed, compatibility with stochastic updates, and adaptability to function approximation. Due to space limitations, proofs for all propositions and theorems were moved to Appendix \ref{app:theory}.

Finally, we empirically validate our analysis on diverse finite MDPs. The results show that the cross entropy softmax updates are as efficient as the direct parametrization updates on hard planning tasks on which policy gradient methods fail to converge to optimality in a reasonable amount of time. Due to space limitations, some details regarding implementation choices, applicative domains, and experimental results were moved to Appendix \ref{app:Experiments}. Moreover, the code is attached to the proceedings.

Our contributions are summarized below:
\begin{itemize}
    \item We identify the slow unlearning behaviour of policy updates following the policy gradient theorem.
    \item We develop two novel policy updates based on the cross-entropy loss tackling the aforementioned issue.
    \item We assess standard policy updates and our policy updates along six analytical dimensions.
    \item We empirically validate our theoretical findings.
\end{itemize}

\section{FAMILIES OF POLICY UPDATES}
\label{sec:updates}
The objective for an agent consists in maximizing the sum of discounted rewards:
\begin{align}
    \!\!\mathcal{J}(\pi) &\doteq \mathbb{E}\!\!\left[\sum_{t=0}^\infty \gamma^t R_t\bigg|\begin{array}{cc}
          \!\!\!\!S_0\sim p_0, S_{t+1}\sim p(\cdot|S_t,A_t),\!\!\!\! \\
          \!\!\!A_t\sim \pi(\cdot|S_t), R_t\sim r(\cdot|S_t,A_t)\!\!\!\!\!
    \end{array} \right]\!\!,\!\! \label{eq:obj}
\end{align}
where state $S_0$ is sampled from the initial state distribution $p_0$, and at each time step $t\geq 0$, action $A_t$ is sampled from the current policy $\pi$, reward $R_t$ is sampled according to the reward kernel $r$, and next state $S_{t+1}$ is sampled according to the transition kernel $p$. $0\leq\gamma<1$ is the discount factor ensuring that the infinite sum of rewards converges.

In this paper, we will consider policies $\pi_\theta$, parametrized by $\theta$, that get recursively updated as follows:
\begin{align}
    \theta' \leftarrow U(\theta, d, q,\eta),
\end{align}
where $d(s)$ is classically the state-visit distribution induced by the current policy $\pi_\theta$, but may be any state distribution following the generalized policy update in \cite{Laroche2021}, $q$ is the current state-action value function estimate for $\pi_\theta$, and $\eta$ is a learning rate scalar. 

\subsection{PG updates: $U_\textsc{pg}$-$U_\textsc{pg-sm}$-$U_\textsc{pg-es}$}
A natural approach is to take the gradient of $\mathcal{J}(\pi_\theta)$ with respect to its parameters. This corresponds to the update prescribed by the policy gradient theorem:
\begin{align}
    U_\textsc{pg}(\theta, d, q,\eta)\! \doteq\! \theta \!+\! \eta \sum_{s\in\mathcal{S}} d(s)\! \sum_{a\in\mathcal{A}} q(s, a) \nabla_{\theta} \pi_\theta(a|s), \label{eq:pg}
\end{align}
where $\nabla_{\theta} \pi_\theta$ denotes the gradient of $\pi_\theta$ with respect to its parameters $\theta$. 
In practice, the vast majority of practitioners use update $U_\textsc{pg}$ in Eq. \eqref{eq:pg} with a softmax parametrization:
\begin{align}
    \pi_\theta(a|s) &\doteq \frac{\exp(f_\theta(s,a))}{\sum_{a'\in\mathcal{A}}\exp(f_\theta(s,a'))} \label{eq:softmax}\\
    % \nabla_\theta \pi_\theta(a|s) &= \pi_\theta(a|s)(1-\pi_\theta(a|s))\nabla_\theta f_\theta,
    \nabla_\theta \pi_\theta(a|s) &= \pi_\theta(a|s) [\nabla_\theta f_\theta(s,a) \nonumber \\ 
    &\hspace{1.5cm}- \sum\nolimits_{a'} \nabla_\theta f_\theta(s,a') \pi_\theta(a'|s)]\label{eq:softmax_grad}
    %\pi_\theta(1-\pi_\theta)\nabla_\theta f_\theta,\label{eq:softmax_grad}
    % \nabla_\theta \pi_\theta(a|s) &= d(s)\pi(a|s)(q_\pi(s,a) - v_\pi(s))\nabla_\theta f_\theta, \\
    % \nabla_\theta \mathcal{J}(\pi_\theta) &= d(s)\pi_\theta(a|s)(q_{\pi_\theta}(s,a) - v_{\pi_\theta}(s))\nabla_\theta f_\theta,
\end{align}
where $f_\theta:\mathcal{S}\times\mathcal{A}\rightarrow\R$ is typically a neural network parametrized by its weights $\theta$. We let $U_\textsc{pg-sm}$ denote the update resulting from this parametrization. % and $(s,a)$ arguments are removed from Eq.\eqref{eq:softmax_grad} for simplicity

Recently, \cite{Mei2020b} discovered that softmax's policy gradient has two issues: slow convergence (aka \textit{damping}) and sensitivity to parameter initialization (aka \textit{gravity well}). They propose the escort transform to address them:
\begin{align}
    \pi_\theta(a|s) &\doteq \frac{|f_\theta(s,a)|^p}{\lVert f_\theta(s,\cdot) \rVert_p^p} \\
    \nabla_\theta \pi_\theta(a|s) &\propto \pi_\theta(a|s)^{1-\frac{1}{p}} [\nabla_\theta f_\theta(s,a) \nonumber \\ 
    &\hspace{-0.6cm}- \sum\nolimits_{a'} \nabla_\theta f_\theta(s,a') \pi_\theta(a|s)^{1 / p} \pi_\theta(a'|s)^{1-\frac{1}{p}} \ \label{eq:escort-grad}% \text{sign}(f_\theta)\pi_\theta^{1-\frac{1}{p}}(1-\pi_\theta)\nabla_\theta f_\theta,
\end{align}
where $p$ is the hyperparameter for the transformation, usually set to 2, $\lVert \cdot \rVert_p$ denotes the $p$-norm, and $\propto$ means proportional to, hiding factors in $p$ and $\lVert f_\theta(s,\cdot) \rVert_p$. We let $U_\textsc{pg-es}$ denote the update resulting from this parametrization.

We will argue in Section \ref{sec:symmetry} that updates of the form of $U_\textsc{pg}$ generally take too much time to unlearn their past steps. Consequently, we investigate other policy updates.

\subsection{Direct parametrization update: $U_\textsc{di}$}
The direct parametrization is arguably the simplest one:
\begin{align}
    % \pi_\theta(a|s) \doteq \theta_{s,a} \quad&\quad \nabla_\theta \pi_\theta = \mathds{1},\\
    \pi_\theta(a|s) \doteq f_\theta(s,a) \quad&\quad \nabla_\theta \pi_\theta = \nabla_\theta f_\theta,
\end{align}
but, since $f_\theta(s,\cdot)$ must live on the probability simplex $\Delta_\mathcal{A}$, an orthogonal projection on $\Delta_\mathcal{A}$ is required after each update $U_\textsc{pg}$:
\begin{align}
    U_\textsc{di}(\theta, d, q,\eta)\doteq \text{Proj}_{\Delta_\mathcal{A}}[U_\textsc{pg}(\theta, d, q,\eta)]
\end{align}
The direct parametrization has been studied quite extensively in the context of finite MDPs. We will argue in Section \ref{sec:deepimpl} that $U_\textsc{di}$ cannot be readily applied with function approximation. We thus investigate other policy updates.

% First, we prove and illustrate that recovering from a convergence to a suboptimal action is costly. Second, we prove that chaining such recoveries induces a cost that is geometric with the length of the chain. In parallel, we prove that the direct parametrization update and the softmax cross-entropy updates do not suffer from such caveat.
\begin{table*}[t!]
\begin{tcolorbox}[tab2,tabularx={c|l||c|c|c|c|c|c}]
    Id                 & Name                    & Conv. \& Opt.        & Rates           & Gravity Well            & Unlearning*            & Stochastic & Deep Impl.          \\\hline\hline
    $U_\textsc{pg-sm}$ & Policy gradient softmax & yes                  & $\mathcal{O}(t^{-1})$  & \textcolor{red}{strong} & \textcolor{red}{slow}* & yes       & yes                 \\\hline   
    $U_\textsc{pg-es}$ & Policy gradient escort  & yes                  & $\mathcal{O}(t^{-1})$  & weak                    & \textcolor{red}{slow}* & yes       & yes                 \\\hline
    $U_\textsc{di}$    & Direct parametrization  & yes                  & exact                  & none*                   & fast*                  & \textcolor{red}{no}      & \textcolor{red}{no} \\\hline   
    $U_\textsc{ce}$    & Cross-entropy softmax*  & \textcolor{red}{no}* & $\mathcal{O}(t^{-1})$  & none*                   & fast*                  & \textcolor{red}{no} & yes                 \\\hline
    $U_\textsc{mce}$   & Modified C-E softmax*   & yes*                 & $\mathcal{O}(t^{-1})$* & none*                   & fast*                  & \textcolor{red}{no} & yes
\end{tcolorbox}
\caption{Summary of the theoretical analysis. The five updates are evaluated across six dimensions. Conv. \& Opt. relates to the existence of a proof of the convergence and global optimality of the update (see Section \ref{sec:conv}). Rates measures the asymptotic speed of convergence (see Section \ref{sec:asymp}). Gravity Well reports how strong they endure a gravitational pull (see Section \ref{sec:gravity}). Unlearning evaluates the speed at which an update can recover from its own past updates (see Section \ref{sec:symmetry}). Stochastic reports whether the update is compatible with stochastic updates (see Section \ref{sec:expected}). Deep Impl. assesses whether the method can be implemented with function approximation (see Section \ref{sec:deepimpl}). The entries have a red font when the answer is considered as disadvantageous. The entries are associated with a star when the answer is novel.}
\label{tab:analysis}
\end{table*}
\subsection{Cross-entropy update: $U_\textsc{ce}$}
Instead of the gradient of the objective function, we propose to follow the gradient derived from the classification problem of selecting the best action according to the current value function $q$: the cross-entropy loss on a softmax parametrization (see Eq. \eqref{eq:softmax}):
\begin{align}
    \!\!\!U_\textsc{ce}(\theta, d, q,\eta) &\doteq \theta + \eta \sum_{s\in\mathcal{S}} d(s) \nabla_{\theta} \log \pi_\theta\left(a_q(s)|s\right), \label{eq:ce1}\\
    (U_\textsc{ce})_{s,a} &= \theta + \eta d(s) \left(\mathds{1}_{a=a_q(s)}-\pi_\theta\left(a|s\right)\right), \label{eq:ce2}
\end{align}
where $a_q(s) = \argmax_{a\in\mathcal{A}} q(s,a)$ is the action that maximizes $q$ in state $s$. The cross-entropy update can be seen as a soft version of the SARSA algorithm (reached as $\eta$ tends to infinity). Unfortunately, the cross-entropy update does not guarantee a monotonous increase of the value. Indeed, an imbalance of the policy for suboptimal actions may lead to an increase in the policy for some of them, and potentially to a decrease in the policy value.

\begin{restatable}[\textbf{Non-monotonicity of $U_\textsc{ce}$}]{proposition}{nonmonotonicity}
    Updating a policy with $U_\textsc{ce}$ may decrease its value.
    \label{prop:non-monotonicity}
\end{restatable}

We will consider $U_\textsc{ce}$ because it works well in practice but the non-monotonicity of its value compromises our proofs of convergence and optimality.

\subsection{Modified cross-entropy update: $U_\textsc{mce}$}
In order to solve the monotonicity issue with $U_\textsc{ce}$, we propose to modify its update such that all suboptimal actions get penalised equally, thereby correcting the imbalance responsible for the non-monotonicity of the update:
\begin{align}
    (U_\textsc{mce})_{s,a_q(s)} &\doteq \theta + \eta d(s) \left(1-\pi_\theta\left(a_q(s)|s\right)\right) \label{eq:mce1} \\
    (U_\textsc{mce})_{s,a\neq a_q(s)} &\doteq \theta - \frac{\eta d(s)}{|\mathcal{A}|-1} \left(1-\pi_\theta\left(a_q(s)|s\right)\right). \label{eq:mce2}
\end{align}

Note that updating $a\neq a_q(s)$ is not necessary but allows $\sum_a \theta_{s,a}$ to remain constant over time, and thus prevents the weights from diverging artificially. We prove its monotonicity under the true updates.

\begin{restatable}[\textbf{Monotonicity of $U_\textsc{mce}$}]{proposition}{monotonicity}
    Updating a policy with $U_\textsc{mce}$ increases its value.
    \label{prop:monotonicity}
\end{restatable}

\subsection{Cross-entropy related updates}

In the context of transfer and multitask learning, \cite{parisotto2016actormimic} train a set of experts on various tasks and then distill the learnt policies into a single agent via the cross-entropy loss. The idea of increasing the probability of greedy actions, while decreasing the probability of bad actions, has also been used in the field of Conservative Policy Iteration~\citep{Kakade2002,Pirotta2013,scherrer2014approximate}. Finally, the Pursuit family of  algorithms introduced in the automata literature also shares some common ground with the idea~\citep{Thathachar1986EstimatorAF,Agache2002}.

\section{THEORETICAL ANALYSIS}
\label{sec:analysis}
In this section, we analyze the five policy updates presented in Section \ref{sec:updates} across six dimensions summarized in Table \ref{tab:analysis}. We first check whether there exists proof of their convergence to global optimality in Section \ref{sec:conv}. In Section \ref{sec:asymp}, we look at their asymptotic convergence rates. In Section \ref{sec:gravity}, we investigate their sensitivity to the gravity well~\citep{Mei2020b}. In Section \ref{sec:symmetry}, we define the unlearning setting and assess the updates' unlearning speed. In Section \ref{sec:expected}, we discuss the compatibility of the update rules to stochastic updates. Finally, we discuss their deep implementations in Section \ref{sec:deepimpl}.

\subsection{Convergence and global optimality}
\label{sec:conv}
The convergence and global optimality of $U_\textsc{pg-sm}$ has been proved under different sets of assumptions/conditions in \cite{Agarwal2019,Mei2020a,Laroche2021}. The convergence and global optimality of $U_\textsc{pg-es}$ has been proved in \cite{Mei2020b}. The convergence and global optimality of $U_\textsc{di}$ has been proved under different sets of assumptions/conditions in \cite{Agarwal2019,Laroche2021}.

While $U_\textsc{ce}$ displays a consistent empiric convergence to optimality in our experiments, the non-monotonicity of its value function is a dent in its theoretical grounding. We conjecture its convergence and global optimality to be true, but they remain an open problem.

Using tools from \cite{Laroche2021}, we prove the convergence of $U_\textsc{mce}$ to the global optimum on finite MDPs:
\begin{restatable}[\textbf{Convergence and optimality of $U_\textsc{mce}$}]{theorem}{optimality}
    Starting from an arbitrary set of parameters $\theta_0$, we consider the process induced by $\theta_{t+1} = U_\textsc{mce}(\theta_t,d_t,q_t,\eta_t),$
    where $q_t=q_{\pi_{\theta_t}}$ is the state-action value of current policy $\pi_{\theta_t}$. Then, under the assumption that the optimal policy is unique, the condition that each state $s$ is updated with weights that sum to infinity over time: $\sum_{t=0}^\infty \eta_t d_t(s)=\infty$,
    is necessary and sufficient to guarantee that the sequence of value functions $(q_t)$ converges to global optimality.
    % \begin{align}
    %     q_\infty = q_\star\doteq\max_{\pi\in\Pi}q_\pi.
    % \end{align}
    \label{thm:optimality}
\end{restatable}

The necessary and sufficient condition on the infinite sum of weights is identical to that for the convergence of $U_\textsc{pg-sm}$ and $U_\textsc{di}$ in the same framework~\citep{Laroche2021}. The optimal policy uniqueness assumption is not required for $U_\textsc{pg-sm}$ and $U_\textsc{di}$, but we conjecture that this is only required for technical reasons and that the theorem holds true even without uniqueness. We run in Section \ref{sec:chainexp} an experiment to empirically support this conjecture.

\subsection{Asymptotic convergence rates}
\label{sec:asymp}
Asymptotic convergence rates for $U_\textsc{pg-sm}$ is well documented to be in $\mathcal{O}(t^{-1})$~\citep{Mei2020a,Laroche2021}. \cite{Mei2020b} prove a convergence rate in $\mathcal{O}(t^{-1})$ for $U_\textsc{pg-es}$. \cite{Laroche2021} prove that $U_\textsc{di}$ will converge to an optimal policy in a finite number of steps.

The convergence of the softmax parametrization under the cross-entropic loss has been studied in the past with a rate of $\mathcal{O}(t^{-1})$~\citep{soudry2018implicit}. Under the assumption that $U_\textsc{ce}$ converges, its rates should be the same. Theorem \ref{thm:rates} shows the same holds true for $U_\textsc{mce}$.
\begin{restatable}[\textbf{Convergence rates of $U_\textsc{mce}$}]{theorem}{rates}
    The process, assumption, and condition defined in Theorem \ref{thm:optimality} guarantee that the sequence of value functions $(q_t)$ asymptotically converges in $\mathcal{O}\left(\left(\sum_{t'=0}^t \eta_{t'} \min_{s\in\mathcal{S}} d_{t'}(s)\right)^{-1}\right)$.
    \label{thm:rates}
\end{restatable}

This is the same rate as $U_\textsc{pg-sm}$ and it reaches rates in $\mathcal{O}\left(t^{-1}\right)$, when the learning rates is kept constant and when off-policy updates prevent $d_{t}(s)\geq d_{\mybot}>0$ from decaying.

The various update rules all converge at least in $\mathcal{O}(t^{-1})$. Given that our setting of interest is RL, where the minimal theoretical cumulative regret is known to be $\mathcal{O}(\log t)$, there is no point in converging faster than $\mathcal{O}(t^{-1})$. For this reason, we consider all the updates converge \textit{fast enough}.

\subsection{Gravity well}
\label{sec:gravity}
\cite{Mei2020b} identify the softmax gravity well issue, \textit{whereby gradient ascent trajectories are drawn towards suboptimal corners of the probability
simplex and subsequently slowed in their progress toward the optimal vertex}. A condition for this to happen is for the action $a_{q_t}$ maximizing $q_t$ not to incur the maximal update step in policy space. Formally, the following condition guarantees the absence of gravity well:
\begin{align}
    a_{q_t}(s) \in \argmax_{a\in\mathcal{A}}\left(\pi_{t+1}(a|s)-\pi_{t}(a|s)\right). \label{eq:gravity_condition}
\end{align}

Theorem \ref{thm:gravitywell} analyses condition \ref{eq:gravity_condition} for all five updates $U_\textsc{pg-sm}$, $U_\textsc{pg-es}$, $U_\textsc{di}$, $U_\textsc{ce}$, and $U_\textsc{mce}$.

\begin{restatable}[\textbf{Gravity well}]{theorem}{gravitywelll}
    $U_\textsc{di}$ and $U_\textsc{mce}$ are guaranteed to satisfy Eq. \eqref{eq:gravity_condition}.
    $U_\textsc{pg-sm}$, $U_\textsc{pg-es}$, and $U_\textsc{ce}$ may transgress Eq. \eqref{eq:gravity_condition}. Furthermore, $U_\textsc{pg-sm}$ may not even satisfy $\pi_{t+1}(a_{q_t}(s)|s)-\pi_{t}(a_{q_t}(s)|s)\geq 0$.
    \label{thm:gravitywell}
\end{restatable}

With policy gradient softmax $U_\textsc{pg-sm}$, we see in the proofs of Th.~\ref{thm:gravitywell} that if $\pi_\theta(a_{q_t}(s)|s)$ is close to 0, it is possible, and even rather easy, for $U_\textsc{pg-sm}$ to induce a larger update step in policy space for a suboptimal action than for $a_{q_t}(s)$ itself. This can last for a significant amount of time because $\pi_\theta(a_{q_t}(s)|s)$ will only observe a small update, hence many steps will be needed to escape the gravity well, as \cite{Mei2020b} empirically observed. Even worse, it may happen that the policy for $a_{q_t}(s)$ decreases.

Similarly to $U_\textsc{pg-sm}$, $U_\textsc{pg-es}$ may induce a larger update step for a suboptimal action than for the optimal one (only when $p>1$). However, that effect is dampened by the $1 - \frac{1}{p}$ power on $\pi_\theta(a|p)$ seen in Eq. \eqref{eq:escort-grad}, and the setting for condition \eqref{eq:gravity_condition} not to be satisfied is much more restricted. Consequently, the gravity well effect exists but is not strong enough to compromise the update efficiency, as predicted by \cite{Mei2020b}.

Regarding $U_\textsc{ce}$, it is rather easy to construct examples where condition \eqref{eq:gravity_condition} is not satisfied. However, we argue that this differs from the gravity well issue, as the action benefiting from a higher update step is not the action with the highest policy. Our experiment confirms that $U_\textsc{ce}$ does not suffer from the gravity well.

Finally, condition \eqref{eq:gravity_condition} is guaranteed to be satisfied with policy updates $U_\textsc{di}$ and $U_\textsc{mce}$.

\subsection{Unlearning what has been learnt}
\label{sec:symmetry}
In RL, the optimal action at a given state is dependent on policy decisions at subsequent states. As a consequence, the learning targets evolve with time and the policy optimization process must be efficient at unlearning what it previously learnt. We use two settings to investigate this property: the unlearning setting measures how fast each algorithm is to recover from bad preliminary $q$ targets, and the domino setting evaluates how this compounds when the bad targets are chained. Our analysis reveals that the symmetry of $U_\textsc{pg}$ with respect to the $q$ target strongly slows down unlearning and that this pitfall compounds geometrically in hard planning tasks.

\subsubsection{Unlearning setting (constant weights)}
% In Reinforcement Learning, local optimality and therefore policy decisions are dependent on further optimality and their corresponding policy decisions. As a consequence, the learning targets evolve with time and the optimization process must be efficient at unlearning the past convergence. 

To evaluate the ability to unlearn convergence stemming from bad preliminary $q$ targets, we consider a single state MDP with two actions and a tabular parametrization $\theta \in \mathbb{R}^2$. Starting from an initial set of parameters $\theta_0$, we apply $n$ updates with $(q(a_1)=1,q(a_2)=0)$ and then measure the number $n'$ of ``opposite'' updates, i.e. with $(q(a_1)=0,q(a_2)=1)$ necessary to unlearn these steps, that is to recover a policy such that $\pi_{\theta_{n+n'}}(a_1)\leq\pi_{\theta_0}(a_1)$.

\begin{restatable}[\textbf{Unlearning setting -- constant weights}]{theorem}{unlearn}
    In the setting where $\eta$ is constant:
    \begin{itemize}
        \item[(i)] Under assumptions enunciated below--and verified by $U_\textsc{pg-sm}$ and $U_\textsc{pg-es}$, $U_\textsc{pg}$ needs $n' \geq n$ updates.
        \item[(ii)] $U_\textsc{di}$ needs $n' = \min\{n\:;\lceil\frac{1}{\eta}\rceil\}$ updates.
        \item[(iii)] $U_\textsc{ce}$ and $U_\textsc{mce}$ need $n' \leq 2 + \frac{1}{\eta}\log(1+2\eta n)$ updates.
    \end{itemize}
    \label{thm:unlearn}
\end{restatable}

Letting $(\theta)_1$ and $(\theta)_2$ denote $\theta$'s components, Theorems \ref{thm:unlearn}(i), \ref{thm:unlearndecay}(i), and \ref{thm:domino}(i) require the following mild assumptions:
\begin{align*}
    &\text{Invariance w.r.t. }\theta\text{: } \pi_{\theta_1} = \pi_{\theta_2} \implies \nabla_\theta \pi_{\theta_1} = \nabla_\theta \pi_{\theta_2}\\
    &\text{Monotonicity: } \pi_\theta(a_1)\! \geq \!\pi_0(a_1) \!\!\! \implies \!\!\! \left\{\!\!\! \begin{array}{l}
         (\nabla_\theta \pi_{\theta}(a_1))_{1}\geq 0 \\
         (\nabla_\theta \pi_{\theta}(a_1))_{2}\leq 0 
    \end{array}\right.\\
    % &\text{Monotonicity2: } \pi_\theta(a_1)\! \geq \!\pi_0(a_1) \!\!\! \implies \!\!\! \left\{\!\!\! \begin{array}{l}
    %      (\nabla_\pi \pi_{\theta}(a_1))_{a_1}\geq 0 \\
    %      (\nabla_\pi \pi_{\theta}(a_1))_{a_2}\leq 0 
    % \end{array}\right.\\
    % &\text{Concavity: } \pi_\theta(a_1)\! \geq \!\pi_0(a_1) \!\!\! \implies \!\!\! \left\{\!\!\! \begin{array}{l}
    %      (\nabla_\theta \pi_{\theta}(a_1))_{1} \text{ decreases} \\
    %      (\nabla_\theta \pi_{\theta}(a_1))_{2} \text{ increases} 
    % \end{array}\right.\\
    &\text{Concavity: } \pi_\theta(a_1)\! \geq \!\pi_0(a_1) \!\!\! \implies \!\!\! \left\{\!\!\!\! \begin{array}{l}
         (\nabla_{\pi}(\nabla_\theta \pi_{\theta}(a_1))_{1})_{a_1}\!\! \leq \!0 \\
         (\nabla_{\pi}(\nabla_\theta \pi_{\theta}(a_1))_{2})_{a_1}\!\! \geq \!0
    \end{array}\right.
    % &\text{Concavity2: } \pi_\theta(a_1)\! \geq \!\pi_0(a_1) \!\!\! \implies \!\!\! \left\{\!\!\!\! \begin{array}{l}
    %      (\nabla_{\pi}(\nabla_\pi \pi_{\theta}(a_1))_{a_1})_{a_1}\!\! \leq \!0 \\
    %      (\nabla_{\pi}(\nabla_\pi \pi_{\theta}(a_1))_{a_2})_{a_1}\!\! \geq \!0
    % \end{array}\right.
\end{align*}

Invariance w.r.t. $\theta$ states that any two parameters implementing the same policy have equal gradients--our theorems do not deal with overparametrization. Monotonicity states that the gradient is monotonic with $\pi$: higher components imply higher policy. It is fairly standard to expect each parameter to be assigned to an action. Concavity states that, as the policy for $a_1$ grows, the absolute value of its gradient with respect to $\theta$ diminishes. Since $\pi_\theta(a_1)$ is a function of $\theta$ from $\mathbb{R}^2$ to $[0,1]$, it is expected that the gradient diminishes as $\pi_\theta(a_1)$ grows to 1.

Theorem \ref{thm:unlearn} states that to unlearn $n$ steps towards a certain action, $U_\textsc{pg}$ requires a number of updates $n'$ at least as large as the number of steps taken initially. As shown in our experiments, this may significantly slow down convergence to the optimal policy. In contrast, $U_\textsc{di}$ updates display no asymptotic dependency in $n$, while $U_\textsc{ce}$ and $U_\textsc{mce}$ updates require a logarithmic number of steps to recover.

% \begin{itemize}
%     \item $\nabla_\theta \pi_{\theta}$ only depends on $\theta$ via $\pi_\theta$,
%     \item $(\nabla_\theta \pi_{\theta}(a_1))_{1}$ (resp. $(\nabla_\theta \pi_{\theta}(a_1))_{2}$) is a positive and decreasing (resp. negative and increasing) function of $\pi_\theta(a_1)$ when $\pi_\theta(a_1) \geq \pi_0(a_1) $.
% \end{itemize}
% The softmax parametrization $U_\textsc{pg-sm}$ is indeed monotonously increasing and concave as long as $\pi_\theta(a_1)\geq\frac{1}{2}$ and therefore falls under Theorem \ref{thm:unlearn}(i). The escort transformation $U_\textsc{pg-es}$ is also concave when $\pi_\theta(a_1)\geq\frac{1}{2}$, but not monotonically increasing as behavior is unclear when the parameters go below 0. This could be regarded as a feature since it embeds some form of asymmetry. However, we consider it as a bug as either $\eta$ is low enough so that it never happens, or $\eta$ is set high and the algorithm may diverge as our experiments reveal (see Section \ref{sec:empirical}). Assuming $\eta$ is such that $\theta$ never goes below 0, then $U_\textsc{pg-es}$ falls under Theorem \ref{thm:unlearn}(i). In contrast, $U_\textsc{di}$ updates display no asymptotic dependency in $n$, and $U_\textsc{ce}$ and $U_\textsc{mce}$ updates require a logarithmic number of steps to recover from past convergence.

\subsubsection{Unlearning setting (decaying weights)}
In practice, the learning weight $\eta_t d_t(s)$ applied to the gradient is likely to decay over time, either because the learning rate $\eta_t$ is required to decay to guarantee convergence of stochastic policy updates, and/or because the state density $d_t(s)$ will decay (depending on $s$) as the policy converges.

In order to model this scenario, we reproduce the unlearning setting with a decaying learning rate: $\forall t\geq 1, \eta_t = \frac{\eta_1}{\sqrt{t}}$ for $\eta_1>0$. From time $t=1$ to $t=n$, updates with $\eta_t$ and $(q(a_1)=1,q(a_2)=0)$ are performed, from time $t=n+1$ onwards, updates with $\eta_t$ and $(q(a_1)=0,q(a_2)=1)$ are applied. We then study $n'$ such that $\pi_{\theta_{n+n'}}(a_1)\leq\pi_{\theta_0}(a_1)$.

\begin{restatable}[\textbf{Unlearning setting -- decaying weights}]{theorem}{unlearndecay}
    In the setting where $\eta_t = \frac{\eta_1}{\sqrt{t}}$:
    \begin{itemize}
        \item[(i)] %If $\pi_0(a_1) \geq \frac{1}{2}$ (resp. $\pi_0(a_1) \geq \frac{p}{2p-1}$), then $U_\textsc{pg-sm}$ (resp. $U_\textsc{pg-es}$)
        Under the assumptions enunciated above--verified by $U_\textsc{pg-sm}$ and $U_\textsc{pg-es}$, $U_\textsc{pg}$ needs $n' \geq 3n - 4\sqrt{n} + 1$.
        \item[(ii)] $U_\textsc{di}$ updates require at most $n'$ equal to\\ $\hspace*{-5pt}\textstyle\min\left\{3n + 4\sqrt{n} + 1\:\:;\: \left(\frac{1}{\eta_1}+1\right)^{2} +\sqrt{n}\left(2+\frac{2}{\eta_1}\right)\right\}.$
        \item[(iii)] $U_\textsc{ce}$ and $U_\textsc{mce}$ updates require at most $n'$ equal to \\$\hspace*{0.6cm}\textstyle\left(4+\frac{\log(1+4\eta_1 \sqrt{n})}{2\eta_1}\right)^{2} +\sqrt{n}\left(8+\frac{\log(1+4\eta_1 \sqrt{n})}{\eta_1}\right).$
    \end{itemize}
    \label{thm:unlearndecay}
\end{restatable}

All updates are severely affected by the decaying weights. Since the decay can stem from the learning rate actually decaying and/or from the state visit $d_t(s)$ decreasing as the behavioural policy converges, these worsened recovery rates are an argument in favor of (i) using expected updates in action space~\citep{Ciosek2018} to allow the use of a constant learning rate, and (ii) performing off-policy updates~\citep{Laroche2021} to properly control the policy update state visitation density.

\subsubsection{Domino setting}
Next, we argue that the number of updates compounds exponentially with the horizon of the task in the following sense: for a $q$ estimate to flip in one state, it is required that all future states have flipped beforehand. To illustrate this effect, we consider the domino setting: several binary decisions are taken sequentially in states $s\in\mathcal{S}=\{s_k\}_{k\in[|\mathcal{S}|]}$. The $q$-function is artificially designed as follows:
\begin{align*}
    q(s_{|\mathcal{S}|},a_1) = 0 \quad\text{and}\quad &q(s_{|\mathcal{S}|},a_2) = 1 \\
    \text{if } \pi_{\theta_t}((a_2|s_{k+1}) \leq \pi_{\theta_0}(a_2|s_{k+1}),\;&\left\{\begin{array}{l}
         q(s_k,a_1) = 1  \\
         q(s_k,a_2) = 0,
    \end{array}\right. \\
    \hspace{-0.2cm}\text{otherwise,}\;&\left\{\begin{array}{l}
         q(s_k,a_1) = 0  \\
         q(s_k,a_2) = 1.
    \end{array}\right.
\end{align*}
Intuitively, the decision in state $s_{k+1}$ needs to be correct for the gradient in state $s_k$ to point in the right direction. When this happens, we say that state $s_{k+1}$ has flipped. We say that the domino setting has been solved in $t$ steps when $\pi_{\theta_t}(a_2|s_{1}) > \pi_{\theta_0}(a_2|s_{1})$.
\begin{restatable}[\textbf{Domino setting}]{theorem}{domino}
    Under the assumption that $\eta$ and $d(s_k)>0$ are constant, in order to solve the setting,
    \begin{itemize}
        \item[(i)] $U_\textsc{pg}$ updates require at least $2^{|\mathcal{S}|-1}$ steps.
        \item[(ii)] $U_\textsc{di}$ updates require at most $1+\frac{|\mathcal{S}|-1}{\eta}$ steps.
        \item[(iii)] $U_\textsc{ce}$ and $U_\textsc{mce}$ updates require at most: \\$\hspace*{2cm}\frac{32e^{8\eta+3}}{\eta^3}(|\mathcal{S}|-1)\log (|\mathcal{S}|-1)\text{ steps}.$
    \end{itemize}
    \label{thm:domino}
\end{restatable}

\myuline{Disclaimer:} The domino setting is a thought experiment for the backpropagation of an optimal policy through a decision chain. However, in practice, the updates will not flip this way. Moreover, its implementation requires a reward function of amplitude $2^{|\mathcal{S}|-1}$. We acknowledge that the domino setting makes things look worse than they really are, but we claim, with the validation of our empirical experiments, that the unlearning slowness of the policy gradient updates are a critical burden for hard planning tasks.

\subsection{Stochastic versus expected updates}
\label{sec:expected}
In Section \ref{sec:symmetry}, we showed drawbacks of $U_\textsc{pg}$'s symmetry property w.r.t. the $q$ function. It can, however, also be an asset, as it allows for stochastic updates. $U_\textsc{di}$'s projection on the simplex breaks the symmetry but, fortunately, only at the frontier of its domain. So, $U_\textsc{di}$ also allows for stochastic updates as long as they obey the classic Robbins-Monro conditions. In contrast, the cross-entropy updates $U_\textsc{ce}$ and $U_\textsc{mce}$ are asymmetric everywhere. Moreover, they ignore the $q$-value gaps, and are thus biased with stochastic updates. Therefore, they must use expected updates~\citep{Ciosek2018}. Given Theorem \ref{thm:unlearndecay}'s analysis, we argue that applying expected updates is actually necessary no matter the update type, so that constant learning rates can be used, and efficient unlearning performance attained.

\subsection{Implementation with function approximation}
\label{sec:deepimpl}
To the best of our knowledge, all actor-critic algorithms with function approximations have implemented policy gradient updates $U_\textsc{pg}$ with or without the true state visitation density: often omitting the discount factor in the state visitation~\citep{Thomas2014,Nota2020,Zhang2020deeper}, and sometimes not correcting off-policy updates~\citep{wang2016sample,espeholt2018impala,vinyals2019grandmaster,schmitt2020off,zahavy2020self}. A vast majority of the implementations also chose by default the softmax parametrization $U_\textsc{pg-sm}$. However, given their shape, implementing cross-entropy updates $U_\textsc{ce}$ and $U_\textsc{mce}$ should be straightforward. The assessment of their actual efficiency is left for future work. Implementing $U_\textsc{pg-es}$ ought not be of any challenge. Nevertheless, since it was already an issue with tabular representations, the instability of $U_\textsc{pg-es}$ when the parameters cross 0 may be a source of concern with function approximation.

The direct parametrization $U_\textsc{di}$ remains to be discussed for function approximation. Sparsemax~\citep{Martins2016,Laha2018} could be seen as an implementation, but, since it omits the projection step, it misses one of its fundamental feature: asymmetry. Indeed, if an action is already assigned a 0 probability, a proper direct update would either decrease it and then project it back to 0, or increase it. In contrast, Sparsemax exhibits a null gradient and therefore immobility in both directions. Adapting $U_\textsc{di}$ to function approximation remains an open problem to this date.

% \begin{figure}[t!]
% 	\centering
% 	\subfloat[\textsc{NoExplo} ($\approx$1000 runs)]{
% 		\includegraphics[trim = 5pt 5pt 5pt 20pt, clip, width=0.41\textwidth]{figures/randommdps/noexplo/results.pdf}
% 		\label{fig:randommdpsno}
% 	}\\
% 	\subfloat[\textsc{LowOffPol} ($\approx$1000 runs)]{
% 		\includegraphics[trim = 5pt 5pt 5pt 20pt, clip, width=0.41\textwidth]{figures/randommdps/lowoffpol/results.pdf}
% 		\label{fig:randommdpslow}
% 	}\\
% 	\subfloat[\textsc{HiOffPol} ($\approx$1000 runs)]{
% 		\includegraphics[trim = 5pt 5pt 5pt 20pt, clip, width=0.41\textwidth]{figures/randommdps/hioffpol/results.pdf}
% 		\label{fig:randommdpshi}
% 	}
% 	\caption{Random MDPs: number of steps to obtain performance equal to 95\% of the gap between $\pi_\star$ and $\pi_u$.}
% 		\label{fig:randommdps}
% \end{figure}

    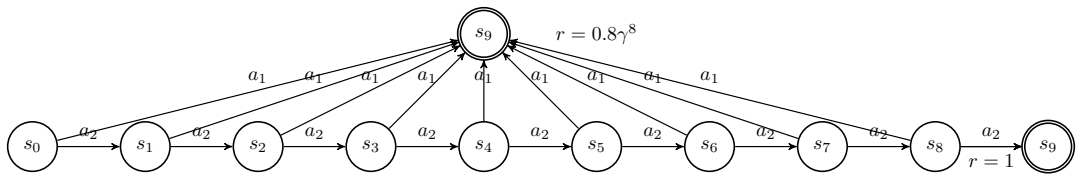
\begin{figure*}[t]
        \begin{center}
            \scalebox{0.75}{
                \begin{tikzpicture}[->, >=stealth', scale=0.6 , semithick, node distance=2cm]
                    \tikzstyle{every state}=[fill=white,draw=black,thick,text=black,scale=1]
                    \node[state]    (x0)                {$s_0$};
                    \node[state]    (x1)[right of=x0]   {$s_1$};
                    \node[state]    (x2)[right of=x1]   {$s_2$};
                    \node[state]    (x3)[right of=x2]   {$s_3$};
                    \node[state]    (x4)[right of=x3]   {$s_4$};
                    \node[state]    (x5)[right of=x4]   {$s_5$};
                    \node[state]    (x6)[right of=x5]   {$s_6$};
                    \node[state]    (x7)[right of=x6]   {$s_7$};
                    \node[state]    (x8)[right of=x7]   {$s_8$};
                    \node[state,accepting]    (x9)[right of=x8]   {$s_9$};
                    \node[state,accepting]    (x-1)[above of=x4]   {$s_9$};
                    \node[] (y)[above of=x5] {$r=0.8\gamma^8$};
                    \path
                    (x0) edge[above]    node{$a_1$}     (x-1)
                    (x1) edge[above]    node{$a_1$}     (x-1)
                    (x2) edge[above]    node{$a_1$}     (x-1)
                    (x3) edge[above]    node{$a_1$}     (x-1)
                    (x4) edge[above]    node{$a_1$}     (x-1)
                    (x5) edge[above]    node{$a_1$}     (x-1)
                    (x6) edge[above]    node{$a_1$}     (x-1)
                    (x7) edge[above]    node{$a_1$}     (x-1)
                    (x8) edge[above]    node{$a_1$}     (x-1)
                    (x0) edge[above]    node{$a_2$}     (x1)
                    (x1) edge[above]    node{$a_2$}     (x2)
                    (x2) edge[above]    node{$a_2$}     (x3)
                    (x3) edge[above]    node{$a_2$}     (x4)
                    (x4) edge[above]    node{$a_2$}     (x5)
                    (x5) edge[above]    node{$a_2$}     (x6)
                    (x6) edge[above]    node{$a_2$}     (x7)
                    (x7) edge[above]    node{$a_2$}     (x8)
                    (x8) edge[above]    node{$a_2$}     (x9)
                    (x8) edge[below]    node{$r=1$}   (x9);
                \end{tikzpicture}
            }
            \caption{Deterministic chain with $|\mathcal{S}|=10$. Initial state is $s_0$. Reward is 0 except when entering final state $s_9$, which is represented with two icons for clarity but is a single state. $r(\cdot,a_2)$ is set such that $q(s_0,a_1)=0.8 q_\star(s_0,a_2)$.}
            \label{fig:chain-env}
        \end{center}
        \vspace{-20pt}
    \end{figure*}
\begin{figure*}[t!]
	\centering
	\subfloat[\textsc{NoExplo} ($\approx$1000 runs)]{
		\includegraphics[trim = 5pt 5pt 5pt 20pt, clip, width=0.32\textwidth]{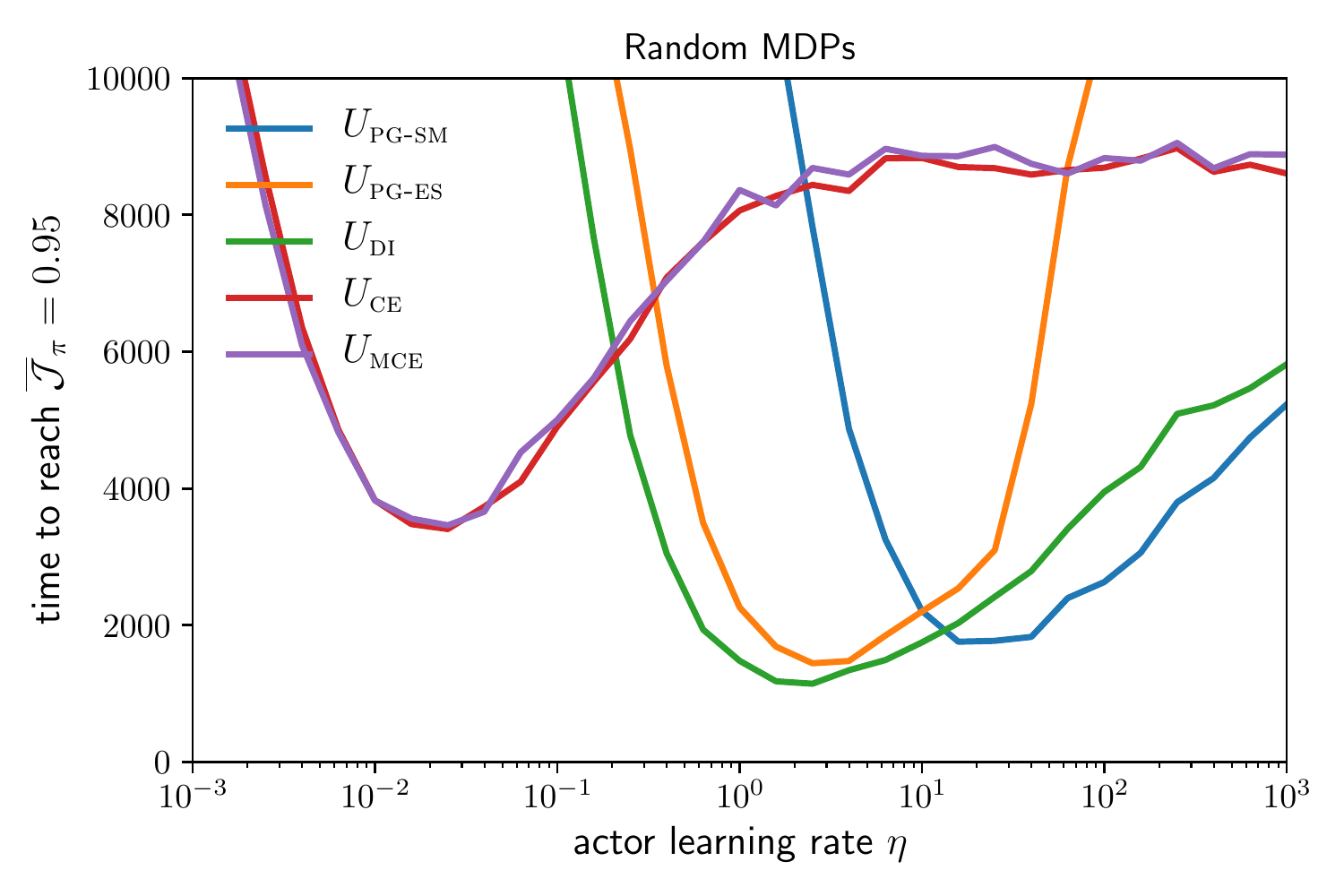}
		\label{fig:randommdpsno}
	}
	\subfloat[\textsc{LowOffPol} ($\approx$1000 runs)]{
		\includegraphics[trim = 5pt 5pt 5pt 20pt, clip, width=0.32\textwidth]{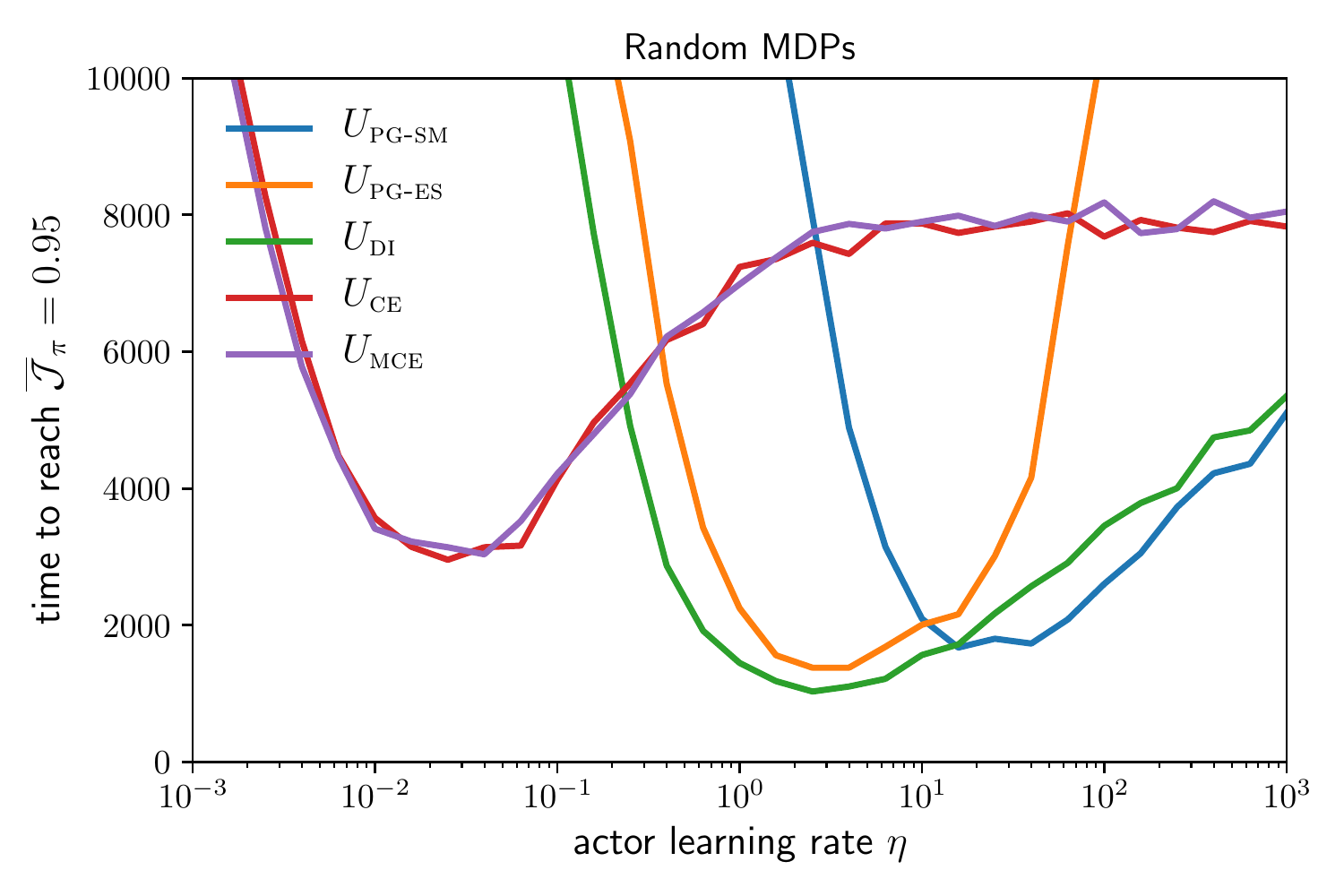}
		\label{fig:randommdpslow}
	}
	\subfloat[\textsc{HiOffPol} ($\approx$1000 runs)]{
		\includegraphics[trim = 5pt 5pt 5pt 20pt, clip, width=0.32\textwidth]{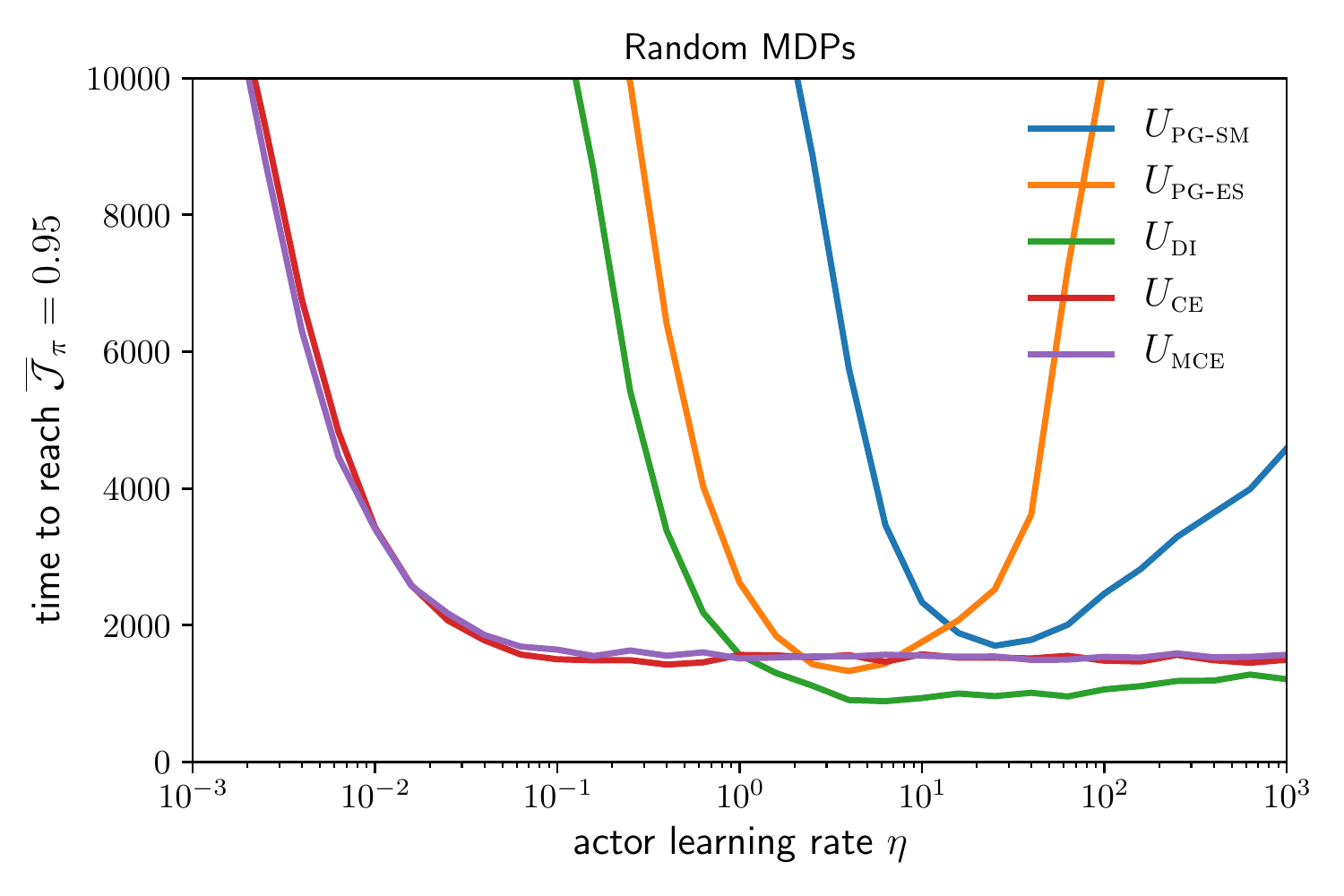}
		\label{fig:randommdpshi}
	}
	\caption{Random MDPs: number of steps to obtain performance equal to 95\% of the gap between $\pi_\star$ and $\pi_u$.}
		\label{fig:randommdps}
\end{figure*}

\section{EMPIRICAL ANALYSIS}
\label{sec:empirical}
This section intends to compare the five updates studied until now in RL experiments (with $p=2$ for $U_\textsc{pg-es}$). In RL, many confounding factors such as exploration or the nature of the environment may compromise the empirical analysis. We will endeavour to control these factors by:
\paragraph{Investigating three exploratory and off-policy updates schemes} designed within the J\&H algorithm\footnote{The precise J\&H implementation is detailed in Appendix \ref{app:Algorithms}.}~\citep{Laroche2021}, where $\epsilon_t$ denotes the probability to give control to Hyde, a pure exploratory agent, at the beginning of each trajectory, and where $o_t$ denotes the probability of updating the actor with a sample collected with Hyde (therefore likely to be off-policy):
        \begin{itemize}
            \item \textsc{NoExplo}: no exploration: $\epsilon_t=0$ and $o_t=0$,
            \item \textsc{LowOffPol}: exploration and low off-policy updates: $\epsilon_t=\min\{1\:;\frac{10}{\sqrt{t}}\}$ and $o_t=\min\{1\:;\frac{10}{\sqrt{t}}\}$,
            \item \textsc{HiOffPol}: exploration and high off-policy updates: $\epsilon_t=\min\{1\:;\frac{10}{\sqrt{t}}\}$ and $o_t=\frac{1}{2}$.
        \end{itemize}
\paragraph{Evaluating on three RL environments}%(full details in Appendix \ref{app:Domains})
        \begin{itemize}
            \item \myuline{Random MDPs:} procedurally generated environments where efficient planning is unlikely to matter and where stochasticity plays an important role.
            \item \myuline{Chain domain:} a deterministic domain with rewards misleading towards suboptimal policies, and thus where planning is the main challenge. The chain domain evaluates the unlearning ability of the updates.
            \item \myuline{Cliff domain:} a slight modification of the chain domain in order to create gravity wells. In addition to their unlearning abilities, the cliff domain should evaluate the updates' resilience to the gravity well pitfall.
        \end{itemize}

\subsection{Random MDPs experiments}
A Random MDP environment with $|\mathcal{S}|=100$ and $|\mathcal{A}|=4$ is procedurally generated at the start of each run. The transition from each state action pair stochastically connects to two states uniformly sampled from $\mathcal{S}$ and with probability generated from a uniform partition of the segment [0,1]. In most cases, the obtained MDP is strongly connected and exploration is barely an issue. However, stochasticity is strong and an accurate $q$ estimate is necessary to find a policy with a good performance.

We evaluate the policies with the number of steps (each step is a transition and an update) they need to reach a normalized performance $\overline{\mathcal{J}}_\pi$ equal to 95\% of the gap between the performance of the optimal policy $\pi_\star$ and that of the uniform one $\pi_u(\cdot|\cdot)=\frac{1}{|\mathcal{A}|}$, formally $\overline{\mathcal{J}}_\pi = \frac{\mathcal{J}(\pi)-\mathcal{J}(\pi_u)}{\mathcal{J}(\pi_\star) - \mathcal{J}(\pi_{u})}$.

The result of the Random MDPs experiments is presented on Figure \ref{fig:randommdps}, where we display the number of steps to reach $\overline{\mathcal{J}}(\pi)=0.95$ versus the learning rate for the actor. The three exploration/off-policiness settings are shown on three separate subfigures: \textsc{NoExplo} in \ref{fig:randommdpsno},  \textsc{LowOffPol} in \ref{fig:randommdpslow}, and  \textsc{HiOffPol} in \ref{fig:randommdpshi}.

The similarity of \textsc{NoExplo} and \textsc{LowOffPol} confirms that the exploration plays a minimal role in this setting. More generally, the policy gradient updates $U_\textsc{pg-sm}$ and $U_\textsc{pg-es}$ are quite unchanged by the presence of off-policy updates, maybe indicating that their slowness to change policies makes them less likely to follow fine $q$ optimality. It is also worth noting that $U_\textsc{pg-es}$ has the narrowest learning rate bandwidth inside which it is efficient: lower, the updates are too slow, higher, the updates get too large and induce divergence because of the \textit{bouncing} behaviour in 0.
$U_\textsc{di}$ gets the best performance in all settings.

$U_\textsc{di}$, $U_\textsc{ce}$, and $U_\textsc{mce}$ all converge to SARSA when $\eta$ tends to $\infty$ and by extrapolation, we may imagine where their curves would meet and therefore SARSA's performance. Thus, we observe that all the policy update algorithms do much better than SARSA on \textsc{NoExplo} and \textsc{LowOffPol}. $U_\textsc{ce}$ and $U_\textsc{mce}$ perform equally, but their strong similarity to SARSA make them the worst on \textsc{NoExplo} and \textsc{LowOffPol}. Indeed, the stochasticity in the environment makes the $q$ predictions of the critic unstable, which prevents the cross-entropy updates to converge. Off-policy updates help because they allow to quickly fix bad $q$ prediction. With high off-policy updates, $U_\textsc{ce}$ and $U_\textsc{mce}$ perform much better with a wide range for the learning rate.

\begin{figure*}[t!]
	\centering
	\subfloat[\textsc{NoExplo}: $|\mathcal{S}|=5$ (100 runs)]{
		\includegraphics[trim = 5pt 5pt 5pt 20pt, clip, width=0.32\textwidth]{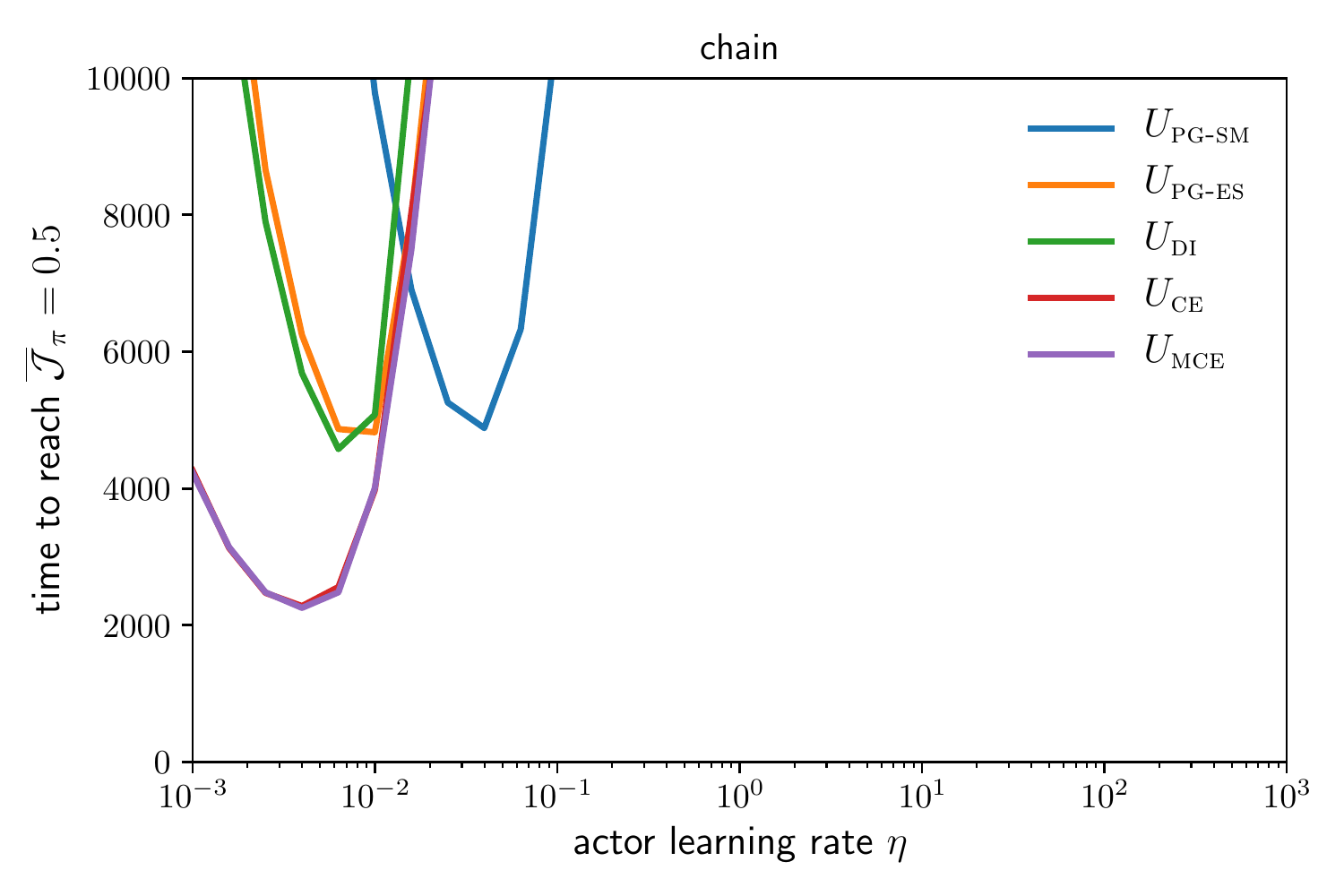}
		\label{fig:chainno}
	}
	\subfloat[\textsc{LowOffPol}: $|\mathcal{S}|=7$ (100 runs)]{
		\includegraphics[trim = 5pt 5pt 5pt 20pt, clip, width=0.32\textwidth]{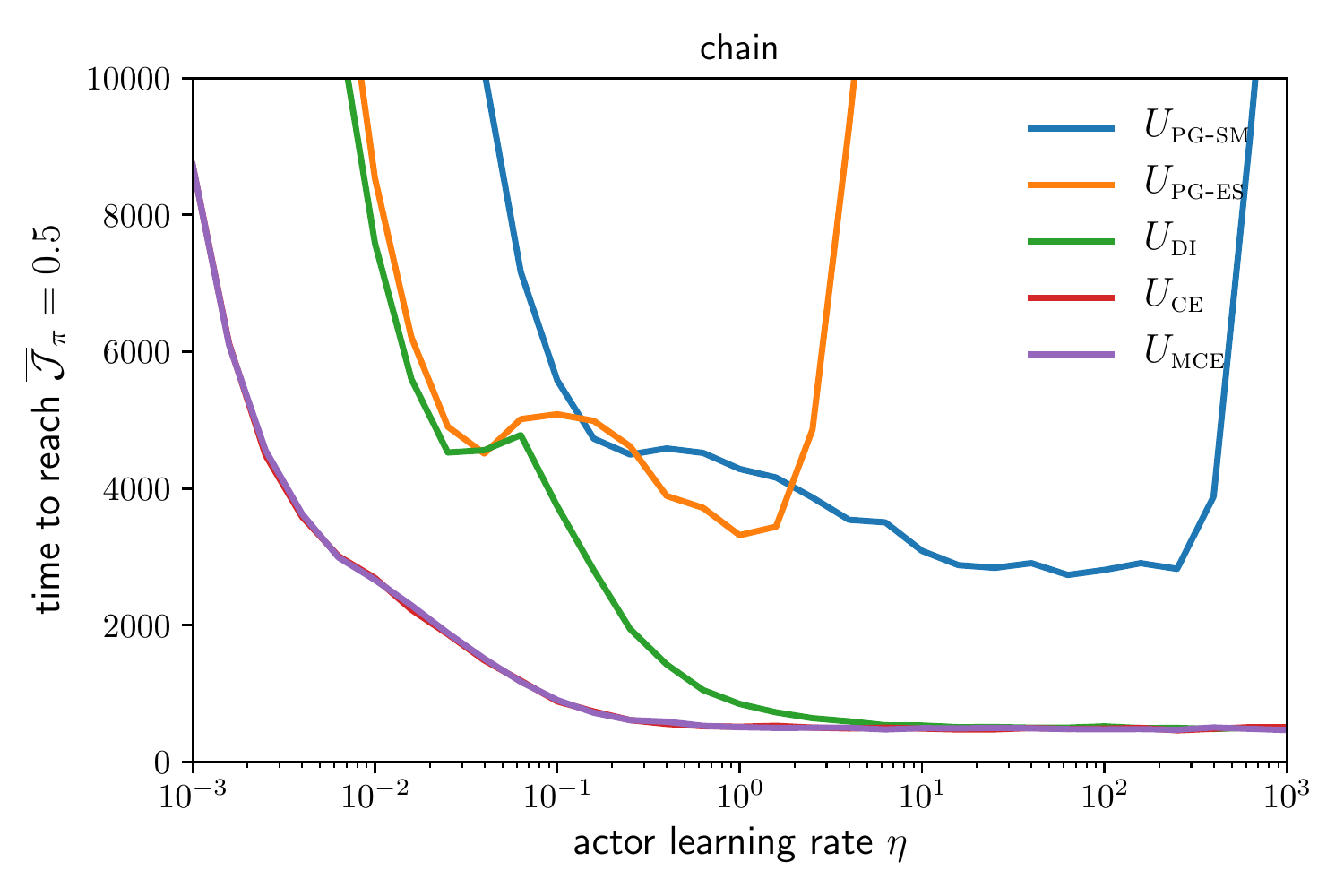}
		\label{fig:chainlow}
	}
	\subfloat[\textsc{HiOffPol}: $|\mathcal{S}|=10$ (100 runs)]{
		\includegraphics[trim = 5pt 5pt 5pt 20pt, clip, width=0.32\textwidth]{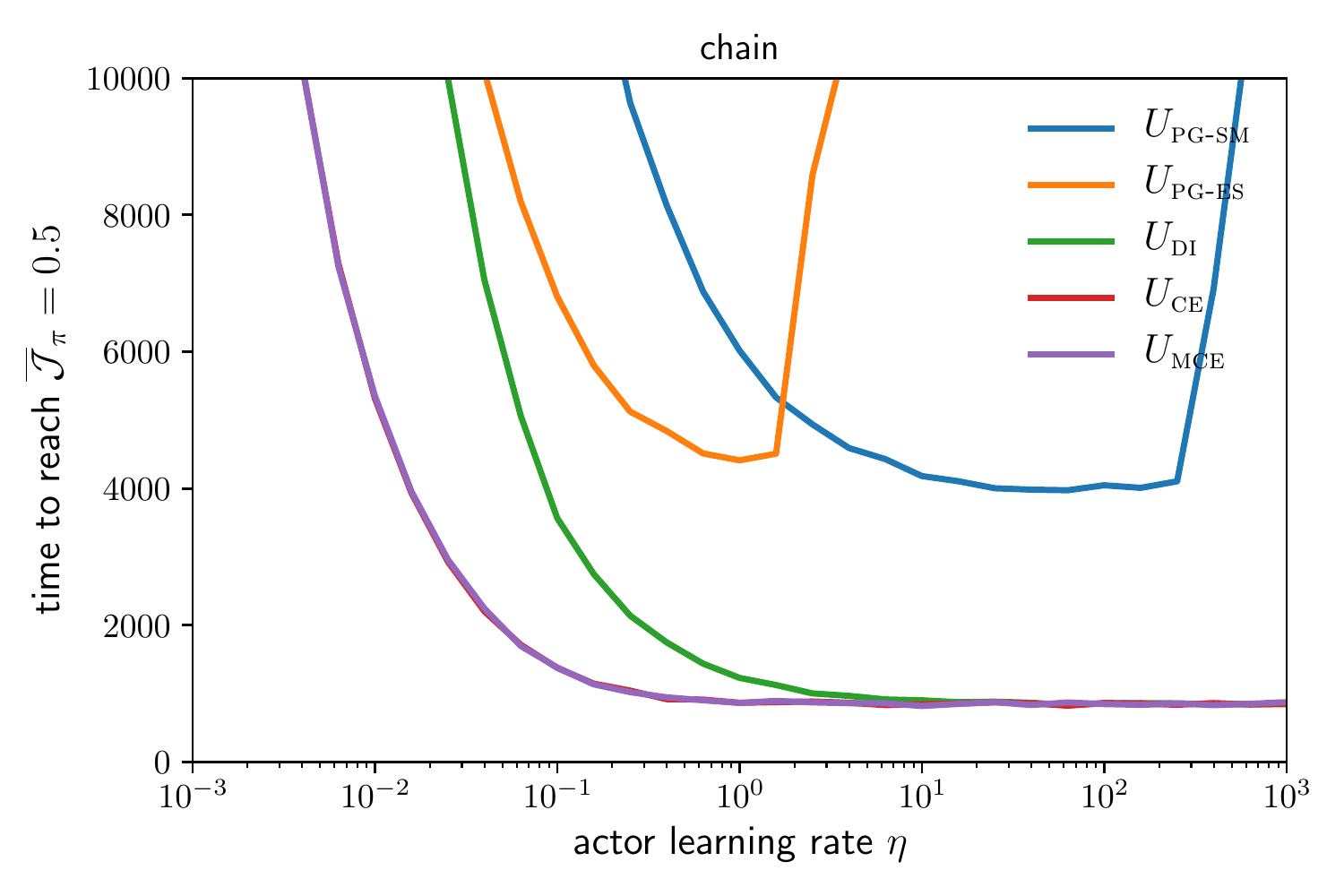}
		\label{fig:chainhi}
	}
	\caption{Chain: number of steps to obtain performance equal to 50\% of the gap between $\pi_\star$ and $\pi_s$.}
		\label{fig:chain}
	\vspace{-10pt}
\end{figure*}
\begin{figure*}[t!]
	\centering
	\subfloat[\textsc{LowOffPol}: ($\approx$20 runs)]{
		\includegraphics[trim = 5pt 5pt 5pt 20pt, clip, width=0.32\textwidth]{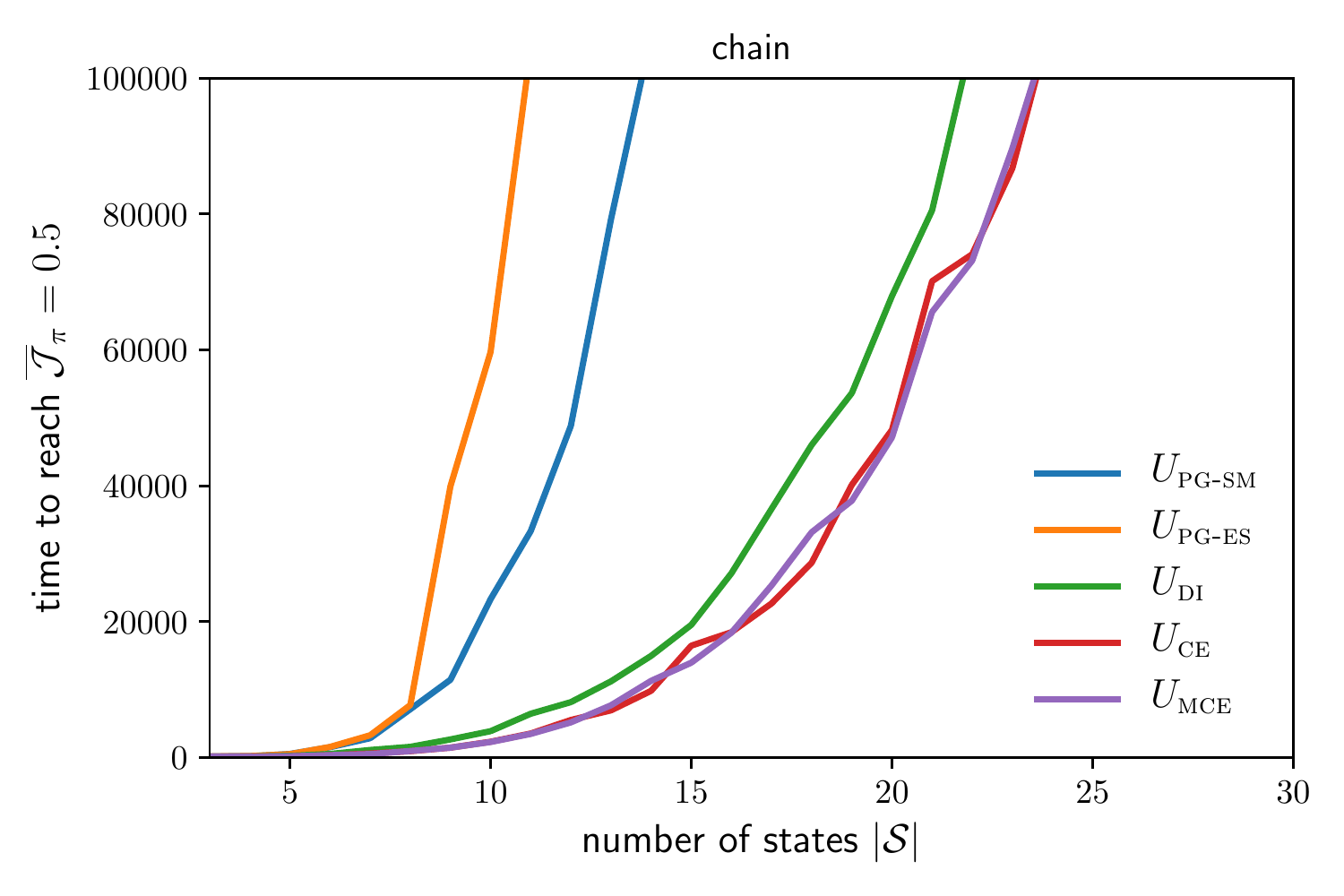}
		\label{fig:nbchainlow}
	}
	\subfloat[\textsc{HiOffPol}: ($\approx$20 runs)]{
		\includegraphics[trim = 5pt 5pt 5pt 20pt, clip, width=0.32\textwidth]{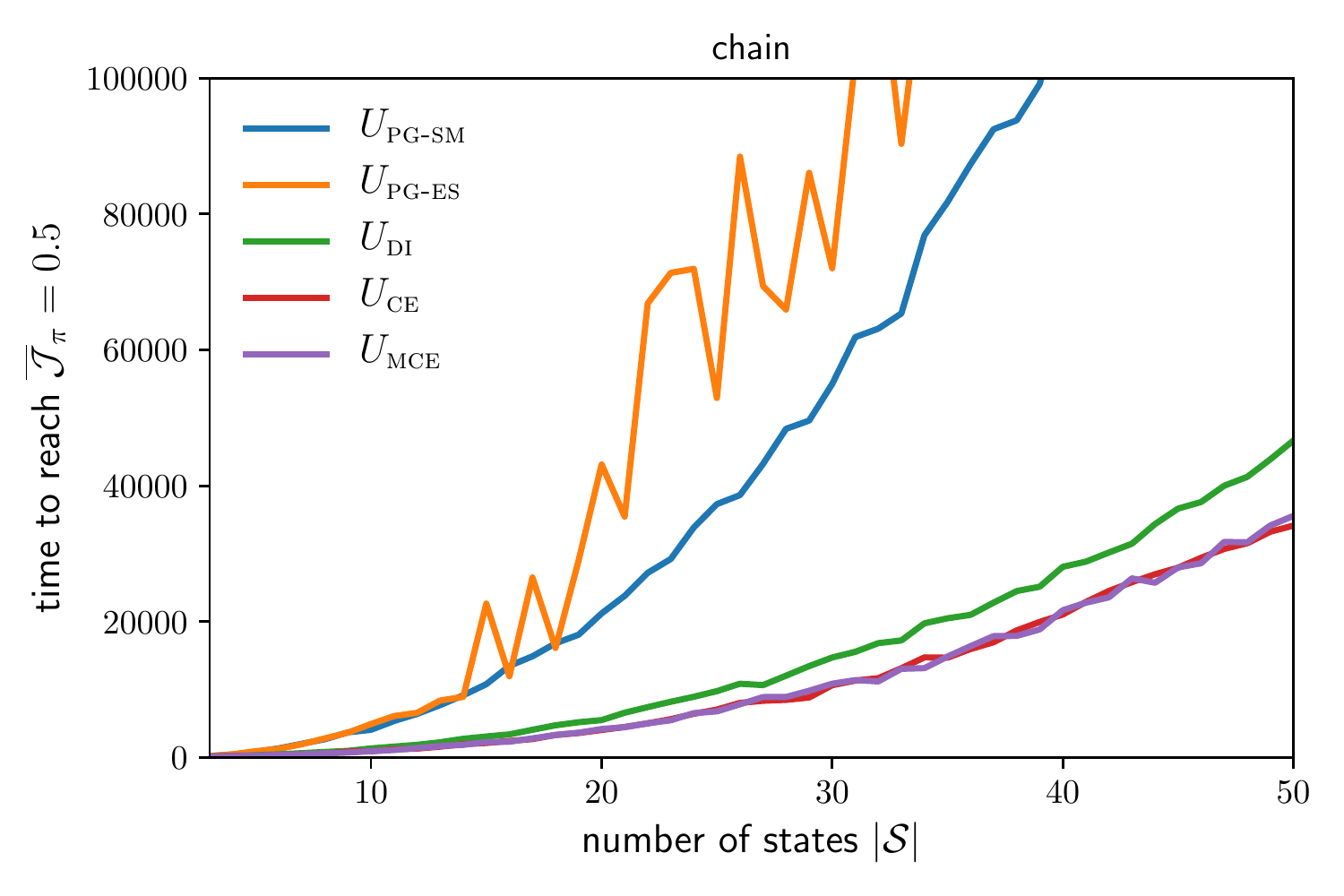}
		\label{fig:nbchainhi}
	}
	\subfloat[\textsc{HiOffPol}: with duplicate actions]{
		\includegraphics[trim = 5pt 5pt 5pt 20pt, clip, width=0.32\textwidth]{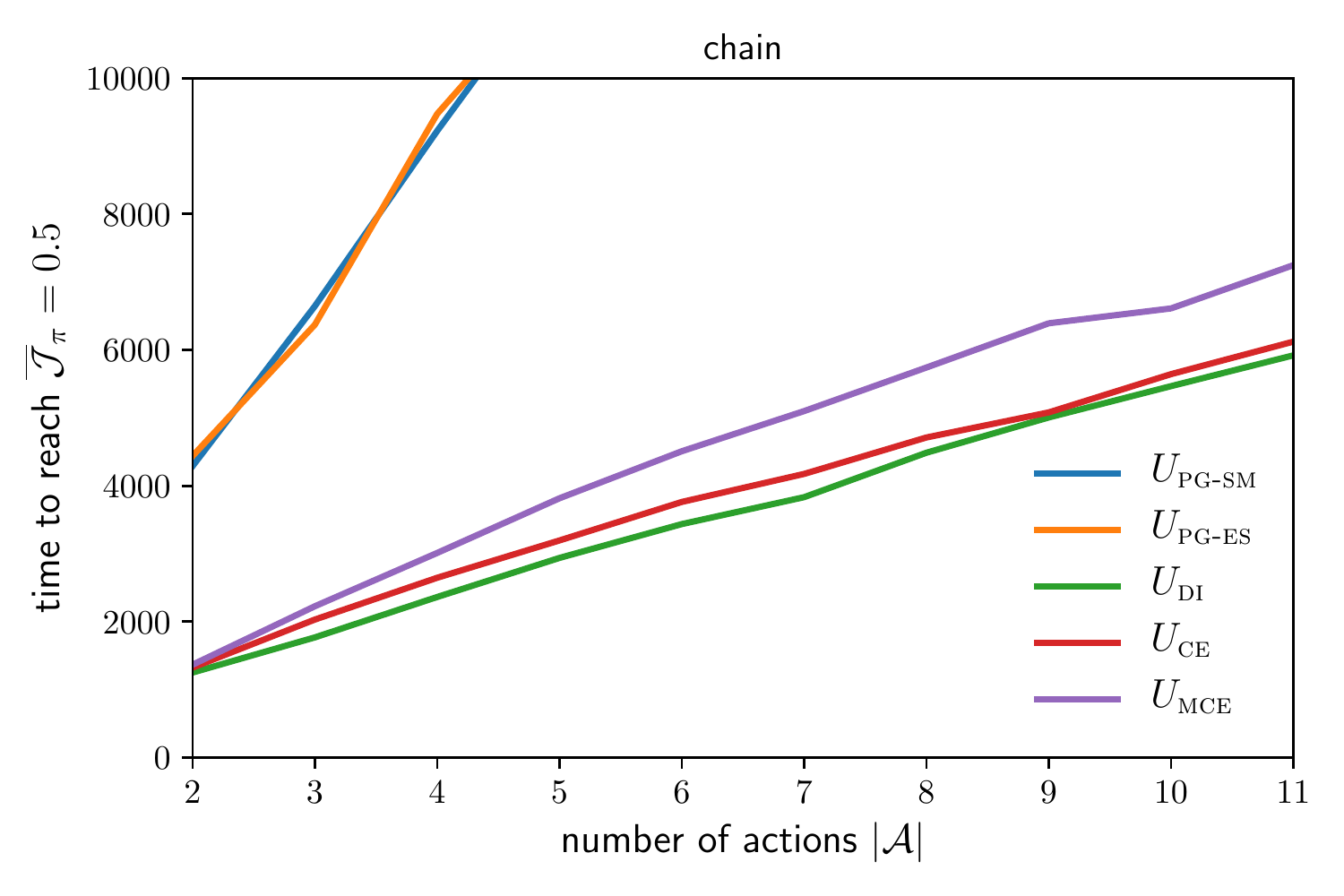}
		\label{fig:nbcliffhi}
	}
% 	\subfloat[\textsc{HiOffPol}: cliff $|\mathcal{S}|=7$ ($\approx$150 runs)]{
% 		\includegraphics[trim = 5pt 5pt 5pt 20pt, clip, width=0.32\textwidth]{figures/cliff/hioffpol/results.pdf}
% 		\label{fig:nbcliffhi}
% 	}
	\caption{Chain: number of steps to obtain performance equal to 50\% of the gap between $\pi_\star$ and $\pi_s$.}
		\label{fig:nbchain}
	\vspace{-10pt}
\end{figure*}

\begin{figure}[b!]
	\centering
	\includegraphics[trim = 5pt 5pt 5pt 20pt, clip, width=0.32\textwidth]{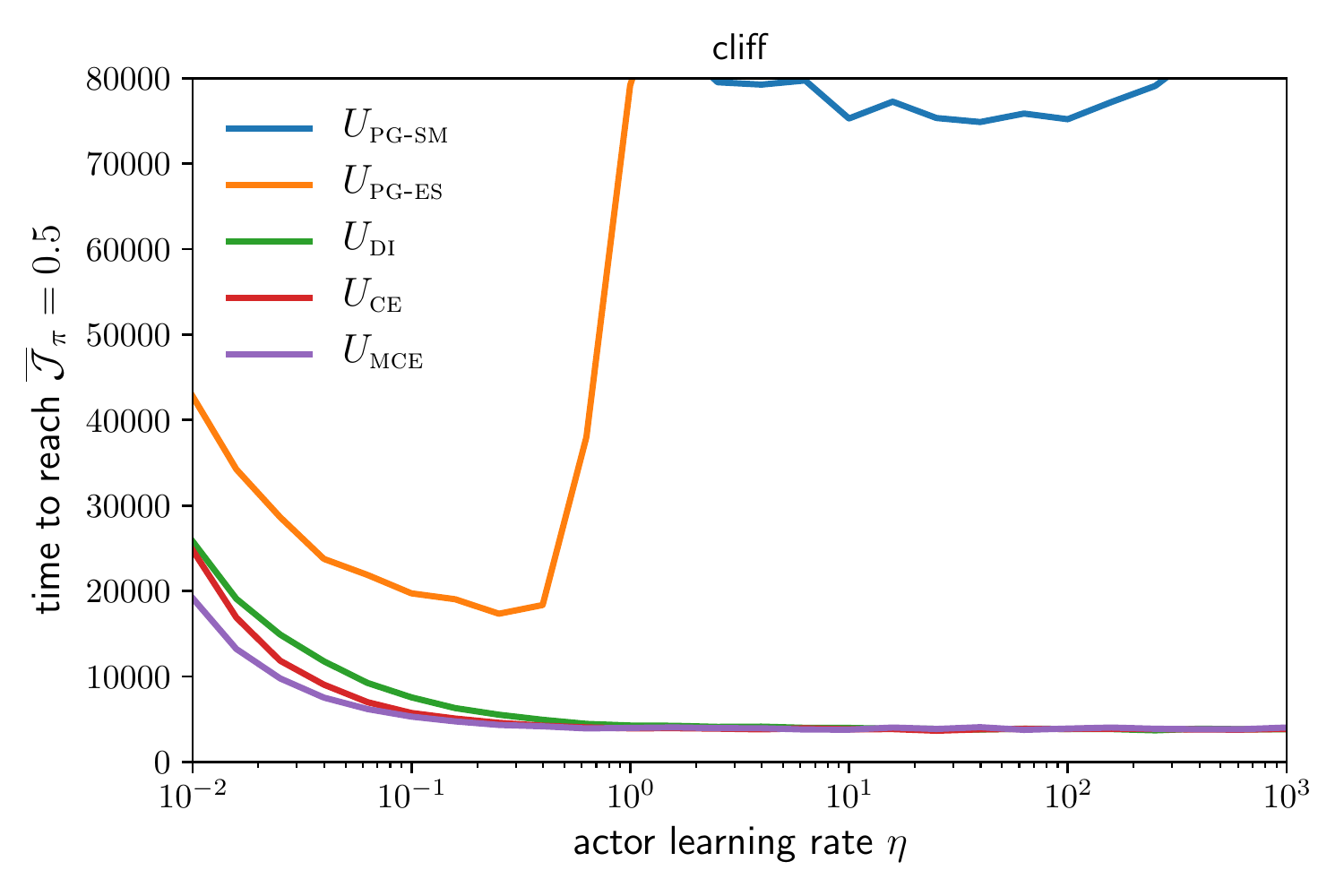}
	\caption{Cliff experiment \textsc{HiOffPol}: $|\mathcal{S}|=7$.}
		\label{fig:duplicates}
\end{figure}
% \begin{figure}[b!]
% 	\centering
% 	\includegraphics[trim = 5pt 5pt 5pt 20pt, clip, width=0.32\textwidth]{figures/rebuttal/duplicate-actions.pdf}
% 	\caption{Chain experiment with duplicated optimal actions. \textsc{HiOffPol} and $|\mathcal{S}|=10$.}
% 		\label{fig:duplicates}
% \end{figure}

\subsection{Chain experiment}
\label{sec:chainexp}
Framing the domino setting in an RL task would have been possible, but we opted for a less processed environment, already existing in the literature. The chain domain, depicted on Figure \ref{fig:chain-env}, implements a deterministic walk along a line, starting from $s_0$ to $s_{|\mathcal{S}|-1}$. In any state, it is possible to jump off the chain (action $a_1$) and get an immediate reward, but the optimal policy consists in walking (action $a_2$) throughout the chain. We expect the PUs to first guide the agent towards $a_1$, and require exploration and updates to propagate the optimal policy $\pi_\star(a_2|s)\approx 1$ and value through the chain. Thus, starting from the end of the chain, the policy in each state will need to be flipped in order to choose the best action at the first state.

We evaluate the policies with the number of steps (each step is a transition and an update) they need to reach a normalized performance $\overline{\mathcal{J}}_\pi$ that is equal to half the gap between that of the optimal policy $\pi_\star$ and that of the suboptimal one $\pi_{s}(a_1|\cdot)=1$, formally $\overline{\mathcal{J}}_\pi = \frac{\mathcal{J}(\pi)-\mathcal{J}(\pi_s)}{\mathcal{J}(\pi_\star) - \mathcal{J}(\pi_{s})}$.

Similarly to the Random MDPs, we first look at the incidence of the actor learning rate for each update rule. The results are presented in Figure \ref{fig:chain}. First, we notice that the size of the chain had to be adapted to each setting. Indeed, with no exploration and no off-policy updates, policy updates tends to completely converge to the suboptimal policy and stop updating the subsequent states much more easily. In all the settings, the cross-entropy updates $U_\textsc{ce}$ and $U_\textsc{mce}$ yield the best results even better than $U_\textsc{di}$. In \textsc{LowOffPol} (Figure \ref{fig:chainlow}) and \textsc{HiOffPol} (Figure \ref{fig:chainhi}), this might just be a hyperparameter shift, but in \textsc{NoExplo} (Figure \ref{fig:chainno}), the advantage is real, which was unexpected to us since it unlearns a bit more slowly in theory. This might be due to the fact that it can both learn and unlearn fast but still explore. In contrast, either $U_\textsc{di}$ has a small learning rate, and it is slow to unlearn, or it has a large learning rate and it completely stops exploring too fast. This analysis seems confirmed by the fact that $U_\textsc{di}$ yields a performance similar to that of $U_\textsc{pg-sm}$ and $U_\textsc{pg-es}$, suggesting it never gets the benefit of its projection's step ability to break the symmetry of the policy gradient.

We conduct an additional experiment where we observe how each update behaves when the length of the chain grows (Figures \ref{fig:nbchainlow} and \ref{fig:nbchainhi}). We set their best learning rate:
% $\eta_\textsc{pg-sm}=0.05,\eta_\textsc{pg-es}=0.01,\eta_\textsc{di}=0.01,\eta_\textsc{ce}=0.005,$ and $\eta_\textsc{mce}=0.005$ for \textsc{NoExplo}, and
$\eta_\textsc{pg-sm}=10,\eta_\textsc{pg-es}=1,\eta_\textsc{di}=1,\eta_\textsc{ce}=1,$ and $\eta_\textsc{mce}=1$ for \textsc{LowOffPol} and \textsc{HiOffPol}. We observe that  $U_\textsc{pg-sm}$ and $U_\textsc{pg-es}$ have approximately the same behaviour and are much slower to converge to the optimal policy. $U_\textsc{pg-es}$ has more visible variance because it diverges sometimes (and less runs have been performed for this experiment). With sufficient exploration, $U_\textsc{di}$, $U_\textsc{ce}$, and $U_\textsc{mce}$ have the same behaviour granted that their learning rate is set sufficiently high (setting $\eta_\textsc{di}=1$ was slightly low and this is the reason why $U_\textsc{di}$ is a bit slower.

To further investigate it, we run our chain environment ($|\mathcal{S}|=10$) where optimal actions have been duplicated $|\mathcal{A}|-1$ times. The results are displayed on Figure \ref{fig:duplicates}. The results show no convergence delay for $U_\textsc{ce}$, as compared with $U_\textsc{di}$, and a slight delay for $U_\textsc{mce}$. Overall, this experiment supports our conjecture that the uniqueness assumption is an artefact of our proof technique and not a structural flaw of the policy update.

\subsection{Cliff experiment}
The cliff experiment is similar to the chain except it includes a third action: $a_3$ jumps and leads to the abyss: $r(s,a_3)=0$ and terminates. It has been designed to create gravity wells: the policy will first converge to $a_1$, and will eventually be able to converge to optimal $a_2$ with enough exploration and updates. Like with the chain domain, we report the time to reach $\overline{\mathcal{J}}_\pi=0.5$.

The results are presented in Figure \ref{fig:nbcliffhi}. As expected, despite the cliff being rather short: $|\mathcal{S}|=7$, $U_\textsc{pg-sm}$ is very slow to converge to the optimal policy, which we interpret as a manifestation of its gravity well. $U_\textsc{pg-es}$ does significantly better but still hits a performance wall with high learning rates because of its divergence behaviour. With its best learning rate, $U_\textsc{pg-es}$ shows more or less the same relative performance gap with the best updates as in the chain domain. We thereby attribute the gap to its unlearning slowness rather than to a gravity well effect.

$U_\textsc{di}$, $U_\textsc{ce}$, and $U_\textsc{mce}$ display the best performance in a similar fashion than in the chain domain at the exception that $U_\textsc{ce}$ and $U_\textsc{mce}$ are not behaving exactly identically (but still similarly). Indeed, with small learning rates, $U_\textsc{mce}$ does a bit better than $U_\textsc{ce}$. We naturally attribute this to its ability to perform monotonic updates that should help to converge faster when starting from a bad parameter initialization, because of past convergence to the suboptimal solution.

\section{CONCLUSION}
In this paper, we identified the unlearning slowness implied by policy gradient updates in actor-critic algorithms. We proposed several alternatives to these updates: the direct parametrization, and two novel cross-entropy-based updates, including one for which we prove convergence to optimality at a rate of $\mathcal{O}(t^{-1})$, under classic assumptions. We further extend the analysis to the study of their ability to enter a gravity well, their unlearning speed, and various implementation constraints. Finally, we empirically validate our theoretical findings on finite MDPs.

In the future, we would like to extend convergence/optimality proofs of policy updates to the stochastic and approximate case, as this kind of results is starting to emerge for softmax policy gradient updates~\citep{zhang2021global,zhang2022chattering}. We would also like to study other non policy gradient policy updates, and find some that solve the both the policy gradient unlearning slowness and the stochasticity issue related to the cross-entropy updates. Finally, we would like to investigate the relationship of these policy updates with the function approximators used in complex tasks.

\bibliographystyle{apalike}
\bibliography{biblioReport}

\begin{thebibliography}{}

\bibitem[Agache and Oommen, 2002]{Agache2002}
Agache, M. and Oommen, B.~J. (2002).
\newblock Generalized pursuit learning schemes: new families of continuous and
  discretized learning automata.
\newblock {\em IEEE Trans. Syst. Man Cybern. Part B}, 32(6):738--749.

\bibitem[Agarwal et~al., 2020]{Agarwal2019}
Agarwal, A., Kakade, S.~M., Lee, J.~D., and Mahajan, G. (2020).
\newblock Optimality and approximation with policy gradient methods in markov
  decision processes.
\newblock In Abernethy, J. and Agarwal, S., editors, {\em Proceedings of 3rd
  Conference on Learning Theory (COLT)}, volume 125 of {\em Proceedings of
  Machine Learning Research}, pages 64--66. PMLR.

\bibitem[Ahmed et~al., 2019]{Ahmed2019}
Ahmed, Z., Le~Roux, N., Norouzi, M., and Schuurmans, D. (2019).
\newblock Understanding the impact of entropy on policy optimization.
\newblock In {\em Proceedings of the 37th International Conference on Machine
  Learning (ICML)}, pages 151--160. PMLR.

\bibitem[Ciosek and Whiteson, 2018]{Ciosek2018}
Ciosek, K. and Whiteson, S. (2018).
\newblock Expected policy gradients.
\newblock In {\em Proceedings of the 32nd AAAI Conference on Artificial
  Intelligence}, volume~32.

\bibitem[Espeholt et~al., 2018]{espeholt2018impala}
Espeholt, L., Soyer, H., Munos, R., Simonyan, K., Mnih, V., Ward, T., Doron,
  Y., Firoiu, V., Harley, T., Dunning, I., Legg, S., and Kavukcuoglu, K.
  (2018).
\newblock {IMPALA:} scalable distributed deep-rl with importance weighted
  actor-learner architectures.
\newblock In {\em Proceedings of the 35th International Conference on Machine
  Learning (ICML)}.

\bibitem[Kakade and Langford, 2002]{Kakade2002}
Kakade, S.~M. and Langford, J. (2002).
\newblock Approximately optimal approximate reinforcement learning.
\newblock In Sammut, C. and Hoffmann, A.~G., editors, {\em ICML}, pages
  267--274. Morgan Kaufmann.

\bibitem[Konda and Tsitsiklis, 1999]{konda2000actor}
Konda, V.~R. and Tsitsiklis, J.~N. (1999).
\newblock Actor-critic algorithms.
\newblock In {\em Proceedings of the 12th Advances in Neural Information
  Processing Systems (NeurIPS)}.

\bibitem[Kumar et~al., 2019]{kumar2019sample}
Kumar, H., Koppel, A., and Ribeiro, A. (2019).
\newblock On the sample complexity of actor-critic method for reinforcement
  learning with function approximation.
\newblock {\em arXiv preprint arXiv:1910.08412}.

\bibitem[Laha et~al., 2018]{Laha2018}
Laha, A., Chemmengath, S.~A., Agrawal, P., Khapra, M., Sankaranarayanan, K.,
  and Ramaswamy, H.~G. (2018).
\newblock On controllable sparse alternatives to softmax.
\newblock In Bengio, S., Wallach, H., Larochelle, H., Grauman, K.,
  Cesa-Bianchi, N., and Garnett, R., editors, {\em Proceedings of the 31st
  Advances in Neural Information Processing Systems (NeurIPS)}, volume~31.
  Curran Associates, Inc.

\bibitem[Laroche and Tachet~des Combes, 2021]{Laroche2021}
Laroche, R. and Tachet~des Combes, R. (2021).
\newblock Dr jekyll and mr hyde: The strange case of off-policy policy updates.
\newblock In {\em Proceedings of the 34th Advances in Neural Information
  Processing Systems (NeurIPS, to appear)}.

\bibitem[Laroche et~al., 2019]{Laroche2019}
Laroche, R., Trichelair, P., and Tachet~des Combes, R. (2019).
\newblock Safe policy improvement with baseline bootstrapping.
\newblock In {\em Proceedings of the 36th International Conference on Machine
  Learning (ICML)}.

\bibitem[Lillicrap et~al., 2016]{lillicrap2015continuous}
Lillicrap, T.~P., Hunt, J.~J., Pritzel, A., Heess, N., Erez, T., Tassa, Y.,
  Silver, D., and Wierstra, D. (2016).
\newblock Continuous control with deep reinforcement learning.
\newblock In {\em Proceedings of the 4th International Conference on Learning
  Representations (ICLR, poster)}.

\bibitem[Martins and Astudillo, 2016]{Martins2016}
Martins, A. and Astudillo, R. (2016).
\newblock From softmax to sparsemax: A sparse model of attention and
  multi-label classification.
\newblock In {\em Proceedings of the 33rd International Conference on Machine
  Learning (ICML)}, pages 1614--1623. PMLR.

\bibitem[Mei et~al., 2020a]{Mei2020b}
Mei, J., Xiao, C., Dai, B., Li, L., Szepesvari, C., and Schuurmans, D. (2020a).
\newblock Escaping the gravitational pull of softmax.
\newblock In Larochelle, H., Ranzato, M., Hadsell, R., Balcan, M.~F., and Lin,
  H., editors, {\em Proceedings of the 33rd Advances in Neural Information
  Processing Systems (NeurIPS)}, volume~33, pages 21130--21140. Curran
  Associates, Inc.

\bibitem[Mei et~al., 2020b]{Mei2020a}
Mei, J., Xiao, C., Szepesvari, C., and Schuurmans, D. (2020b).
\newblock On the global convergence rates of softmax policy gradient methods.
\newblock In {\em Proceedings of the 37th International Conference on Machine
  Learning (ICML)}, pages 6820--6829. PMLR.

\bibitem[Mnih et~al., 2016]{mnih2016asynchronous}
Mnih, V., Badia, A.~P., Mirza, M., Graves, A., Lillicrap, T.~P., Harley, T.,
  Silver, D., and Kavukcuoglu, K. (2016).
\newblock Asynchronous methods for deep reinforcement learning.
\newblock In {\em Proceedings of the 33rd International Conference on Machine
  Learning (ICML)}.

\bibitem[Nadjahi* et~al., 2019]{Nadjahi2019}
Nadjahi*, K., Laroche*, R., and Tachet~des Combes, R. (2019).
\newblock Safe policy improvement with soft baseline bootstrapping.
\newblock In {\em Proceedings of the 17th European Conference on Machine
  Learning and Principles and Practice of Knowledge Discovery in Databases
  (ECML-PKDD)}.

\bibitem[Nota and Thomas, 2020]{Nota2020}
Nota, C. and Thomas, P.~S. (2020).
\newblock Is the policy gradient a gradient?
\newblock In {\em Proceedings of the 19th International Conference on
  Autonomous Agents and Multiagent Systems (AAMAS)}.

\bibitem[Parisotto et~al., 2016]{parisotto2016actormimic}
Parisotto, E., Ba, J.~L., and Salakhutdinov, R. (2016).
\newblock Actor-mimic: Deep multitask and transfer reinforcement learning.

\bibitem[Pirotta et~al., 2013]{Pirotta2013}
Pirotta, M., Restelli, M., Pecorino, A., and Calandriello, D. (2013).
\newblock Safe policy iteration.
\newblock In {\em ICML (3)}, volume~28 of {\em JMLR Workshop and Conference
  Proceedings}, pages 307--315. JMLR.org.

\bibitem[Qiu et~al., 2021]{qiu2021finite}
Qiu, S., Yang, Z., Ye, J., and Wang, Z. (2021).
\newblock On finite-time convergence of actor-critic algorithm.
\newblock {\em IEEE Journal on Selected Areas in Information Theory},
  2(2):652--664.

\bibitem[Scherrer, 2014]{scherrer2014approximate}
Scherrer, B. (2014).
\newblock Approximate policy iteration schemes: A comparison.

\bibitem[Schmitt et~al., 2020]{schmitt2020off}
Schmitt, S., Hessel, M., and Simonyan, K. (2020).
\newblock Off-policy actor-critic with shared experience replay.
\newblock In {\em Proceedings of the 37th International Conference on Machine
  Learning (ICML)}, pages 8545--8554. PMLR.

\bibitem[Silver et~al., 2016]{silver2016mastering}
Silver, D., Huang, A., Maddison, C.~J., Guez, A., Sifre, L., van~den Driessche,
  G., Schrittwieser, J., Antonoglou, I., Panneershelvam, V., Lanctot, M.,
  Dieleman, S., Grewe, D., Nham, J., Kalchbrenner, N., Sutskever, I.,
  Lillicrap, T.~P., Leach, M., Kavukcuoglu, K., Graepel, T., and Hassabis, D.
  (2016).
\newblock Mastering the game of go with deep neural networks and tree search.
\newblock {\em Nature}.

\bibitem[Silver et~al., 2014]{Silver2014}
Silver, D., Lever, G., Heess, N., Degris, T., Wierstra, D., and Riedmiller, M.
  (2014).
\newblock Deterministic policy gradient algorithms.
\newblock In {\em Proceedings of the 31st International Conference on Machine
  Learning (ICML)}, pages 387--395. PMLR.

\bibitem[Sim\~ao et~al., 2020]{Simao2020}
Sim\~ao, T.~D., Laroche, R., and Tachet~des Combes, R. (2020).
\newblock Safe policy improvement with estimated baseline bootstrapping.
\newblock In {\em Proceedings of the 19th International Conference on
  Autonomous Agents and Multi-Agent Systems (AAMAS, in review)}.

\bibitem[Soudry et~al., 2018]{soudry2018implicit}
Soudry, D., Hoffer, E., Nacson, M.~S., Gunasekar, S., and Srebro, N. (2018).
\newblock The implicit bias of gradient descent on separable data.
\newblock {\em Journal of Machine Learning Research}, 19(70):1--57.

\bibitem[Sutton and Barto, 1998]{Sutton1998}
Sutton, R.~S. and Barto, A.~G. (1998).
\newblock {\em Reinforcement Learning: An Introduction}.
\newblock The MIT Press.

\bibitem[Sutton et~al., 1999]{sutton2000policy}
Sutton, R.~S., McAllester, D.~A., Singh, S.~P., and Mansour, Y. (1999).
\newblock Policy gradient methods for reinforcement learning with function
  approximation.
\newblock In {\em Proceedings of the 12th Advances in Neural Information
  Processing Systems (NeurIPS)}.

\bibitem[Thathachar and Sastry, 1986]{Thathachar1986EstimatorAF}
Thathachar, M. A.~L. and Sastry, P.~S. (1986).
\newblock Estimator algorithms for learning automata.

\bibitem[Thomas, 2014]{Thomas2014}
Thomas, P. (2014).
\newblock Bias in natural actor-critic algorithms.
\newblock In {\em Proceedings of the 31st International Conference on Machine
  Learning (ICML)}.

\bibitem[Vinyals et~al., 2019]{vinyals2019grandmaster}
Vinyals, O., Babuschkin, I., Czarnecki, W.~M., Mathieu, M., Dudzik, A., Chung,
  J., Choi, D.~H., Powell, R., Ewalds, T., Georgiev, P., Oh, J., Horgan, D.,
  Kroiss, M., Danihelka, I., Huang, A., Sifre, L., Cai, T., Agapiou, J.~P.,
  Jaderberg, M., Vezhnevets, A.~S., Leblond, R., Pohlen, T., Dalibard, V.,
  Budden, D., Sulsky, Y., Molloy, J., Paine, T.~L., G{\"{u}}l{\c{c}}ehre,
  {\c{C}}., Wang, Z., Pfaff, T., Wu, Y., Ring, R., Yogatama, D., W{\"{u}}nsch,
  D., McKinney, K., Smith, O., Schaul, T., Lillicrap, T.~P., Kavukcuoglu, K.,
  Hassabis, D., Apps, C., and Silver, D. (2019).
\newblock Grandmaster level in starcraft {II} using multi-agent reinforcement
  learning.
\newblock {\em Nature}.

\bibitem[Wang et~al., 2017]{wang2016sample}
Wang, Z., Bapst, V., Heess, N., Mnih, V., Munos, R., Kavukcuoglu, K., and
  de~Freitas, N. (2017).
\newblock Sample efficient actor-critic with experience replay.
\newblock In {\em Proceedings of the 5th International Conference on Learning
  Representations (ICLR)}.

\bibitem[Williams, 1992]{Williams1992}
Williams, R.~J. (1992).
\newblock Simple statistical gradient-following algorithms for connectionist
  reinforcement learning.
\newblock {\em Machine Learning}, 8(3–4):229–256.

\bibitem[Zahavy et~al., 2020]{zahavy2020self}
Zahavy, T., Xu, Z., Veeriah, V., Hessel, M., Oh, J., van Hasselt, H.~P.,
  Silver, D., and Singh, S. (2020).
\newblock A self-tuning actor-critic algorithm.
\newblock In Larochelle, H., Ranzato, M., Hadsell, R., Balcan, M.~F., and Lin,
  H., editors, {\em Proceedings of the 33rd Advances in Neural Information
  Processing Systems (NeurIPS)}, volume~33, pages 20913--20924. Curran
  Associates, Inc.

\bibitem[Zhang et~al., 2020a]{zhang2020global}
Zhang, K., Koppel, A., Zhu, H., and Basar, T. (2020a).
\newblock Global convergence of policy gradient methods to (almost) locally
  optimal policies.
\newblock {\em SIAM Journal on Control and Optimization}, 58(6):3586--3612.

\bibitem[Zhang et~al., 2020b]{Zhang2020deeper}
Zhang, S., Laroche, R., van Seijen, H., Whiteson, S., and Tachet~des Combes, R.
  (2020b).
\newblock A deeper look at discounting mismatch in actor-critic algorithms.
\newblock {\em arXiv preprint arXiv:2010.01069}.

\bibitem[Zhang et~al., 2022]{zhang2022chattering}
Zhang, S., Tachet, R., and Laroche, R. (2022).
\newblock On the chattering of sarsa with linear function approximation.

\bibitem[Zhang et~al., 2021]{zhang2021global}
Zhang, S., Tachet~des Combes, R., and Laroche, R. (2021).
\newblock Global optimality and finite sample analysis of softmax off-policy
  actor critic under state distribution mismatch.
\newblock {\em arXiv preprint arXiv:2111.02997}.

\end{thebibliography}

\clearpage

\appendix
\onecolumn

\section{THEORY}\label{app:theory}

\subsection{Notations and preliminaries}\label{app:prelim}

\subsubsection{Softmax parametrization: $U_\textsc{pg-sm}$}

For completeness, we reproduce the computations for the softmax parametrization present in e.g. ~\cite{Agarwal2019}:
\begin{align}
    \nabla_\theta \pi_\theta(a|s) &= \nabla_\theta \left(\frac{e^{f_\theta(s,a)}}{\sum_{a'} e^{f_\theta(s,a)}}\right) \\
    &= \nabla_\theta f_\theta(s,a) \frac{e^{f_\theta(s,a)}}{\sum_{a'} e^{f_\theta(s,a)}} - \sum_{a''} e^{f_\theta(s,a)} \nabla_\theta f_\theta(s,a'') \frac{e^{f_\theta(s,a'')}}{\left(\sum_{a'} e^{f_\theta(s,a')}\right)^2} \\
    &= \pi_\theta(s,a) \left[\nabla_\theta f_\theta(s,a) - \sum_{a''} \nabla_\theta f_\theta(s,a'') \pi_\theta(s,a'')\right].
\end{align}

This leads to the following update step for $U_\textsc{pg-sm}$ with a tabular parametrization: $\theta \in \mathbb{R}^\mathcal{A}$ and $f_\theta(s,a) = \theta_a$. In that case, $(\nabla_\theta f_\theta(s,a))_a = 1$ and its other coordinates are 0. This gives:
\begin{align}
(U_\textsc{pg-sm}(\theta))_{s,a}-(\theta)_{s,a} &\propto \sum_{a'\in\mathcal{A}} q(s, a') \nabla_{\theta} \pi_\theta(a'|s) \\ 
&= \pi_\theta(s,a) q(s,a) - \sum_{a'} \pi_\theta(s,a') q(s,a') \\
&= \pi_\theta(s,a) \textnormal{adv}(s, a).
\end{align}

\subsubsection{Escort parametrization: $U_\textsc{pg-es}$}

Let us compute the policy gradients for the escort parametrization, assuming for the sake of simplicity that for all $a$, $f_\theta(s,a) \geq 0$ (accounting for the sign simply amounts to multiplying any gradient $\nabla_\theta f_\theta(s,a)$ by $\text{sign} f_\theta(s,a)$):
\begin{align}
    \nabla_\theta \pi_\theta(a|s) &= \nabla_\theta \left(\frac{|f_\theta(s,a)|^p}{\sum_{a'} |f_\theta(s,a')|^p}\right) \\
    &= p \nabla_\theta f_\theta(s,a) \frac{|f_\theta(s,a)|^{p-1}}{\sum_{a'} |f_\theta(s,a')|^p} - \sum_{a''} |f_\theta(s,a)|^p p \nabla_\theta f_\theta(s,a'') \frac{|f_\theta(s,a'')|^{p-1}}{\left(\sum_{a'} |f_\theta(s,a')|^p\right)^2} \\
    &= p \nabla_\theta f_\theta(s,a) \frac{\pi_\theta(s,a)^{1-\frac{1}{p}}}{\left(\sum_{a'} |f_\theta(s,a')|^p\right)^{1 / p}} - \pi_\theta(s,a) \sum_{a''} p \nabla_\theta f_\theta(s,a'') \frac{|f_\theta(s,a'')|^{p-1}}{\sum_{a'} |f_\theta(s,a')|^p} \\
    &= \frac{p}{\|f_\theta(s,\cdot)\|_p} \pi_\theta(s,a)^{1-\frac{1}{p}} \left[ \nabla_\theta f_\theta(s,a) - \sum_{a''} \nabla_\theta f_\theta(s,a'') \pi_\theta(s,a)^{1 / p} \pi_\theta(s,a'')^{1-\frac{1}{p}} \right].
\end{align}

We now compute the update step for the escort update $U_\textsc{pg-es}$ with a tabular parametrization: $\theta \in \mathbb{R}^\mathcal{A}$ and $f_\theta(s,a) = \theta_a$. In that case, $(\nabla_\theta f_\theta(s,a))_a = 1$ and its other coordinates are 0. This gives:
\begin{align}
(U_\textsc{pg-es}(\theta))_{s,a}-(\theta)_{s,a} &\propto \sum_{a'\in\mathcal{A}} q(s, a') \nabla_{\theta} \pi_\theta(a'|s) \\ 
&= \frac{p}{\|f_\theta(s,\cdot)\|_p} \left[ \pi_\theta(s,a)^{1-\frac{1}{p}} (1 - \pi_\theta(s,a)) q(s,a) - \sum_{a' \neq a} \pi_\theta(s,a') \pi_\theta(s,a)^{1-\frac{1}{p}} q(s,a')\right] \\
&= \frac{p}{\|f_\theta(s,\cdot)\|_p} \pi_\theta(s,a)^{1-\frac{1}{p}} \left[  q(s,a) - \sum_{a'} \pi_\theta(s,a') q(s,a')\right] \\
&= \frac{p}{\|f_\theta(s,\cdot)\|_p} \pi_\theta(s,a)^{1-\frac{1}{p}} \textnormal{adv}(s, a).
\end{align}

% \subsubsection{Cross-entropy update: $U_\textsc{ce}$}

% \subsubsection{Modified cross-entropy update: $U_\textsc{mce}$}

\subsection{Convergence, optimality, and rates proofs (Sections \ref{sec:updates}, \ref{sec:conv}, and \ref{sec:asymp})}
\label{app:conv}
\nonmonotonicity*
\begin{proof}
    Consider an MDP with 3 actions $\mathcal{A}=\{a_1, a_2, a_3\}$. Assume the parameters in state $s$ are:
    \begin{align*}
        \theta_{s,a_1} = 10 \quad\quad \theta_{s,a_2} = 0 \quad\quad \theta_{s,a_3} = 0,
    \end{align*}
    and the value function is:
    \begin{align*}
        q(s,a_1) = 0.99 \quad\quad q(s,a_2) = 1 \quad\quad q(s,a_3) = 0.
    \end{align*}
    Let us choose $\eta=d(s)=1$ for simplicity. Then, update $U_\textsc{ce}$, defined as $\theta'_{s,a} \doteq \theta_{s,a} + \mathds{1}_{a=a_q(s)} - \pi_{\theta_t}\left(a|s\right)$, leads to:
    \begin{align*}
        \theta'_{s,a_1} \approx 10-1 = 9 \quad\quad \theta'_{s,a_2} \approx 0+1=1 \quad\quad \theta'_{s,a_3} \approx 0-0 = 0,
    \end{align*}
    meaning that $\pi_{\theta'}(a_2|s) \approx e^2 \pi_{\theta}(a_3|s)$, while $\pi_{\theta'}(a_3|s) \approx e \pi_{\theta}(a_3|s)$, inducing a disadvantageous update, given how $q(s,a_2)$ is closer to $q(s,a_1)$ than $q(s,a_3)$ is.
    Numerically we find that:
    \begin{align*}
        v_{\pi_\theta}(s) \approx 0.98996 > 0.98988 \approx v_{\pi_{\theta'}}(s),
    \end{align*}
    which confirms the result.
\end{proof}

\monotonicity*
\begin{proof}
    Let us fix any $s \in \mathcal{S}$. We recall that
    \begin{align}
        (\theta_{t+1})_{s,a_q(s)} &\doteq (\theta_{t})_{s,a_q(s)} + \eta_t d_t(s) \nabla_{\theta} \log \pi_{\theta_t}\left(a_q(s)|s\right) \\
        &= (\theta_{t})_{s,a_q(s)} + \eta_t d_t(s) \left(1- \pi_{\theta_t}\left(a_q(s)|s\right)\right) \\
        \forall a\neq a_q(s),\quad (\theta_{t+1})_{s,a} &\doteq (\theta_{t})_{s,a} - \frac{1}{|\mathcal{A}|-1}\eta_t d_t(s) \left(1- \pi_{\theta_t}\left(a_q(s)|s\right)\right) \\
        \text{with}\quad &\left\{\begin{array}{l}
          \pi_{\theta_t}(a|s) =  \displaystyle\frac{\exp((\theta_t)_{s,a})}{\sum_{a'\in\mathcal{A}}\exp((\theta_t)_{s,a'})} \\
          q_t(s,a) = \mathbb{E}\left[\sum_{t=0}^\infty \gamma^t R_t\bigg|\begin{array}{cc}
          S_0 = s, A_0=a, S_{t+1}\sim p(\cdot|S_t,A_t), \\
          A_t\sim \pi_{\theta_t}(\cdot|S_t), R_t\sim r(\cdot|S_t,A_t)
        \end{array} \right]\\
          a_q(s) = \argmax_{a\in\mathcal{A}} q_t(s,a).
        \end{array}\right.
    \end{align}
    
    Let $u_t \doteq \eta_t d_t(s) \left(1- \pi_{\theta_t}\left(a_q(s)|s\right)\right)$. For all $a\neq a_q(s)$:
    \begin{align}
        \pi_{t+1}(a|s) &= \frac{\exp((\theta_{t+1})_{s,a})}{\sum_{a'\in\mathcal{A}}\exp((\theta_{t+1})_{s,a'})} \\
        &= \frac{\exp((\theta_{t})_{s,a}-\frac{u_t}{|\mathcal{A}|-1})}{\exp((\theta_{t})_{s,a_q(s)}+u_t) + \sum_{a'\neq a_q(s)}\exp((\theta_{t})_{s,a'}-\frac{u_t}{|\mathcal{A}|-1})} \\
        &= \frac{\exp((\theta_{t})_{s,a})}{\exp\left((\theta_{t})_{s,a_q(s)}+u_t+\frac{u_t}{|\mathcal{A}|-1}\right) + \sum_{a'\neq a_q(s)}\exp((\theta_{t})_{s,a'})}  \label{eq:ut}  \\
        &\leq \frac{\exp((\theta_{t})_{s,a})}{\exp((\theta_{t})_{s,a_q(s)}) + \sum_{a'\neq a_q(s)}\exp((\theta_{t})_{s,a'})}    \\
        &=  \pi_{t}(a|s). \label{eq:monotonic-decrease-policy}
    \end{align}
    
    We prove now that the new policy is advantageous by rearranging the policy masses:
    \begin{align}
        \textnormal{adv}_t(s, \pi_{t+1}) &\doteq \sum_{a\in\mathcal{A}}(\pi_{t+1}(a|s)-\pi_{t}(a|s))q_t(s, a) \\
        &= (\pi_{t+1}(a_q(s)|s)-\pi_{t}(a_q(s)|s))q_t(s, a_q(s))+\sum_{a\neq a_q(s)}(\pi_{t+1}(a|s)-\pi_{t}(a|s))q_t(s, a) \\
        &= \sum_{a\neq a_q(s)}\underbrace{(\pi_{t+1}(a|s)-\pi_{t}(a|s))}_{\leq 0 \text{ from Eq. \eqref{eq:monotonic-decrease-policy}}}\underbrace{(q_t(s, a) - q_t(s, a_q(s)))}_{\leq 0 \text{ from } a_q(s) \text{'s optimality}} \label{eq:advantage}\\
        &\geq 0.
    \end{align}
    which is true for all $s$ and therefore allows to apply the policy improvement theorem to conclude the proof.
\end{proof}

\optimality*
\begin{proof}
    We use the same structure of proof as \cite{Laroche2021}. Monotonicy of value functions has been proven in Proposition \ref{prop:monotonicity}. Next, we prove the convergence of the value functions $q_t$.
     
    \begin{corollary}[\textbf{Convergence under $U_\textsc{mce}$}]
        The sequence of values $(q_t)$ converges to some value function: $q_\infty \doteq \lim_{t\to\infty} q_t$.
        \label{cor:convergence}
    \end{corollary}
    \begin{proof}
        By the monotonicity and boundedness of the state value functions, the monotone convergence theorem guarantees the sequence $(v_t)$ converges to $v_\infty \doteq \lim_{t\to\infty} v_t$.
        Applying Bellman's equation then proves the existence of $q_\infty \in \mathbb{R}^{|\mathcal{S}|\times|\mathcal{A}|}$ that is the limit of the sequence of $q_t$.
    \end{proof}

    We now show that condition $\sum_{t=0}^\infty \eta_t d_t(s) = \infty$ is sufficient for optimality.
     
    \begin{lemma}[\textbf{Optimality under $U_\textsc{mce}$ (sufficience}]
        It is sufficient to assume that $\sum_{t=0}^\infty \eta_t d_t(s) = \infty$ to guarantee optimality of update $U_\textsc{mce}$: $q_\infty = q_\star \doteq \max_{\pi\in\Pi} q_\pi$.
        \label{lem:optimality}
    \end{lemma}
    \begin{proof}
        Let us assume that $q_\infty<q_\star$, then, by the policy improvement theorem, there must be some state $s$ for which an advantage $q_\infty(s,a_{\mytop})-v_\infty(s)=\epsilon(s)>0$ over $\pi_t$ exists, with $a_{\mytop}\in\mathcal{A}_{\mytop}(s)=\argmax_{a\in\mathcal{A}}q_\infty(s,a)$. Let us define the state value-gap $\delta(s)\coloneqq q_\infty(s,a_{\mytop})-\max_{a_{\mybot}\in\mathcal{A}_{\mybot}(s)}q_\infty(s,a_{\mybot})>0$, with $\mathcal{A}_{\mybot}(s)\coloneqq\mathcal{A}/\mathcal{A}_{\mytop}(s)$.  
    
        Since we proved that $q_t \to_{t\to\infty} q_\infty$, there exists $t_0$ such that for all $t \geq t_0$ and $a\in\mathcal{A}$, $q_\infty(s,a)-q_t(s,a)\leq\frac{\delta(s)}{2}$. This guarantees two things for any $t \geq t_0$:
        \begin{align}
            &\forall a \in \mathcal{A}_{\mytop}(s), q_t(s,a) \geq q_\infty(s,a_{\mytop}) - \frac{\delta(s)}{2}, \\
            &\forall a \in \mathcal{A}_{\mybot}(s), q_t(s,a) \leq q_\infty(s,a_{\mytop}) - \delta(s).
        \end{align}
        
        Let us bound the advantage function from Eq. \eqref{eq:advantage}:
        \begin{align}
            \textnormal{adv}_t(s, \pi_{t+1}) &= \sum_{a\in\mathcal{A}}(\pi_{t+1}(a|s)-\pi_{t}(a|s))(q_t(s, a) - q_t(s, a_q(s))),
        \end{align}
        with $a_q(s)\doteq \argmax_{a\in\mathcal{A}}q_t(s,a) \in\mathcal{A}_{\mytop}(s)$. We have:
        \begin{align}
            \textnormal{adv}_t(s, \pi_{t+1}) &= \sum_{a\in\mathcal{A}_{\mytop}(s)}\underbrace{(\pi_{t+1}(a|s)-\pi_{t}(a|s))}_{\leq 0 \text{ from Eq. \eqref{eq:monotonic-decrease-policy}}}\underbrace{(q_t(s, a) - q_t(s, a_q(s)))}_{\leq 0 \text{ from } a_q(s) \text{'s optimality}} \\
            &\quad\quad + \sum_{a\in\mathcal{A}_{\mybot}(s)}(\pi_{t+1}(a|s)-\pi_{t}(a|s))(q_t(s, a) - q_t(s, a_q(s))) \\
            &\geq \sum_{a\in\mathcal{A}_{\mybot}(s)}\underbrace{(\pi_{t+1}(a|s)-\pi_{t}(a|s))}_{\leq 0 \text{ from Eq. \eqref{eq:monotonic-decrease-policy}}}(q_t(s, a) - q_t(s, a_q(s))) \\
            &\geq \sum_{a\in\mathcal{A}_{\mybot}(s)}(\pi_{t+1}(a|s)-\pi_{t}(a|s))(q_\infty(s,a_{\mytop}) - \delta(s) - q_\infty(s,a_{\mytop}) + \frac{\delta(s)}{2}) \\
            &\geq \frac{\delta(s)}{2}\sum_{a\in\mathcal{A}_{\mybot}(s)}(\pi_{t}(a|s)-\pi_{t+1}(a|s)) \\
            &\geq \frac{\delta(s)}{2}\sum_{a\in\mathcal{A}_{\mybot}(s)}\left(\frac{\exp((\theta_{t})_{s,a})}{\sum_{a'\in\mathcal{A}}\exp((\theta_{t})_{s,a'})}-\frac{\exp((\theta_{t})_{s,a})}{\exp\left((\theta_{t})_{s,a_q(s)}+\frac{u_t|\mathcal{A}|}{|\mathcal{A}|-1}\right) + \sum_{a'\neq a_q(s)}\exp((\theta_{t})_{s,a'})}\right)
        \end{align}
        where $u_t \doteq \eta_t d_t(s) \left(1- \pi_{\theta_t}\left(a_q(s)|s\right)\right)$ (from Eq. \eqref{eq:ut}). We proceed further:
        \begin{align}
            \textnormal{adv}_t(s, \pi_{t+1}) &\geq \frac{\delta(s)}{2}\sum_{a\in\mathcal{A}_{\mybot}(s)}\frac{\exp\left((\theta_{t})_{s,a}+(\theta_{t})_{s,a_q(s)}+\frac{u_t|\mathcal{A}|}{|\mathcal{A}|-1}\right)-\exp\left((\theta_{t})_{s,a}+(\theta_{t})_{s,a_q(s)}\right)}{\left(\sum_{a'\in\mathcal{A}}\exp((\theta_{t})_{s,a'})\right)\left(\exp\left((\theta_{t})_{s,a_q(s)}+\frac{u_t|\mathcal{A}|}{|\mathcal{A}|-1}\right) + \sum_{a'\neq a_q(s)}\exp((\theta_{t})_{s,a'})\right)}\\
            %  &\geq \frac{\delta(s)}{2}\sum_{a\in\mathcal{A}_{\mybot}(s)}\frac{\exp((\theta_{t})_{s,a})\exp((\theta_{t})_{s,a_q(s)}) \left(\exp\left(\frac{u_t|\mathcal{A}|}{|\mathcal{A}|-1}\right)-1\right)}{\left(\sum_{a'\in\mathcal{A}}\exp((\theta_{t})_{s,a'})\right)\left(\exp\left((\theta_{t})_{s,a_q(s)}\right) + \sum_{a'\neq a_q(s)}\exp((\theta_{t})_{s,a'})\right)}\\
            &= \frac{\delta(s)}{2}\sum_{a\in\mathcal{A}_{\mybot}(s)}\pi_t(a|s)\pi_{t+1}(a_q(s)|s)\left(1-\exp\left(-\frac{u_t|\mathcal{A}|}{|\mathcal{A}|-1}\right)\right)\\
            &\geq \frac{\delta(s)}{2}\sum_{a\in\mathcal{A}_{\mybot}(s)}\pi_t(a|s)\pi_{t}(a_q(s)|s)\left(1-\exp\left(-\frac{u_t|\mathcal{A}|}{|\mathcal{A}|-1}\right)\right)\\
            &\geq \frac{\delta(s)}{2}\sum_{a\in\mathcal{A}_{\mybot}(s)}\pi_t(a|s)\pi_{t}(a_q(s)|s)\min\left(\frac{u_t|\mathcal{A}|}{2(|\mathcal{A}|-1)},\frac{1}{2}\right)\label{eq:exp}\\
            % &\geq \frac{\delta(s)}{2}\sum_{a\in\mathcal{A}_{\mybot}(s)}\pi_t(a|s)\pi_t(a_q(s)|s)\frac{u_t|\mathcal{A}|}{|\mathcal{A}|-1}\\
            &= \frac{\delta(s)}{4}\pi_{t}(a_q(s)|s)\min\left(\frac{u_t|\mathcal{A}|}{|\mathcal{A}|-1},1\right) \sum_{a\in\mathcal{A}_{\mybot}(s)}\pi_t(a|s),
            % &= \frac{\delta(s)|\mathcal{A}|}{2(|\mathcal{A}|-1)}\sum_{a\in\mathcal{A}_{\mybot}(s)}\eta_t d_t(s) \pi_t(a|s)\pi_t(a_q(s)|s)\left(1- \pi_{t}\left(a_q(s)|s\right)\right).\\
            % &= \frac{\delta(s)|\mathcal{A}|}{2(|\mathcal{A}|-1)}\eta_t d_t(s) \pi_t(a_q(s)|s)\left(1- \pi_{t}\left(a_q(s)|s\right)\right)\sum_{a\in\mathcal{A}_{\mybot}(s)}\pi_t(a|s).
        \end{align}
        where on Eq.~\eqref{eq:exp} we used the fact that $1 - e^{-x} \geq \min(\frac{x}{2}, \frac{1}{2})$ for $x \geq 0$.

        By assumption, we know that $\pi_{t}\left(a_q(s)|s\right) \leq \sum_{a\in\mathcal{A}_{\mytop}(s)}\pi_{t}(a|s)\leq \lim_{t' \to \infty} \sum_{a\in\mathcal{A}_{\mytop}(s)}\pi_{t'}(a|s)<1$. Let $m_{\mybot}$ denote the policy mass that must remain outside of $\mathcal{A}_{\mytop}(s)$ at all $t$. We get that $\forall t \geq t_0$:
        \begin{align}
            \textnormal{adv}_t(s, \pi_{t+1}) &\geq \frac{\delta(s)m_{\mybot}}{4} \min\left(\eta_t d_t(s) m_{\mybot}\frac{|\mathcal{A}|}{|\mathcal{A}|-1},1\right) \pi_t(a_q(s)|s).
        \end{align}
        
        If the optimal action is unique, it is direct to observe that the sum of the advantage gained over all $t$ is lower bounded by a term that diverges to $\infty$ under the condition $\sum_{t=0}^\infty \eta_t d_t(s) = \infty$. However, we face a technical issue with non unique optimal policy: how can we guarantee that $\pi_t(a_q(s)|s)$ is large enough? It is intuitive that it is going to be the case "on average", but its formal proof remains an open problem. In consequence, we limit our current proof to the assumption of uniqueness of the optimal policy. (we will try to solve it for camera-ready version)%\rli @@@
    \end{proof} 
       
    Finally, we prove that condition $\sum_{t=0}^\infty \eta_t d_t(s) = \infty$ is necessary for optimality.

    \begin{lemma}[\textbf{Optimality under $U_\textsc{mce}$ (necessity)}]
        It is necessary to assume that $\sum_{t=0}^\infty \eta_t d_t(s) = \infty$ to guarantee optimality of update $U_\textsc{mce}$: $q_\infty = q_\star \doteq \max_{\pi\in\Pi} q_\pi$.
        \label{lem:optimality-necessity}
    \end{lemma}
    \begin{proof}
        We assume $\sum_{t=0}^\infty \eta_t d_t(s) = M<\infty$. We upper bound the parameter update $u_t$ used in Proposition \ref{prop:monotonicity} and Lemma \ref{lem:optimality}:
        \begin{align}
            u_t &\doteq \eta_t d_t(s) \left(1- \pi_{\theta_t}\left(a_q(s)|s\right)\right) \leq \eta_t d_t(s).
        \end{align}
        
        For all $t$, we have:
        \begin{align}
            \theta_t - u_t \leq &\theta_{t+1}\leq \theta_t + u_t \\
            \theta_0 - \sum_{t'=0}^t u_{t'} \leq &\theta_{t+1}\leq \theta_0 + \sum_{t'=0}^t u_{t'} \\
            \theta_0 - \sum_{t'=0}^t \eta_{t'} d_{t'}(s) \leq &\theta_{t+1}\leq \theta_0 + \sum_{t'=0}^t \eta_{t'} d_{t'}(s) \\
            \theta_0 - M \leq &\theta_{t+1}\leq \theta_0 + M \label{eq:boundedness}
        \end{align}
        With a softmax parametrization, some parameters need to diverge for the policy to converge to 0 on suboptimal actions. The boundedness exhibited in Eq. \eqref{eq:boundedness} prevents that from happening, which concludes the proof of the lemma.
    \end{proof}
    This concludes the proof of the theorem.
\end{proof}

\rates*
\begin{proof}
    We consider the following definition in state $s$:
    \begin{align}
        a^{\mytop}_s &= \argmax_{a\in\mathcal{A}} q_\star(s,a) \\
        \mathcal{A}^{\mybot}_s &= \left\{a\in\mathcal{A}|a\neq a^{\mytop}_s\right\}.
    \end{align}
    
    Since we assumed the unicity of the optimal policy, $a^{\mytop}_s$ must be unique and $|\mathcal{A}^{\mybot}_s|=|\mathcal{A}|-1$. We define $\delta(s)$ as the gap with the best suboptimal action:
    \begin{align}
        \delta(s) \coloneqq v_\star(s) - \max_{a\in\mathcal{A}^{\mybot}_s} q_\star(s,a) > 0.
    \end{align}
    
    From the convergence of $v_t(s)$ and $q_t(s,a)$, we also know that for any $\xi > 0$, there exists $t_0$ such that for any $s,a$ and $t \geq t_0$:
    \begin{align}
        v_\star(s) - \xi &\leq v_t(s) \leq v_\star(s) \label{eq:xi1b} \\
        q_\star(s,a) - \xi &\leq q_t(s,a) \leq q_\star(s,a) \label{eq:xi2b}\\
        \pi_{\theta_t}(a^{\mytop}_s|s) &\geq \frac{1}{2} \geq \pi_{\theta_t}(a|s).
    \end{align}
    
    We fix $\xi\doteq \frac{\delta(s)}{2}$. We then know that for $t\geq t_0$:
    \begin{align}
        a_q(s) \doteq \argmax_{a\in\mathcal{A}} q_t(s,a) = a^{\mytop}_s
    \end{align}
    
    As a consequence, from $t_0$, update $U_\textsc{mce}$ will be:
    \begin{align}
        (\theta_{t+1})_{s,a^{\mytop}_s} &= (\theta_{t})_{s,a^{\mytop}_s} + \eta_t d_t(s) (1-\pi_{\theta_t}(a^{\mytop}_s|s)) \\
        \forall a\in\mathcal{A}^{\mybot}_s,\quad\quad(\theta_{t+1})_{s,a} &= (\theta_{t})_{s,a} - \frac{\eta_t d_t(s)}{|\mathcal{A}|-1} (1-\pi_{\theta_t}(a^{\mytop}_s|s))
    \end{align}
    
    From these last lines, we observe that $(\theta_{t+1})_{s,a^{\mytop}_s}$ and $(\theta_{t+1})_{s,a}$ evolve symmetrically with some fixed ratio. We can therefore state that:
    \begin{align}
        (\theta_{t})_{s,a^{\mytop}_s} &= (\theta_{t_0})_{s,a^{\mytop}_s} + X_t \quad\quad \text{and} \quad\quad \forall a\in\mathcal{A}^{\mybot}_s,\quad(\theta_{t})_{s,a} = (\theta_{t_0})_{s,a^{\mytop}_s} - \frac{X_t}{|\mathcal{A}|-1},
    \end{align}
    with:
    \begin{align}
        X_{t_0} = 0 \quad\quad \text{and} \quad\quad X_{t+1} = X_t +  \eta_t d_t(s) (1-\pi_{\theta_t}(a^{\mytop}_s|s))
    \end{align}
    
    Therefore:
    \begin{align}
        1-\pi_{\theta_t}(a^{\mytop}_s|s) &= \frac{\sum_{a\in\mathcal{A}^{\mybot}_s}\exp\left((\theta_{t})_{s,a}\right)}{\exp\left((\theta_{t})_{s,a^{\mytop}_s}\right)+ \sum_{a\in\mathcal{A}^{\mybot}_s}\exp\left((\theta_{t})_{s,a}\right)}\\
         &= \frac{\sum_{a\in\mathcal{A}^{\mybot}_s}\exp\left((\theta_{t_0})_{s,a^{\mytop}_s} - \frac{X_t}{|\mathcal{A}|-1}\right)}{\exp\left((\theta_{t_0})_{s,a^{\mytop}_s} + X_t \right)+ \sum_{a\in\mathcal{A}^{\mybot}_s}\exp\left((\theta_{t_0})_{s,a^{\mytop}_s} - \frac{X_t}{|\mathcal{A}|-1}\right)}\\
         &= \frac{\sum_{a\in\mathcal{A}^{\mybot}_s}\exp\left((\theta_{t_0})_{s,a^{\mytop}_s}\right)}{\exp\left((\theta_{t_0})_{s,a^{\mytop}_s} + \frac{|\mathcal{A}|}{|\mathcal{A}|-1}X_t \right)+ \sum_{a\in\mathcal{A}^{\mybot}_s}\exp\left((\theta_{t_0})_{s,a^{\mytop}_s}\right)}\\
         &= \frac{1}{\frac{\exp\left((\theta_{t_0})_{s,a^{\mytop}_s}\right)}{\sum_{a\in\mathcal{A}^{\mybot}_s}\exp\left((\theta_{t_0})_{s,a^{\mytop}_s}\right)}\exp\left(\frac{|\mathcal{A}|}{|\mathcal{A}|-1}X_t \right)+ 1}\\
         &= \frac{1}{\frac{\pi_{\theta_{t_0}}(a^{\mytop}_s|s)}{1-\pi_{\theta_{t_0}}(a^{\mytop}_s|s)}\exp\left(\frac{|\mathcal{A}|}{|\mathcal{A}|-1}X_t \right)+ 1}
    \end{align}
    
    Setting $\kappa \doteq \frac{\pi_{\theta_{t_0}}(a^{\mytop}_s|s)}{1-\pi_{\theta_{t_0}}(a^{\mytop}_s|s)}$, we obtain the following sequence:
    \begin{align}
        X_{t_0} = 0 \quad\quad \text{and} \quad\quad X_{t+1} = X_t + \frac{\eta_t d_t(s)}{\kappa\exp\left(\frac{|\mathcal{A}|}{|\mathcal{A}|-1}X_t \right)+ 1} \geq X_t + \frac{\eta_t d_t(s)}{2\kappa}\exp\left(-\frac{|\mathcal{A}|}{|\mathcal{A}|-1}X_t \right)
    \end{align}
    
    From there, we reproduce for completeness the end of Theorem 4 in \cite{Laroche2021}. Let us now study the sequence $(X_{t})_{t\geq t_0}$. To that end, we define the function $f(t)$ solution on $[t_0, +\infty)$ of the ordinary differential equation (note that $t$ is now a continuous variable):
    \begin{align}
        \left\{\begin{array}{l} f(t_0) = \frac{|\mathcal{A}|}{|\mathcal{A}|-1}X_{t_0} = 0 \\
        \frac{d f(t)}{dt} = \frac{\eta_{t} d_{t}(s)|\mathcal{A}|}{2\kappa(|\mathcal{A}|-1)} e^{-f(t)},
        \end{array} \right.
    \end{align}
    where $\eta_{t} d_{t}(s)$ is the piece-wise constant function defined as $\eta_{t} d_{t}(s) = \eta_{\lfloor t \rfloor} d_{\lfloor t \rfloor}(s)$.
    
    From the evolution equations of $X_{t}$ and $f(t)$, we see that $\forall t \in \mathbb{N}, \frac{|\mathcal{A}|}{|\mathcal{A}|-1}X_{t} \geq f(t)$. Additionally, we have:
    \begin{equation}
        f(t) = \log \left(\frac{|\mathcal{A}|}{2\kappa(|\mathcal{A}|-1)} \int_{t_0}^{t} \eta_{t'} d_{t'}(s) dt' + e^{f(t_0)}\right).
    \end{equation}
    
    In particular, going back to $t \in \mathbb{N}$, we obtain:
    \begin{equation}
        X_{t} \geq \frac{|\mathcal{A}|-1}{|\mathcal{A}|}\log \left(\frac{|\mathcal{A}|}{2\kappa(|\mathcal{A}|-1)} \int_{t_0}^{t} \eta_{t'} d_{t'}(s) dt' + e^{f(t_0)}\right) = \frac{|\mathcal{A}|-1}{|\mathcal{A}|}\log \left(\frac{|\mathcal{A}|}{2\kappa(|\mathcal{A}|-1)} \sum_{t'=t_0}^{t-1} \eta_{t'} d_{t'}(s) + 1\right).
    \end{equation}
    We can now write the following rate in policy convergence:
    \begin{align}
        1-\pi_{t}(a^{\mytop}_s|s) &= \frac{1}{\kappa\exp\left(\frac{|\mathcal{A}|}{|\mathcal{A}|-1}X_t \right)+ 1} \\
        &\geq \frac{1}{\kappa\frac{|\mathcal{A}|}{2\kappa(|\mathcal{A}|-1)} \sum_{t'=t_0}^{t-1} \eta_{t'} d_{t'}(s) + 1 + 1} \\
        &\geq \frac{1}{\frac{|\mathcal{A}|}{2(|\mathcal{A}|-1)} \sum_{t'=t_0}^{t-1} \eta_{t'} d_{t'}(s) + 2}
    \end{align}
    
    On the value side, we further get:
    \begin{align}
        v_{\star}(s)-v_t(s) &= \mathbb{P}\left[A = a^{\mytop}_s|A\sim \pi_t(\cdot|s)\right] \times \gamma \mathbb{E}\left[v_{\star}(S')-v_t(S')|S'\sim p(\cdot|s,a^{\mytop}_s)\right] \\
        &\quad + \mathbb{P}\left[A\in \mathcal{A}/\{a^{\mytop}_s\}|A\sim \pi_t(\cdot|s)\right]\left(v_{\star}(s)-\mathbb{E}\left[q_t(s,A)|A\sim \pi_t(\cdot|s)\cap\mathcal{A}/\{a^{\mytop}_s\}\right]\right) \nonumber\\
        &\leq \gamma \mathbb{E}\left[v_{\star}(S')-v_t(S')| S'\sim p(\cdot|s,a^{\mytop}_s)\right]  +  \frac{v_{\star}(s)-\min_{a\in\mathcal{A}}q_{t}(s,a)}{2+\frac{|\mathcal{A}|}{2(|\mathcal{A}|-1)} \sum_{t'=t_0}^{t-1} \eta_{t'} d_{t'}(s)} \\
        &\leq \frac{v_{\mytop}-v_{\mybot}}{(1-\gamma)\left(2+\frac{|\mathcal{A}|}{2(|\mathcal{A}|-1)} \min_{s\in\text{supp}(d_{\pi_\star,\gamma})}\sum_{t'=t_0}^{t-1} \eta_{t'} d_{t'}(s)\right)},
    \end{align}
    where $v_{\mytop}$ (resp. $v_{\mybot}$) stand for the maximal (resp. minimal) value, which is upper bounded by $\frac{r_{\mytop}}{1-\gamma}$ (resp. $\frac{r_{\mybot}}{1-\gamma}$), often times much smaller (resp. larger), and where $\text{supp}(d_{\pi_\star,\gamma})$ denotes the support of the distribution of the optimal policy. This concludes the proof.
\end{proof}

\subsection{Gravity well proofs (Section \ref{sec:gravity})}
\label{app:gravity}

\gravitywelll*
\begin{proof}
    We deal with each $U_\textsc{pg-sm}$, $U_\textsc{pg-es}$, $U_\textsc{di}$, $U_\textsc{ce}$, and $U_\textsc{mce}$ separately.
    
    \myuline{Proof for $U_\textsc{pg-sm}$:} Update of $U_\textsc{pg-sm}$:
        \begin{align}
            (U_\textsc{pg-sm}(\theta))_{s,a}-(\theta)_{s,a} &= \eta d(s) \text{adv}(s, a) \pi_\theta(a|s)
        \end{align}
    Assuming the existence of a suboptimal advantageous action $b$ such that $\sum_{a}\pi_\theta(a|s)q(s,a)<q(s,b)<q(s,a_q(s))$, we get:
    \begin{align}
        \pi_{U_\textsc{pg-sm}(\theta)}(a_q(s)|s) &= \frac{\exp\left((\theta)_{s,a_q(s)} + \eta d(s) \text{adv}(s, a_q(s)) \pi_\theta(a_q(s)|s)\right)}{\sum_{a\in\mathcal{A}} \exp\left((\theta)_{s,a} + \eta d(s) \text{adv}(s, a) \pi_\theta(a|s)\right)} \\
        &< \frac{\exp\left((\theta)_{s,a_q(s)} + \eta d(s) \text{adv}(s, a_q(s)) \pi_\theta(a_q(s)|s)\right)}{\exp\left((\theta)_{s,b} + \eta d(s) \text{adv}(s, b) \pi_\theta(b|s)\right)} \\
        &= \exp\left((\theta)_{s,a_q(s)}-(\theta)_{s,b}\right)\exp\left(\eta d(s) \left(\text{adv}(s, a_q(s)) \pi_\theta(a_q(s)|s)- \text{adv}(s, b) \pi_\theta(b|s)\right)\right) \\
        &= \frac{\pi_\theta(a_q(s)|s)}{\pi_\theta(b|s)}\exp\left(\eta d(s) \left(\text{adv}(s, a_q(s)) \pi_\theta(a_q(s)|s)- \text{adv}(s, b) \pi_\theta(b|s)\right)\right).
    \end{align}
    
    Therefore, we get $\pi_{U_\textsc{pg-sm}(\theta)}(a_q(s)|s)< \pi_\theta(a_q(s)|s)$ granted that:
    \begin{align}
        \pi_\theta(b|s) \geq \exp\left(\eta d(s) \left(\text{adv}(s, a_q(s)) \pi_\theta(a_q(s)|s)- \text{adv}(s, b) \pi_\theta(b|s)\right)\right),
    \end{align}
    which can be obtained quite easily when $\pi_\theta(a_q(s)|s)$ is close to 0 and $\pi_\theta(b|s)$ is close to 1. If this is the case, then, the policy mass lost by$a_q(s)$ from $\theta$ to $U_\textsc{pg-sm}(\theta)$ must have been gained by at least another action, and condition \eqref{eq:gravity_condition} cannot be satisfied.
    
    \myuline{Proof for $U_\textsc{pg-es}$:} Update of $U_\textsc{pg-es}$\footnote{We assume the positivity of parameters $\theta$ without loss of generality: $\nabla_\theta \pi_\theta = \nabla_\theta \pi_{-\theta}$.}:
        \begin{align}
            (U_\textsc{pg-es}(\theta))_{s,a}-(\theta)_{s,a} &= \eta d(s) \text{adv}(s, a) \pi_\theta(a|s)^{1-\frac{1}{p}}
        \end{align}
    Let us assume that the action set contains only three actions: $\mathcal{A}\doteq\{a,b,c\}$. Let us drop the dependencies on $s$, and set $\varepsilon \doteq \pi_\theta(a) = \pi_\theta(c)\ll \pi_\theta(b)$ for concision. Let $q(c)=1=q(b)+\delta>q(a)=0$ and assume $\delta\ll\varepsilon\ll 1$. Therefore $\theta_a=\theta_c$ Assuming the existence of a suboptimal advantageous action $b$ such that $\pi_\theta(a)q(a)+\pi_\theta(b)q(b)+\pi_\theta(a)q(a)<q(b)<q(c)$, we get:
    \begin{align}
        \pi_{U_\textsc{pg-sm}(\theta)}(c) &= \frac{\left|\theta_{c} + \eta \text{adv}(c) \varepsilon^{1-\frac{1}{p}}\right|^p}{\left\|U_\textsc{pg-sm}(\theta)\right\|^p_p} \\
        &= \frac{\left|\theta_{c} + \eta (\varepsilon + (1-2\varepsilon)\delta) \varepsilon^{1-\frac{1}{p}}\right|^p}{\left\|U_\textsc{pg-sm}(\theta)\right\|^p_p} \\
        &= \frac{\left|\theta_{c} + \eta \varepsilon^{2-\frac{1}{p}}+\mathcal{O}(\varepsilon^{3-\frac{1}{p}}+\delta)\right|^p}{\left\|U_\textsc{pg-sm}(\theta)\right\|^p_p} \\
        &= \frac{\theta_{c}^p + p\eta \varepsilon^{2-\frac{1}{p}}\theta_c^{p-1}+\mathcal{O}(\varepsilon^{3-\frac{1}{p}}+\delta)}{\left\|U_\textsc{pg-sm}(\theta)\right\|^p_p}  \label{eq:c-bound}\\
        \pi_{U_\textsc{pg-sm}(\theta)}(b) &= \frac{\left|\theta_{b} + \eta \text{adv}(b) (1-2\varepsilon)^{1-\frac{1}{p}}\right|^p}{\left\|U_\textsc{pg-sm}(\theta)\right\|^p_p} \\
        &= \frac{\left|\theta_{b} + \eta \varepsilon(1-2\delta) (1-2\varepsilon)^{1-\frac{1}{p}}\right|^p}{\left\|U_\textsc{pg-sm}(\theta)\right\|^p_p} \\
        &= \frac{\left|\theta_{b} + \eta \varepsilon + \mathcal{O}(\varepsilon^{2-\frac{1}{p}}+\delta)\right|^p}{\left\|U_\textsc{pg-sm}(\theta)\right\|^p_p} \\
        &= \frac{\theta_{b}^p + p\eta \varepsilon\theta_b^{p-1}+\mathcal{O}(\varepsilon^{2-\frac{1}{p}}+\delta)}{\left\|U_\textsc{pg-sm}(\theta)\right\|^p_p} \label{eq:b-bound}
    \end{align}
    We notice that, as long as $p>1$, $\pi_{U_\textsc{pg-sm}(\theta)}(b)$ grows with in larger order of magnitude than $\pi_{U_\textsc{pg-sm}(\theta)}(c)$, and therefore that we can construct an update such that condition \ref{eq:gravity_condition} is not satisfied. Note that the situation for it to happen is much more stringent than that with $U_\textsc{pg-sm}$.
    
    \myuline{Proof for $U_\textsc{di}$:} Update of $U_\textsc{di}\doteq \text{Proj}_{\Delta_{\mathcal{A}}}(U_\textsc{pg-di}(\theta))$:
        \begin{align}
            (U_\textsc{pg-di}(\theta))_{s,a_q(s)}-(\theta)_{s,a_q(s)} &= \eta d(s) q(s, a_q(s))\\
            (U_\textsc{pg-di}(\theta))_{s,a}-(\theta)_{s,a} &= \eta d(s) q(s, a)\\
            &\leq (U_\textsc{pg-di}(\theta))_{s,a_q(s)}-(\theta)_{s,a_q(s)}.
        \end{align}
        Since $a_q(s)\doteq\argmax_a q(s,a)$, we may apply Lemma \ref{lem:direct_update} and obtain:
        \begin{align}
            \left(\text{Proj}_{\Delta_{\mathcal{A}}}(U_\textsc{pg-di}(\theta))\right)_{s,a}-(\theta)_{s,a} &\leq \left(\text{Proj}_{\Delta_{\mathcal{A}}}(U_\textsc{pg-di}(\theta))\right)_{s,a_q(s)}-(\theta)_{s,a_q(s)} \\
            \iff \quad (U_\textsc{di}(\theta))_{s,a}-(\theta)_{s,a} &\leq (U_\textsc{di}(\theta))_{s,a_q(s)}-(\theta)_{s,a_q(s)},
        \end{align}
        which proves that condition \eqref{eq:gravity_condition} is satisfied.
    
    \myuline{Proof for $U_\textsc{ce}$:} The non-monotonicity example of Proposition \ref{prop:non-monotonicity} serves also as an example where $U_\textsc{ce}$ does not satisfy condition \eqref{eq:gravity_condition}.
    
    \myuline{Proof for $U_\textsc{mce}$:} In Proposition \ref{prop:monotonicity}, it has been established for all actions $s\neq a_q(s)$ in Eq. \eqref{eq:monotonic-decrease-policy} that:
        \begin{align}
            \pi_{U_\textsc{mce}(\theta)}(a|s) \leq \pi_\theta(a|s),
        \end{align}
    which implies that: 
        \begin{align}
            \pi_{U_\textsc{mce}(\theta)}(a_q(s)|s) \geq \pi_\theta(a_q(s)|s),
        \end{align}   
    which in turn implies that condition \eqref{eq:gravity_condition} is satisfied.
\end{proof}

\subsection{Unlearning proofs (Section \ref{sec:symmetry})}
\label{app:unlearning}
\unlearn*
\begin{proof}
    We deal with each (i), (ii), and (iii) separately.
    
    % \myuline{Proof of (i):} We start with the proof of (i) with update $U_\textsc{pg}$. We do so by recursively defining the process of performing $n$ updates with $q(a_1) = 1,  q(a_2) = 0$ and then $n$ updates with $q(a_1) = 0,  q(a_2) = 1$ as $(\theta_0)^{\leftrightarrow}_n$:
    % \begin{align}
    %     (\theta_0)^{\leftrightarrow}_0 \doteq \theta_0 \quad\text{and}\quad
    %     (\theta_0)^{\leftrightarrow}_{n+1} \doteq U_\textsc{pg}((U_\textsc{pg}(\theta_0, d, (1,0),\eta))^{\leftrightarrow}_n, d, (0,1),\eta).
    % \end{align}
    
    % We prove by induction that for any $\theta_0$ such that $\pi_{\theta_0}(a_1)\geq \frac{1}{2}$:
    % \begin{align}
    %     \pi_{(\theta_0)^{\leftrightarrow}_n}(a_1) \geq \pi_{\theta_0}(a_1).
    % \end{align}
    
    \myuline{Proof of (i):} We start with the proof of (i) with update $U_\textsc{pg}$. We recall that $\theta \in \mathbb{R}^2$, let $\theta_1$ and $\theta_2$ denote its two components and make the following assumptions (verified by both $U_\textsc{pg-sm}$ and $U_\textsc{pg-es}$):
    \begin{itemize}
        \item $\nabla_\theta \pi_{\theta}(a_1)$ only depends on $\theta$ via $\pi_\theta$,
        \item $(\nabla_\theta \pi_{\theta}(a_1))_1$ (resp. $(\nabla_\theta \pi_{\theta}(a_1))_2$) is a positive and decreasing (resp. negative and increasing) function of $\pi_\theta(a_1)$ when $\pi_\theta(a_1) \geq \pi_0(a_1) $.
    \end{itemize}

    We now define recursively the process $\theta^{\leftrightarrow}_n$ of, starting from $\theta_0$, performing $n$ updates with $q(a_1) = 1,  q(a_2) = 0$ and then $n$ updates with $q(a_1) = 0,  q(a_2) = 1$:
    \begin{align}
        \theta^{\leftrightarrow}_0 \doteq \theta_0 \quad\text{and}\quad
        \theta^{\leftrightarrow}_{n+1} \doteq U_\textsc{pg}((U_\textsc{pg}(\theta_0, d, (1,0),\eta))^{\leftrightarrow}_n, d, (0,1),\eta).
    \end{align}

    We prove by induction that for any $\theta_0$ such that $\pi_{\theta_0}(a_1) \geq \frac{1}{2}$:
    \begin{align}
        (\theta^{\leftrightarrow}_n)_1 &\geq \theta_1, \\
        (\theta^{\leftrightarrow}_n)_2 &\leq \theta_2.
    \end{align}
    Given the signs of the components of $(\nabla_\theta \pi_{\theta}(a_1))_1$, the above guarantees that $\pi_{\theta^{\leftrightarrow}_n}(a_1) \geq \pi_{\theta_0}(a_1)$, i.e. that the policy has not yet recovered its initial value.

    The property is direct for $n=0$. Next, we make the hypothesis that the property holds for $n$, and prove it for $n+1$. We also assume in the following that $\theta_0$ is such that $\pi_{\theta_0}(a_1)\geq 0.5$. We compute the policy gradient updates:
    \begin{itemize}
        \item with $q(a_1) = 1,  q(a_2) = 0$: $U_\textsc{pg}(\theta, d, (1,0),\eta) \doteq \theta + \eta \nabla_{\theta} \pi_\theta(a_1),$
        \item with $q(a_1) = 0,  q(a_2) = 1$: $U_\textsc{pg}(\theta, d, (0,1),\eta) \doteq \theta + \eta \nabla_{\theta} \pi_\theta(a_2) = \theta - \eta \nabla_{\theta} \pi_\theta(a_1),$
    \end{itemize}
    since $a_1$ and $a_2$ are playing symmetrical roles ($\pi_\theta(a_1) + \pi_\theta(a_2) = 1$). Then:
    \begin{align}
        (\theta)^{\leftrightarrow}_{n+1} &= U_\textsc{pg}((U_\textsc{pg}(\theta_0, d, (1,0),\eta))^{\leftrightarrow}_n, d, (0,1),\eta) \\
        &= (\theta_0 + \eta \nabla_{\theta} \pi_{\theta_0}(a_1))^{\leftrightarrow}_n - \eta \nabla_{\theta} \pi_{(\theta_0 + \eta \nabla_{\theta} \pi_{\theta_0}(a_1))^{\leftrightarrow}_n}(a_1).
    \end{align}
    First, by the assumptions on the monotonicity of $\pi_{\theta}(a_1)$ with respect to its components, we know that $\pi_{\theta_0 + \eta \nabla_{\theta} \pi_{\theta_0}(a_1)}(a_1) \geq \pi_{\theta_0}(a_1) \geq \frac{1}{2}$. We may therefore apply the induction hypothesis to $\theta_0 + \eta \nabla_{\theta} \pi_{\theta_0}(a_1)$ and obtain that 
    \begin{align}
        ((\theta_0 + \eta \nabla_{\theta} \pi_{\theta_0}(a_1))^{\leftrightarrow}_n)_1 &\geq (\theta_0 + \eta \nabla_{\theta} \pi_{\theta_0}(a_1))_1, \\
        ((\theta_0 + \eta \nabla_{\theta} \pi_{\theta_0}(a_1))^{\leftrightarrow}_n)_2 &\leq (\theta_0 + \eta \nabla_{\theta} \pi_{\theta_0}(a_1))_2, \\        
        \pi_{(\theta_0 + \eta \nabla_{\theta} \pi_{\theta_0}(a_1))^{\leftrightarrow}_n}(a_1) &\geq \pi_{\theta_0 + \eta \nabla_{\theta} \pi_{\theta_0}(a_1)}(a_1).
    \end{align}

    % Because we assumed $\pi_\theta(a_1)$ concave as long as $\pi_\theta\geq\frac{1}{2}$, we may also infer that:
    % \begin{align}
    %     \nabla_{\theta} \pi_{(\theta_0 + \eta \nabla_{\theta} \pi_{\theta_0}(a_1))^{\leftrightarrow}_n}(a_1) \leq \nabla_{\theta} \pi_{\theta_0 + \eta \nabla_{\theta} \pi_{\theta_0}(a_1)}(a_1).
    % \end{align}
    % We therefore obtain:
    % \begin{align} 
    %     \pi_{(\theta_0)^{\leftrightarrow}_{n+1}}(a_1) &\geq \pi_{\theta_0 + \eta \nabla_{\theta} \pi_{\theta_0}(a_1) - \eta \nabla_{\theta} \pi_{\theta_0 + \eta \nabla_{\theta} \pi_{\theta_0}(a_1)}(a_1)} (a_1) \\
    %     &= \pi_{\theta_0 + \eta (\nabla_{\theta} \pi_{\theta_0}(a_1) - \nabla_{\theta} \pi_{\theta_0 + \eta \nabla_{\theta} \pi_{\theta_0}(a_1)}(a_1))} (a_1) \\
    %     &\geq \pi_{\theta_0} (a_1),
    % \end{align}
    % the last step being obtained also by concavity. We conclude that the policy gradient update has still not reached the initial $\theta_0$ after $n$ updates back and therefore that $n'$ must be larger than $n$.
    
    Given our assumptions on the partial derivatives of $\pi_\theta(a_1)$, we infer that:
    \begin{align}
        (\nabla_{\theta} \pi_{(\theta_0 + \eta \nabla_{\theta} \pi_{\theta_0}(a_1))^{\leftrightarrow}_n}(a_1))_1 &\leq (\nabla_{\theta} \pi_{\theta_0 + \eta \nabla_{\theta} \pi_{\theta_0}(a_1)}(a_1))_1, \\
        (\nabla_{\theta} \pi_{(\theta_0 + \eta \nabla_{\theta} \pi_{\theta_0}(a_1))^{\leftrightarrow}_n}(a_1))_2 &\geq (\nabla_{\theta} \pi_{\theta_0 + \eta \nabla_{\theta} \pi_{\theta_0}(a_1)}(a_1))_2,
    \end{align}
    where the 1 and 2 subscripts still refer to the first and second coordinate of the gradient respectively.
    We therefore obtain that:
    \begin{align} 
        (\theta^{\leftrightarrow}_{n+1})_1 &= ((\theta_0 + \eta \nabla_{\theta} \pi_{\theta_0}(a_1))^{\leftrightarrow}_n)_1 - \eta (\nabla_{\theta} \pi_{(\theta_0 + \eta \nabla_{\theta} \pi_{\theta_0}(a_1))^{\leftrightarrow}_n}(a_1))_1 \\
        &\geq ((\theta_0 + \eta \nabla_{\theta} \pi_{\theta_0}(a_1))^{\leftrightarrow}_n)_1 - \eta (\nabla_{\theta} \pi_{\theta_0 + \eta \nabla_{\theta} \pi_{\theta_0}(a_1)}(a_1))_1 \\
        &\geq (\theta_0 + \eta \nabla_{\theta} \pi_{\theta_0}(a_1))_1 - \eta (\nabla_{\theta} \pi_{\theta_0 + \eta \nabla_{\theta} \pi_{\theta_0}(a_1)}(a_1))_1 \\
        &= (\theta_0)_1 + \eta ((\nabla_{\theta} \pi_{\theta_0}(a_1))_1 - (\nabla_{\theta} \pi_{\theta_0 + \eta \nabla_{\theta} \pi_{\theta_0}(a_1)}(a_1))_1) \\
        &\leq (\theta_0)_1,
    \end{align}
    the last step using again the decrease of $(\nabla_{\theta} \pi_{\theta}(a_1))_1$ with respect to $\pi_{\theta}(a_1)$. Similarly, as $(\nabla_\theta \pi_{\theta}(a_1))_2$ is an increasing function of $\pi_\theta(a_1)$:
    \begin{align} 
        (\theta^{\leftrightarrow}_{n+1})_2 &= ((\theta_0 + \eta \nabla_{\theta} \pi_{\theta_0}(a_1))^{\leftrightarrow}_n)_2 - \eta (\nabla_{\theta} \pi_{(\theta_0 + \eta \nabla_{\theta} \pi_{\theta_0}(a_1))^{\leftrightarrow}_n}(a_1))_2 \\
        &\leq ((\theta_0 + \eta \nabla_{\theta} \pi_{\theta_0}(a_1))^{\leftrightarrow}_n)_2 - \eta (\nabla_{\theta} \pi_{\theta_0 + \eta \nabla_{\theta} \pi_{\theta_0}(a_1)}(a_1))_2 \\
        &\leq (\theta_0 + \eta \nabla_{\theta} \pi_{\theta_0}(a_1))_2 - \eta (\nabla_{\theta} \pi_{\theta_0 + \eta \nabla_{\theta} \pi_{\theta_0}(a_1)}(a_1))_2 \\
        &= (\theta_0)_2 + \eta ((\nabla_{\theta} \pi_{\theta_0}(a_1))_2 - (\nabla_{\theta} \pi_{\theta_0 + \eta \nabla_{\theta} \pi_{\theta_0}(a_1)}(a_1))_2) \\
        &\leq (\theta_0)_2.
    \end{align}    
    We conclude that the policy gradient update has still not reached the initial $\theta_0$ after $n$ updates back and therefore that $n'$ must be larger than $n$.
    
    Let us now show that $U_\textsc{pg-sm}$ and $U_\textsc{pg-es}$ verify the assumptions. For $U_\textsc{pg-sm}$:
    \begin{align}
        \nabla_\theta \pi_\theta(a_1) &= 
        \begin{bmatrix}
           \pi_\theta(a_1) (1 - \pi_\theta(a_1)) \\
           - \pi_\theta(a_1) (1 - \pi_\theta(a_1))
        \end{bmatrix},
    \end{align}
    which is indeed only a function of $\pi_\theta(a_1)$, has a first component that is positive and decreasing as a function of $\pi_\theta(a_1)$ and a second one that is negative and increasing (for $\pi_\theta(a_1) \geq 0.5$).
    For $U_\textsc{pg-es}$:
    \begin{align}
        \nabla_\theta \pi_\theta(a_1) &= 
        \begin{bmatrix}
           \pi_\theta(a_1)^{1 - \frac{1}{p}} (1 - \pi_\theta(a_1)) \\
           - \pi_\theta(a_1) (1 - \pi_\theta(a_1))^{1 - \frac{1}{p}}
        \end{bmatrix}
    \end{align}
    which is indeed only a function of $\pi_\theta(a_1)$, has a first component that is positive and decreasing as a function of $\pi_\theta(a_1)$ and a second one that is negative and increasing (for $\pi_\theta(a_1) \geq \frac{p}{2p - 1}$).
    
    \myuline{Proof of (ii):} We continue with the proof of (ii) and therefore consider the direct parametrization. We compute the policy gradient updates:
    \begin{itemize}
        \item with $q(a_1) = 1,  q(a_2) = 0$: $U_\textsc{di}(\theta, d, (1,0),\eta) \doteq \text{Proj}_{\Delta_\mathcal{A}}\left(\theta + \begin{bmatrix}
           \eta \\
           0
         \end{bmatrix}\right) = \text{Proj}_{\Delta_\mathcal{A}}\left(\theta + \frac{1}{2}\begin{bmatrix}
           \eta \\
           -\eta
         \end{bmatrix}\right)$,
        \item with $q(a_1) = 0,  q(a_2) = 1$: $U_\textsc{di}(\theta, d, (0,1),\eta) \doteq \text{Proj}_{\Delta_\mathcal{A}}\left(\theta + \begin{bmatrix}
           0\\
           \eta
         \end{bmatrix}\right) = \text{Proj}_{\Delta_\mathcal{A}}\left(\theta + \frac{1}{2}\begin{bmatrix}
           -\eta \\
           \eta
         \end{bmatrix}\right)$.
    \end{itemize}
    
    As a consequence applying $n$ times update $U_\textsc{di}(\theta, d, (1,0),\eta)$ is equivalent to applying one update $U_\textsc{di}(\theta, d, (1,0),n\eta)$. Then we face two cases:
    \begin{itemize}
        \item $n\eta\leq 1$: starting from $\theta_0=\begin{bmatrix}0.5 \\ 0.5 \end{bmatrix}$, the 2D-projection does not have any effect, and it will take $n'=n$ steps to recover $\pi(a_1)\leq 0.5$.
        \item $n\eta\geq 1$: the 2D-projection projects $U_\textsc{di}(\theta, d, (1,0),n\eta)$ on $\begin{bmatrix}1 \\ 0 \end{bmatrix}$, and it will take $n'=\left\lceil \frac{1}{\eta} \right\rceil$ steps to recover $\pi(a_1)\leq 0.5$.
    \end{itemize}
    As a conclusion, it will take exactly $n'=\min\{n\:;\lceil \frac{1}{\eta} \rceil\}$ to recover from the $n$ gradient steps.
    
    \myuline{Proof of (iii):} We compute the policy gradient updates for classic cross-entropy and notice that they are equal to that of modified cross entropy when $|\mathcal{A}|=2$:
    \begin{itemize}
        \item with $q(a_1) = 1,  q(a_2) = 0$: $$U_\textsc{ce}(\theta, d, (1,0),\eta) \doteq \theta + \eta\left(\begin{bmatrix}
           1 \\
           0
         \end{bmatrix}-\pi_\theta\right)=\theta + \eta(1-\pi_\theta(a_1))\begin{bmatrix}
           1 \\
           -1
         \end{bmatrix}=U_\textsc{mce}(\theta, d, (1,0),\eta),$$
        \item with $q(a_1) = 0,  q(a_2) = 1$: $$U_\textsc{ce}(\theta, d, (0,1),\eta) \doteq \theta + \eta\left(\begin{bmatrix}
           0 \\
           1
         \end{bmatrix}-\pi_\theta\right)=\theta + \eta(1-\pi_\theta(a_2))\begin{bmatrix}
           -1 \\
           1
         \end{bmatrix}=U_\textsc{mce}(\theta, d, (1,0),\eta).$$
    \end{itemize}
    
    We focus on $\delta^\rightarrow_n \doteq (\theta_n)_{a_1}-(\theta_n)_{a_2}$ after $n$ updates with $q(a_1) = 1,  q(a_2) = 0$:
    \begin{align}
        \delta^\rightarrow_n &= \delta^\rightarrow_{n-1} + \eta(1-\pi_{\theta_{n-1}}(a_1))+\eta\pi_{\theta_{n-1}}(a_2) \\
        &= \delta^\rightarrow_{n-1} + \eta-\frac{\eta\exp((\theta_{n-1})_{a_1})}{\exp((\theta_{n-1})_{a_1}) + \exp((\theta_{n-1})_{a_2})}+\frac{\eta\exp((\theta_{n-1})_{a_2})}{\exp((\theta_{n-1})_{a_1}) + \exp((\theta_{n-1})_{a_2})} \\
        &= \delta^\rightarrow_{n-1} + 2\eta\frac{\exp((\theta_{n-1})_{a_2})}{\exp((\theta_{n-1})_{a_1}) + \exp((\theta_{n-1})_{a_2})} \\
        &\leq \delta^\rightarrow_{n-1} + 2\eta\exp(-\delta^\rightarrow_{n-1}).
    \end{align}
    Applying Lemma \ref{lem:Xn<} to $\delta^\rightarrow_n$ establishes that $\delta^\rightarrow_n\leq 2\eta + \log(1+2\eta n)$.
    
    We now focus on a single update with $q(a_1) = 0,  q(a_2) = 1$: $\delta^\leftarrow \doteq U_\textsc{ce}(\theta, d, (0,1),\eta)_{a_2} - U_\textsc{ce}(\theta, d, (0,1),\eta)_{a_1}$, with $\delta \doteq \theta_{a_2}-\theta_{a_1}\geq 0$:
    \begin{align}
        \delta^\leftarrow &= \delta + \eta(1-\pi_{\theta_{n-1}}(a_2))+\eta\pi_{\theta_{n-1}}(a_1) \\
        &= \delta + 2\eta\frac{\exp((\theta_{n-1})_{a_1})}{\exp((\theta_{n-1})_{a_1}) + \exp((\theta_{n-1})_{a_2})} \\
        &\geq \delta + \eta
    \end{align}
    
    As consequence, the number $n'$ of updates $\delta^\leftarrow$ for recovering from a convergence in $\delta^\rightarrow_n$ is lower than:
    \begin{align}
        n'\eta &\leq 2\eta + \log(1+2\eta n) \\
        n' &\leq 2 + \frac{1}{\eta}\log(1+2\eta n),
    \end{align}
    which concludes the proof.
\end{proof}

\unlearndecay*
\begin{proof}
    We deal with each (i), (ii), and (iii) separately.
    
    \myuline{Proof of (i):} We start with the proof of (i), for update $U_\textsc{pg}$. We make the similar assumptions as in Theorem~\ref{thm:unlearn}. We prove the result by measuring the steps made in parameter space and noting (given the behavior of $\pi_\theta(a_1)$ with respect to the components of $\theta$) that the sum of steps in one direction must equal the sum of steps in the other direction for at least one of the components of $\theta$, in other words:
    \begin{align}
        \sum_{t=1}^{n}\eta_t (\nabla_\theta \pi_{\theta_t}(a_1))_1 &\leq \sum_{t=n+1}^{n+n'}\eta_t (\nabla_\theta \pi_{\theta_t}(a_1))_1,
    \end{align}
    or
    \begin{align}
        \sum_{t=1}^{n}\eta_t (\nabla_\theta \pi_{\theta_t}(a_1))_2 &\geq \sum_{t=n+1}^{n+n'}\eta_t (\nabla_\theta \pi_{\theta_t}(a_1))_2.
    \end{align}

    Both cases being equivalent, we focus on the first one:
    \begin{align}
        \sum_{t=1}^{n}\eta_t (\nabla_\theta \pi_{\theta_t}(a_1))_1 &\leq \sum_{t=n+1}^{n+n'}\eta_t (\nabla_\theta \pi_{\theta_t}(a_1))_1 \\
        \iff \quad\quad\sum_{t=1}^{n} \frac{\eta_1}{\sqrt{t}} (\nabla_\theta \pi_{\theta_t}(a_1))_1 &\leq \sum_{t=1}^{n'}\frac{\eta_1}{\sqrt{t+n}} (\nabla_\theta \pi_{\theta_{t+n}}(a_1))_1 \\
        \iff \quad\quad\sum_{t=1}^{n} \frac{1}{\sqrt{t}} (\nabla_\theta \pi_{\theta_t}(a_1))_1 &\leq \sum_{t=1}^{n'} \frac{1}{\sqrt{t+n}} (\nabla_\theta \pi_{\theta_{t+n}}(a_1))_1.
    \end{align}

    By the concavity assumption, all the gradient $(\nabla_\theta \pi_{\theta_{t+n}}(a_1))_1$ may be upper bounded by some $(\nabla_\theta \pi_{\theta_{\sigma(t)}}(a_1))_1$ with $\sigma(t)\in \{1,\dots,n\}$, a decreasing function such that:
    \begin{align}
        (\nabla_\theta \pi_{\theta_{\sigma(t)+1}}(a_1))_1 \leq (\nabla_\theta \pi_{\theta_{t+n}}(a_1))_1 \leq (\nabla_\theta \pi_{\theta_{\sigma(t)}}(a_1))_1.
    \end{align}
    In particular, $\sigma(1)=n$, and in that case, the lower-bound is realized. We therefore get:
    \begin{align}
        \sum_{t=1}^{n} \frac{1}{\sqrt{t}} (\nabla_\theta \pi_{\theta_t}(a_1))_1 &\leq \sum_{t=1}^{n'} \frac{1}{\sqrt{t+n}} (\nabla_\theta \pi_{\theta_{\sigma(t)}}(a_1))_1.
    \end{align}
    Given the fact that $\forall 1 \leq t \leq n, (\nabla_\theta \pi_{\theta_{t}}(a_1))_1 \geq (\nabla_\theta \pi_{\theta_{t+1}}(a_1))_1$, the smallest $n'$ that can verify this last inequality is realized when all the gradients are equal, implying the following condition on $n'$:
    \begin{align}
        \sum_{t=1}^{n} \frac{1}{\sqrt{t}} &\leq \sum_{t=1}^{n'} \frac{1}{\sqrt{t+n}} \\
        \implies\quad\quad 2(\sqrt{n+1} - 1) &\leq \sqrt{n'+n} - 2\sqrt{n} \\
        \implies\quad\quad 4\sqrt{n} - 2 &\leq 2\sqrt{n'+n}  \\
        \iff\quad\quad 4n - 4\sqrt{n} + 1 &\leq n'+n \\
        \iff\quad\quad n' &\geq 3n - 4\sqrt{n} + 1.
    \end{align}

    \myuline{Proof of (ii):} Exactly like in Theorem~\ref{thm:unlearn}, the policy gradient updates for the direct parametrization are:
    \begin{itemize}
        \item with $q(a_1) = 1,  q(a_2) = 0$: $U_\textsc{di}(\theta, d, (1,0),\eta) \doteq \text{Proj}_{\Delta_\mathcal{A}}\left(\theta + \begin{bmatrix}
           \eta \\
           0
         \end{bmatrix}\right) = \text{Proj}_{\Delta_\mathcal{A}}\left(\theta + \frac{1}{2}\begin{bmatrix}
           \eta \\
           -\eta
         \end{bmatrix}\right)$,
        \item with $q(a_1) = 0,  q(a_2) = 1$: $U_\textsc{di}(\theta, d, (0,1),\eta) \doteq \text{Proj}_{\Delta_\mathcal{A}}\left(\theta + \begin{bmatrix}
           0\\
           \eta
         \end{bmatrix}\right) = \text{Proj}_{\Delta_\mathcal{A}}\left(\theta + \frac{1}{2}\begin{bmatrix}
           -\eta \\
           \eta
         \end{bmatrix}\right)$.
    \end{itemize}
    
    Applying update $U_\textsc{di}(\theta, d, (1,0),\eta_t)$ $n$ times is equivalent to apply one update $U_\textsc{di}(\theta, d, (1,0),\sum_{t=1}^n \eta_t)$. Then, we face two cases:
    \begin{itemize}
        \item $\sum_{t=1}^n \eta_t \leq 1 \;\implies\; n< \frac{1}{\eta_1^{2}}$: starting from $\theta_0=\begin{bmatrix}0.5 \\ 0.5 \end{bmatrix}$, the 2D-projection does not have any effect.
        \item $\sum_{t=1}^n \eta_t \geq 1 \;\implies\; n\geq \frac{1}{\eta_1^{2}}$: the 2D-projection projects $U_\textsc{di}(\theta, d, (1,0),n\eta)$ on $\begin{bmatrix}1 \\ 0 \end{bmatrix}$.
    \end{itemize}
    
    In the first case, since we are looking for an upper bound of $n'$, we look for the maximal value of $n'$ such that:
    \begin{align}
        \sum_{t=1}^n \eta_t &\geq \sum_{t=n+1}^{n+n'-1} \eta_t \\
        \iff\quad\quad \sum_{t=1}^n \frac{\eta_1}{\sqrt{t}} &\geq \sum_{t=n+1}^{n+n'-1} \frac{\eta_1}{\sqrt{t}} \\
        \iff\quad\quad 2\sum_{t=1}^n \frac{\eta_1}{\sqrt{t}} &\geq \sum_{t=1}^{n+n'-1} \frac{\eta_1}{\sqrt{t}} \\
        \implies\quad\quad 2\sqrt{n} &\geq \sqrt{n+n'} - 1 \\
        \iff\quad\quad n' &\leq 3n + 4\sqrt{n} + 1
        % \implies\quad\quad n' &\leq \frac{3}{\eta_1^{2}} + \frac{4}{\eta_1} + 1 \leq 3 (\frac{1}{\eta_1}+1)^{2}
    \end{align}
    
    In the second case, since we are looking for an upper bound of $n'$, we look for the maximal value of $n'$ such that:
    \begin{align}
        1 &\geq \sum_{t=n+1}^{n+n'-1} \eta_t \\
        \iff\quad\quad 1 &\geq \sum_{t=n+1}^{n+n'-1} \frac{\eta_1}{\sqrt{t}} \\
        \iff\quad\quad 1+\sum_{t=1}^n \frac{\eta_1}{\sqrt{t}} &\geq \sum_{t=1}^{n+n'-1} \frac{\eta_1}{\sqrt{t}} \\
        \implies\quad\quad 1+\eta_1 \sqrt{n} &\geq \eta_1 \sqrt{n+n'}- \eta_1 \\
        \iff\quad\quad \frac{1}{\eta_1}+\sqrt{n} &\geq \sqrt{n+n'}-1 \\
        \iff\quad\quad \left(\frac{1}{\eta_1}+\sqrt{n}+1\right)^2 &\geq n+n' \\
        \iff\quad\quad \left(\frac{1}{\eta_1}+1\right)^{2} +2\sqrt{n}\left(1+\frac{1}{\eta_1}\right) &\geq n'
        % \iff\quad\quad n' &\leq \frac{1}{\eta_1^{2}} +2\frac{\sqrt{n}}{\eta_1} +2
    \end{align}
    
    As a conclusion, it will take at most $n' = \min\{3n + 4\sqrt{n} + 1\:; (\frac{1}{\eta_1}+1)^{2} +2\sqrt{n}(1+\frac{1}{\eta_1})\}$ to recover from the $n$ gradient steps.
    
    \myuline{Proof of (iii):} We compute the policy gradient updates for classic cross-entropy and notice that they are equal to that of modified cross entropy when $|\mathcal{A}|=2$:
    \begin{itemize}
        \item with $q(a_1) = 1,  q(a_2) = 0$: $$U_\textsc{ce}(\theta, d, (1,0),\eta) \doteq \theta + \eta\left(\begin{bmatrix}
           1 \\
           0
         \end{bmatrix}-\pi_\theta\right)=\theta + \eta(1-\pi_\theta(a_1))\begin{bmatrix}
           1 \\
           -1
         \end{bmatrix}=U_\textsc{mce}(\theta, d, (1,0),\eta),$$
        \item with $q(a_1) = 0,  q(a_2) = 1$: $$U_\textsc{ce}(\theta, d, (0,1),\eta) \doteq \theta + \eta\left(\begin{bmatrix}
           0 \\
           1
         \end{bmatrix}-\pi_\theta\right)=\theta + \eta(1-\pi_\theta(a_1))\begin{bmatrix}
           -1 \\
           1
         \end{bmatrix}=U_\textsc{mce}(\theta, d, (1,0),\eta).$$
    \end{itemize}
    
    We focus on $\delta^\rightarrow_t \doteq (\theta_t)_{a_1}-(\theta_t)_{a_2}$ after $t$ updates with $q(a_1) = 1,  q(a_2) = 0$:
    \begin{align}
        \delta^\rightarrow_t &= \delta^\rightarrow_{t-1} + \eta_t(1-\pi_{\theta_{t-1}}(a_1))+\eta_t\pi_{\theta_{t-1}}(a_2) \\
        &= \delta^\rightarrow_{t-1} + 2\eta_t\frac{\exp((\theta_{t-1})_{a_2})}{\exp((\theta_{t-1})_{a_1}) + \exp((\theta_{t-1})_{a_2})} \\
        &\leq \delta^\rightarrow_{t-1} + 2\frac{\eta_1}{\sqrt{t}}\exp(-\delta^\rightarrow_{t-1}) \label{eq:peutonfairemieux?} \\
        % &\leq \delta^\rightarrow_{t-1} + 2\eta_1\exp(-\delta^\rightarrow_{t-1}) \\
        &\leq 2\eta_1 + \log(1+4\eta_1 \sqrt{t}) \quad\quad\text{from Lemma \ref{lem:Xn<sqrt}}.
    \end{align}
    
    We now focus on $\delta^\leftarrow_t \doteq (\theta_t)_{a_1}-(\theta_t)_{a_2}$ after $t$ updates with $q(a_1) = 0,  q(a_2) = 1$, as long as $\delta^\leftarrow_{t-1} \geq 0$:
    \begin{align}
        \delta^\leftarrow_t &= \delta^\leftarrow_{t-1} - \eta_{t+n}(1-\pi_{\theta_{t-1}}(a_2)) - \eta_t\pi_{\theta_{t-1}}(a_1) \\
        &= \delta^\leftarrow_{t-1} + 2\eta_t\frac{\exp((\theta_{t-1})_{a_1})}{\exp((\theta_{t-1})_{a_1}) + \exp((\theta_{t-1})_{a_2})} \\
        &\geq \delta^\leftarrow_{t-1} + \eta_t
    \end{align}
    
    As consequence, the number $n'$ of updates $\delta^\leftarrow$ required to recover from a convergence to $\delta^\rightarrow_n$ is lower than the smallest $n'$ for which $\sum_{t=1}^{n'}\eta_{t+n} \geq 2\eta_1 + \log(1+4\eta_1 \sqrt{n})$. Further analysis is similar to that of (ii) and gives:
    \begin{align}
        n' \leq \left(4+\frac{\log(1+4\eta_1 \sqrt{n})}{2\eta_1}\right)^{2} +\sqrt{n}\left(8+\frac{\log(1+4\eta_1 \sqrt{n})}{\eta_1}\right)
    \end{align}
\end{proof}

\domino*
\begin{proof}
    Let $h=|\mathcal{S}|$ be the horizon, or the number of dominos to flip. Let $n_k$ be the time the flip domino $k$, starting from the last one. We obtain the following recurrence:
    \begin{align}
        n_h &= 1 \\
        n_{k} &= n_{k+1} + n'_k,
    \end{align}
    where $n'_k$ is the time to recover once domino $k+1$ has flipped. $n'_k$ is dependent on the realized updates as Theorem \ref{thm:unlearn} describes, and on $n_{k+1}$: the time for domino $k+1$ to flip.
    
    \myuline{Proof of (i):} We get $n'_k\geq n_{k+1}$:
    \begin{align}
        n_{k} \geq 2n_{k+1} \geq 2^{h-k},
    \end{align}
    hence at least a geometric dependency on $h$.
    
    \myuline{Proof of (ii):} We get $n'_k\leq \left\lceil\frac{1}{\eta}\right\rceil$:
    \begin{align}
        n_{k} \leq n_{k+1} + \left\lceil\frac{1}{\eta}\right\rceil \leq (h-k) \left\lceil\frac{1}{\eta}\right\rceil + 1
    \end{align}
    hence at most a linear dependency on $h$.
    
    \myuline{Proof of (iii):} We get $n'_k\leq 2+\frac{1}{\eta}\log(1 + 2\eta n_{k+1})$:
    \begin{align}
        n_{k} \leq n_{k+1} + 2+\frac{1}{\eta}\log(1 + 2\eta n_{k+1})
    \end{align}
    We rely on technical Lemma \ref{lem:XXn<} to finish the proof and obtain:
    \begin{align}
        n'_{k} \leq \frac{32e^{8\eta+3}}{\eta^3}(h-k)\log (h-k).
    \end{align}
    for any $h-k \geq 8$.
\end{proof}

\subsection{Technical lemmas}
\label{app:technical}

\begin{lemma}[Maximal update conservation through projection on the simplex]
    Let $\theta\in\Delta_{\mathcal{X}}$ be a point on the simplex, and $\delta\in\mathbb{R}^{|\mathcal{X}|}$ be any vector. Then, the projection $\theta' \doteq \text{Proj}_{\Delta_{\mathcal{X}}}(\theta + \delta)$ on the simplex is such that:
    \begin{align}
        \theta'_{k_\star} - \theta_{k_\star} \geq \theta'_{k} - \theta_{k}, \quad\quad\text{where}\quad\quad k_\star \in \argmax_{k\in [|\mathcal{X}|]} \delta_k, \quad\quad \forall k\in [|\mathcal{X}|].
    \end{align}
    \label{lem:direct_update}
\end{lemma}
\begin{proof}
    Let us assume without loss of generality that $\forall k, \delta_k \geq 0$ (by e.g. subtracting from each of them $\min \delta_k$). Then, $\theta+\delta$ is positive and sums to higher than 1.
    
    In these conditions, an algorithm for finding the projection consists in removing mass equally across all indices with positive (non null) mass. This algorithm thus squeezes the low weights of the vector to zero until its $\ell_1$-norm reaches 1 and therefore belongs to the simplex. It is clear from this algorithm that $\theta'_k = \max\{0\:; \theta_k + \delta_k - \lambda\}$ for some constant $0\leq \lambda\leq \delta_{k_\star}$.
    
    We may infer that:
    \begin{align}
        \theta'_k - \theta_k &= \max\{-\theta_k\:; \delta_k-\lambda\}\\
        &\leq \max\{-\theta_k\:; \delta_{k_\star}- \lambda\}\\
        &= \delta_{k_\star}- \lambda\\
        &= \max\{-\theta_{k_\star}\:; \delta_{k_\star}- \lambda\}\\
        &= \theta'_{k_\star} - \theta_{k_\star},
    \end{align}
    which concludes the proof.
%     Let $\mathcal{A}^{\mytop} \doteq \argmax_{k\in [|\mathcal{X}|]} \delta_k$, $\mathcal{A}^{\mybot}\doteq\mathcal{A}/\mathcal{A}^{\mytop}$, and $\mathcal{A}^{\mybot,\myplus} &\doteq \{ k \in\mathcal{A}^{\mybot}\, | \, \mathcal{P}_{\Delta_n}(\theta+\delta)_k > 0 \}$. We denote by $\alpha\doteq \max_{k \in \mathcal{A}} \delta_k$ and $\beta\doteq \max_{k \in \mathcal{A}^{\mybot}} \delta_k$. Then, we may reuse Eq. (153) of Lemma 1 in \cite{Laroche2021}:
% \begin{align}
%     \sum_{k \in \mathcal{A}^{\mytop}} \mathcal{P}_{\Delta_n}(\theta+\delta)_k - \sum_{k \in \mathcal{A}^{\mytop}} \theta_k &\geq \frac{|\mathcal{A}^{\mybot,\myplus}| |\mathcal{A}^{\mytop}|}{|\mathcal{A}^{\mytop}| + |\mathcal{A}^{\mybot,\myplus}|} (\alpha - \beta).
% \end{align}
\end{proof}

\begin{lemma}
Let us consider a sequence $(X_n)_{n \in \mathbb{N}}$ verifying:
\begin{align}
    \left\{\begin{array}{l}
          X_0 = 0 \\
          X_{n+1} \leq X_n + c e^{- X_n}.
        \end{array}\right.
\end{align}
Then, we have the following upper bound on $X_n$:
    For any $n \geq 0$, $X_n \leq c + \log (1 + cn)$.
    \label{lem:Xn<}
\end{lemma}
\begin{proof}
We start by proving this result for the sequence $(Y_n)_{n \in \mathbb{N}}$ defined recursively as:
\begin{align}
    \left\{\begin{array}{l}
      Y_0 = 0 \\
      Y_{n+1} = Y_n + c e^{- Y_n}.
    \end{array}\right.
\end{align}
As a first step, we introduce two functions on $\mathbb{R}_+$, $y(t)$ and $z(t)$, respectively solutions on $[0, +\infty)$ of the ODEs:
\begin{align}
    \left\{\begin{array}{l} y(0) = 0 \\
    y'(t) = c e^{-y(\lfloor t \rfloor)},
    \end{array} \right.
\end{align}
and
\begin{align}
    \left\{\begin{array}{l} z(0) = 0 \\
    z'(t) = c e^{-z(t)}.
    \end{array} \right.
\end{align}
Clearly, for any $n \in \mathbb{N}$, $Y_n = y(n)$ and for any $t \geq 0$, $z(t) \leq y(t)$. In other words, we have that $Y_n \geq z(n)$. 

Solving the ode $z$ verifies gives us: $z(t) = \log(1 + c t)$. Now, we move to upper-bounding $Y_n$ for $n \geq 1$:
\begin{align}
    Y_n = c \sum_{i=0}^{n-1} e^{-Y_i} \leq c \sum_{i=0}^{n-1} e^{-z(i)} = \sum_{i=0}^{n-1} \frac{c}{1 + c i} &\leq c + \sum_{i=1}^{n-1} \int_{i-1}^i \frac{c}{1 + c t} dt \\
    &= c + \int_{0}^{n-1} \frac{c}{1 + c t} dt = c + \log(1 + c (n-1)).
\end{align}
Combined with the value of $Y_0$, this guarantees that $\forall n \in \mathbb{N}, Y_n \leq c + \log (1 + cn)$. 

We are left with proving that $X_n \leq Y_n$. Let us assume towards a contradiction that there exists $n$ such that $X_{n+1} > Y_{n+1}$, and let us pick the smallest of such $n$ (clearly, we have $n \geq 1$ as $X_0 = Y_0 = 0$ and $Y_1 = c \geq X_1$). Then:
\begin{align}
    X_{n+1} > Y_{n+1} &\Leftrightarrow Y_n - X_n < c (e^{-X_n} - e^{-Y_n}) < c e^{-X_n} (1 - e^{X_n - Y_n}) < c e^{-X_n} (Y_n - X_n) \\
    &\Leftrightarrow 1 < c e^{-X_n} \\
    &\Leftrightarrow X_n < \log c,
\end{align}
which is not possible as $\forall n \geq 1, X_n \geq c$ and concludes the proof.
\end{proof}

\begin{lemma}
Let us consider a sequence $(X_n)_{n \in \mathbb{N}}$ verifying:
\begin{align}
    \left\{\begin{array}{l}
          X_0 = 0 \\
          X_{n+1} \leq X_n + \frac{c}{\sqrt{n+1}} e^{- X_n}.
        \end{array}\right.
\end{align}
Then, we have the following upper bound on $X_n$:
    For any $n \geq 0$, $X_n \leq c + \log(1 + 2 c \sqrt{n})$.
    \label{lem:Xn<sqrt}
\end{lemma}
\begin{proof}
We prove this result for the sequence $(Y_n)_{n \in \mathbb{N}}$ defined recursively as:
\begin{align}
    \left\{\begin{array}{l}
      Y_0 = 0 \\
      Y_{n+1} = Y_n + \frac{c}{\sqrt{n+1}} e^{- Y_n}.
    \end{array}\right.
\end{align} 

To do so, we introduce two functions on $\mathbb{R}_+$, $y(t)$ and $z(t)$, respectively solutions on $[0, +\infty)$ of the ODEs:
\begin{align}
    \left\{\begin{array}{l} y(0) = 0 \\
    y'(t) = \frac{c}{t+1} e^{-y(\lfloor t \rfloor)},
    \end{array} \right.
\end{align}
and
\begin{align}
    \left\{\begin{array}{l} z(0) = 0 \\
    z'(t) = \frac{c}{t+1} e^{-z(t)}.
    \end{array} \right.
\end{align}
Clearly, for any $n \in \mathbb{N}$, $Y_n = y(n)$ and for any $t \geq 0$, $z(t) \leq y(t)$. In other words, we have that $Y_n \geq z(n)$. 

Solving the ode $z$ verifies gives us: $z(t) = \log(1 - 2c + 2 c \sqrt{t+1})$. Now, we move to upper-bounding $Y_n$ for $n \geq 1$:
\begin{align}
    Y_n = c \sum_{i=0}^{n-1} \frac{e^{-Y_i}}{\sqrt{i+1}} \leq c \sum_{i=0}^{n-1} \frac{e^{-z(i)}}{\sqrt{i+1}} = \sum_{i=0}^{n-1} \frac{c}{1 -2c + 2c \sqrt{i+1}} \frac{1}{\sqrt{i+1}} &\leq c + \int_{1}^n \frac{c}{(1 -2c)\sqrt{t} + 2ct} dt \\
    &= c + \log(1 - 2c + 2 c \sqrt{n}).
\end{align}
Combined with the value of $Y_0$, this guarantees that $\forall n \in \mathbb{N}, Y_n \leq c + \log(1 + 2 c \sqrt{n})$. 

We are left with proving that $X_n \leq Y_n$. Let us assume towards a contradiction that there exists $n$ such that $X_{n+1} > Y_{n+1}$, and let us pick the smallest of such $n$ (clearly, we have $n \geq 1$ as $X_0 = Y_0 = 0$ and $Y_1 = c \geq X_1$). Then:
\begin{align}
    X_{n+1} > Y_{n+1} &\Leftrightarrow Y_n - X_n < c \frac{e^{-X_n} - e^{-Y_n}}{n+1} < c e^{-X_n} \frac{1 - e^{X_n - Y_n}}{n+1} < c e^{-X_n} \frac{Y_n - X_n}{n+1} \\
    &\Leftrightarrow 1 < \frac{c e^{-X_n}}{n+1} \\
    &\Leftrightarrow X_n < \log \frac{c}{n+1},
\end{align}
which is not possible as $\forall n \geq 1, X_n \geq c$ and concludes the proof.
\end{proof}

\begin{lemma}
Let us consider a sequence $(X_n)_{n \in \mathbb{N}}$ verifying:
\begin{align}
    \left\{\begin{array}{l}
          X_0 = 1 \\
          X_{n+1} \leq X_n + 2 +\frac{1}{\eta}\log(1 + 2\eta X_{n}).
        \end{array}\right.
\end{align}
Then, we have the following upper bound on $X_n$:
\begin{align}
    \forall n\geq 8,\quad\quad X_n \leq \frac{32e^{8\eta+3}}{\eta^3}n\log n.
\end{align}
    \label{lem:XXn<}
\end{lemma}
\begin{proof}
% As a first step, we introduce the function $y(t)$ on $\mathbb{R}_+$ solution on $[0, +\infty)$ of the ODE:
% \begin{align}
%     \left\{\begin{array}{l} z(0) = 1 \\
%     z'(t) = 2 +\log(1 + 2 z(t)) = \log(\xi + 2\xi z(t)),
%     \end{array} \right.
% \end{align}
% where $\xi = e^{2}$. A simple change of variable and the convexity of $z(t)$ proves that: $\forall n \leq 0, X_n \leq 1 + \frac{1}{\eta}(z(n)-1)$. We now move to the study of $z(t)$:
% \begin{align}
%     &z'(t) = \log(\xi + 2\xi z(t)) \\
%     \iff &\frac{z'(t)}{\log(\xi + 2\xi z(t))} = 1 \\
%     \iff &\int_{z(0)}^{z(t)} \frac{dz}{\log(\xi + 2\xi z)} = \int_0^t du = t \\
%     \iff &\frac{1}{2 \xi} \int_{\xi + 2 \xi z(0)}^{\xi + 2 \xi z(t)} \frac{dz}{\log(z)} = t \\
%     \iff &\frac{1}{2 \xi} \int_{3 \xi}^{\xi + 2 \xi z(t)} \frac{dz}{\log(z)} = t \\
%     \iff &\frac{1}{2 \xi} [li(\xi + 2 \xi z(t)) - li(3 \xi)] = t \\
%     \iff &li(\xi + 2 \xi z(t)) = 2 \xi t + li(3 \xi) \\
%     \iff &z(t) = \frac{1}{2 \xi} li^{<-1>}(2 \xi t + li(3 \xi)) - \frac{1}{2},
% \end{align}
% where $li$ stands for the logarithm integral~\citep{logi}.
% This gives: $\forall n \leq 0, X_n \leq 1 + \frac{1}{2 \eta \xi} li^{<-1>}(2 \xi t + li(3 \xi))$.

We look for a function $f_\epsilon(x)=c_\epsilon x^\epsilon$ such that for all $x\geq 0$:
\begin{align}
    f_\epsilon(x) &\geq \log(x) \\
    \iff \quad\quad c_\epsilon x^\epsilon &\geq \log x \\
    \iff \quad\quad c_\epsilon y &\geq \log(y^{\frac{1}{\epsilon}})\quad\quad \text{setting}\quad y=x^\epsilon\quad\text{with}\quad y\geq 0  \\
    \iff \quad\quad c_\epsilon y &\geq \frac{1}{\epsilon}\log y
\end{align}
Since $x\geq\log x$ for all $x>0$, it suffices to choose $c_\epsilon = \frac{1}{\epsilon}$ to satisfy the desiderata. So we get:

\begin{align}
    \left\{\begin{array}{l}
          X_0 = 1 \\
          X_{n+1} \leq X_n + 2 +\frac{1}{\eta}\log(1 + 2\eta X_{n}) = X_n +\frac{1}{\eta}\log(e^{2\eta}(1 + 2\eta X_{n})) \\
          \leq X_n + \cfrac{(e^{2\eta}(1 + 2\eta X_{n}))^\epsilon}{\eta \epsilon} = X_n + \cfrac{e^{2\eta\epsilon}(1 + 2\eta X_{n})^{\epsilon}}{\epsilon\eta} .
        \end{array}\right.
\end{align}

As a first step, we introduce the function $y(t)$ on $\mathbb{R}_+$ solution on $[0, +\infty)$ of the ODE:
\begin{align}
    \left\{\begin{array}{l} y(0) = 1 \\
    y'(t) = \cfrac{e^{2\eta\epsilon}(1 + 2\eta y(t))^{\epsilon}}{\epsilon\eta}.
    \end{array} \right.
\end{align}
Let $z(t) = 1 + 2\eta y(t) $, then $z'(t) = 2\eta y'(t) $ and we get:
\begin{align}
    \left\{\begin{array}{l} z(0) = 1 + 2 \eta \\
    z'(t) = \cfrac{2 e^{2\eta\epsilon}}{\eta\epsilon}z(t)^{\epsilon}.
    \end{array} \right.
\end{align}
Solving the ode gives:

\begin{align}
    &z'(t) = \cfrac{2 e^{2\eta\epsilon}}{\eta\epsilon}z(t)^{\epsilon} \\
    \iff &\frac{z'(t)}{z(t)^{\epsilon}} = \cfrac{2 e^{2\eta\epsilon}}{\eta\epsilon} \\
    \iff &\int_{z(0)}^{z(t)} \frac{dz}{z^\epsilon} = \int_0^t \cfrac{2 e^{2\eta\epsilon}}{\eta\epsilon} du = \cfrac{2 e^{2\eta\epsilon}}{\eta\epsilon} t \\
    \iff &\frac{1}{1 - \epsilon}\left[ z(t)^{1 - \epsilon} - z(0)^{1 - \epsilon}\right] = \cfrac{2 e^{2\eta\epsilon}}{\eta\epsilon} t \\
    \iff &z(t)^{1 - \epsilon} = (1 + 2 \eta)^{1-\epsilon} +\cfrac{2 (1 - \epsilon)e^{2\eta\epsilon}}{\eta\epsilon} t \\
    \iff &z(t) = \left[(1 + 2 \eta)^{1-\epsilon} +\cfrac{2 (1 - \epsilon)e^{2\eta\epsilon}}{\eta\epsilon} t\right]^\frac{1}{1 - \epsilon}.
\end{align}

This results in:
\begin{align}
    y(t) = \frac{1}{2 \eta} \left[ \left[(1 + 2 \eta)^{1-\epsilon} +\cfrac{2 (1 - \epsilon)e^{2\eta\epsilon}}{\eta\epsilon} t\right]^\frac{1}{1 - \epsilon} - 1 \right].
\end{align}

We can thus upper-bound $X_n$ as follows: 
\begin{align}
    X_n \leq \frac{1}{2 \eta} \left[(1 + 2 \eta)^{1-\epsilon} +\cfrac{2e^{2\eta}}{\eta\epsilon} n\right]^\frac{1}{1 - \epsilon}
\end{align}

We choose $\epsilon = \frac{1}{\log n}$ for $n\geq 8 >e^2$ and obtain:
\begin{align}
    X_n &\leq \frac{1}{2 \eta} \left[(1 + 2 \eta)^{1-\frac{1}{\log n}} +\cfrac{2e^{2\eta}}{\eta} n\log n\right]^{1+\frac{1}{\log n - 1}} \\
    &\leq \frac{1}{2 \eta} \left[1 + 2 \eta +\cfrac{2e^{2\eta}}{\eta} n\log n\right]^{1+\frac{1}{\log n - 1}} \\
    \log X_n &\leq -\log (2\eta) + \left(1+\frac{1}{\log n - 1}\right)\log\left[1 + 2 \eta +\cfrac{2e^{2\eta}}{\eta} n\log n\right] \\
    &\leq -\log (2\eta) + \left(1+\frac{1}{\log n - 1}\right)\left[2\log 2 + \log(1 + 2 \eta) +\log\left(\cfrac{2e^{2\eta}}{\eta} n\log n\right)\right]\\
    &\leq -\log \eta - \log 2 + \left(1+\frac{1}{\log n - 1}\right)\left[3\log 2 + 4 \eta - \log \eta + \log n + \log \log n\right]\\
    % &= - 2 \log \eta + 2 \log 2 + 4\eta + \log n + \log\log n + \frac{\log n}{\log n - 1} + \kappa(n)\\
    &= - 2 \log \eta + 2 \log 2 + 4\eta + \log n + \log\log n + 1 + \kappa(n) \\
    X_n &\leq \exp\left(- 2 \log \eta + 2 \log 2 + 4\eta + \log n + \log\log n + 1 + \kappa(n)\right) \\
    &\leq \frac{4e^{4\eta+1}}{\eta^2}n\log n \exp\left(\kappa(n)\right),
\end{align}
where $\kappa(n)\doteq \frac{3log 2 + 4\eta - \log \eta + 1 + \log\log n}{\log n -1}\leq 2+3\log 2 + 4\eta - \log \eta$ granted that $n\geq e^2$. $\kappa(n)$ is generally in $o(1)$ and therefore $\exp(\kappa(n))$ will converge to 1 asymptotically. For the all-time upper bound, we use:
\begin{align}
    X_n &\leq \frac{32e^{8\eta+3}}{\eta^3}n\log n.
\end{align}
\end{proof}

\clearpage

\section{Experiments}
\label{app:Experiments}

\subsection{Domains}
\label{app:Domains}
The Random MDP domain is taken from \cite{Laroche2019,Nadjahi2019,Simao2020,Laroche2021}, where it is fully described in Section B.1.3. We set the three hyperparameters as follows: number of states = 100, number of actions = 4, connectivity of transitions = 2.

The Chain domain is borrowed from \cite{Laroche2021}, where it is fully described in Section C.1. We set hyper parameter $\beta$ to 0.7. The number of states varies from one figure to another where it is always specified.

The cliff experiment is novel to this work and is a direct adaptation of the chain algorithm where a third action is added. This new action is terminal and yields a 0 reward. Intuitively, it should only marginally slow down the algorithms, but in practice some of the algorithms get significantly slowed down.

In all domains the discount factor $\gamma$ is set to 0.99.

\subsection{Algorithms}
\label{app:Algorithms}
We use the Jekyll\&Hyde algorithm from \cite{Laroche2021}, which we recall below:

\begin{algorithm}[H]
\caption{Dr Jekyll \& Mr Hyde algorithm. After initialization of parameters and buffers, we enter the main loop. At every time step, an action, chosen by the behavioral policy, is executed in the environment to produce a transition $\tau_t$ (line 5). $\tau_t$ is stored in the replay buffer of the personality in control (either Dr Jekyll or Mr Hyde, line 6). If the trajectory is done, the algorithm samples a new personality to be in control during the next one (line 7). Then, the updates of the models start (line 8). The update for Mr Hyde's policy $\tilde{\pi}$ is performed with $Q$-kearning trained on UCB rewards. Dr Jekyll's critic $\mathring{q}$ is trained with SARSA. When $(\epsilon_t) = 0$, Jekyll\&Hyde amounts to on-policy expected updates from a single replay buffer.}
\textbf{Input:} Scheduling of exploration $(\epsilon_t)$, scheduling of off-policiness $(o_t)$ and actor learning rate $\eta$.
\begin{algorithmic}[1]
    \State Initialize Dr Jekyll's replay buffer $\mathring{D}=\emptyset$, actor $\mathring{\pi}$, and critic $\mathring{q} $. \Comment{exploitation agent}
    \State Initialize Mr Hyde's replay buffer $\tilde{D}=\emptyset$, and policy $\tilde{\pi}$. \Comment{exploration agent}
    % \State Initialize $\tilde{D}=\emptyset$ and parameters $\tilde{\theta}$ and $\tilde{\omega}$ of Mr Hyde's actor $\tilde{\pi}\doteq \pi_{\tilde{\theta}}$ and critic $\tilde{q} \doteq q_{\tilde{\omega}}$.
    \State Set the behavioural policy and working replay buffer to Dr Jekyll's: $\pi_{b} \leftarrow \mathring{\pi}$ and $D_{b} \leftarrow \mathring{D}$.
    % \State Sample initial state: $s_0\sim p_0$.
    \For{$t=0$ to $\infty$}
        % Online behaviour 
        \State Sample a transition $\tau_t = \langle s_t,a_t\sim\pi_{b}(\cdot|s_t),s_{t+1}\sim p(\cdot|s_t,a_t),r_t\sim r(\cdot|s_t,a_t)\rangle$.
        \State Add it to the working replay buffer $D_b\leftarrow D_b\cup\{\tau_t\}$.
        \InlineIfThen{$\tau$ was terminal,}{$(\pi_{b}, D_b) \leftarrow (\tilde{\pi}, \tilde{D})$ w.p. $\epsilon_t$, $(\pi_{b}, D_b) \leftarrow (\mathring{\pi}, \mathring{D})$ otherwise.}
        % Models updates
        % \For{$K$ update steps}
        \If{Update step,}
        % \State Perform stochastic update step on Dr Jekyll's critic $\mathring{\omega}$ with $\tau$ (depends on its algorithm).
        % \State \multiline{Perform deterministic update step in state $s$ on both actors $\theta\in\{\mathring{\theta},\tilde{\theta}\}$ with their respective critics $q\in\{\mathring{q},\tilde{q}\}$, policies $\pi\in\{\mathring{\pi},\tilde{\pi}\}$, and actor learning rates $\eta\in\{\mathring{\eta},\tilde{\eta}\}$:}
            \State $\tau \doteq \langle s,a,s',r \rangle \sim \tilde{D}$ w.p. $o_t$, $\tau \doteq \langle s,a,s',r \rangle \sim \mathring{D}$ otherwise.
            \State Update Mr Hyde's policy $\tilde{\pi}$ on $\tau$. \Comment{with $Q$-learning trained on UCB rewards}
            \State Update Dr Jekyll's critic $\mathring{q}$ on $\tau$. \Comment{with SARSA update}
            \State Expected update of Dr Jekyll's actor $\mathring{\pi}$ in state $s$. \Comment{with Eq. \eqref{eq:expected_update}}
        \EndIf
        % \EndFor
    \EndFor
\end{algorithmic}
% \vspace{-10pt}
\label{alg:J&H}
\end{algorithm}

Step 12 requires the use of approximate expected policy update \citep{Silver2014,lillicrap2015continuous,Ciosek2018}:
    \begin{align}
         \widehat{U}(\theta,s) &\doteq \sum_{a\in\mathcal{A}} \mathring{q}(s, a) \nabla_{\theta} \pi(a|s). \label{eq:expected_update}
    \end{align}

We set Jekyll\&Hyde hyperparameters as follows:
\begin{itemize}
    \item Critic learning rate: $\eta_c=0.1$, and critic initialization: $q_0=0$,
    \item Scheduling of exploration: $\epsilon_t=0$ in \textsc{NoExplo}, $\epsilon_t=\frac{10}{\sqrt{t}}$ in \textsc{LowOffPol}, and $\epsilon_t=\frac{10}{\sqrt{t}}$ in \textsc{HiOffPol},
    \item Scheduling of off-policiness: $o_t=0$ in \textsc{NoExplo}, $o_t=\frac{10}{\sqrt{t}}$ in \textsc{LowOffPol}, and $o_t=0.5$ in \textsc{HiOffPol},
    \item In all the experiments, Mr Hyde is trained with $Q$-learning on a $\gamma$ discounted objective from UCB rewards: $\tilde{r}(s,a)\doteq\frac{1}{\sqrt{n_{s,a}}}$. It has the same learning rate and initialization as the critic of Dr Jekyll.
\end{itemize}

% \subsection{Additional experimental results}
% \label{app:AdditionalExperiments}
\end{document}